\documentclass{article}

     \usepackage[preprint]{neurips_2021}

\usepackage{microtype}
\usepackage{graphicx}
\usepackage{subfigure}
\usepackage{booktabs} 
\usepackage{chngpage}

\usepackage[utf8]{inputenc} 
\usepackage[T1]{fontenc}    
\usepackage{hyperref}       
\usepackage{url}            
\usepackage{booktabs}       
\usepackage{amsfonts}       
\usepackage{nicefrac}       
\usepackage{microtype}      
\usepackage{graphicx}
\usepackage{amsmath}
\usepackage{amssymb}
\usepackage{bbm}
\usepackage{xcolor}
\usepackage{amsthm}
\usepackage{multirow}
\usepackage{dblfloatfix} 
\usepackage{wrapfig}
\usepackage{hhline}
\newtheorem{theorem}{Theorem}
\newtheorem{remark}{Remark}

\title{Emergent Properties of Foveated Perceptual Systems}

\author{%
Arturo Deza$^{1,2}$\\
Center for Brains, Minds and Machines \\
Massachusetts Institute of Technology$^{1}$ \\
deza@mit.edu
\And
Talia Konkle$^2$ \\
Department of Psychology \\
Harvard University$^2$ \\
talia\_konkle@harvard.edu
}
\begin{document}

\maketitle

\begin{abstract}
The goal of this work is to characterize the representational impact that foveation operations have for machine vision systems, inspired by the foveated human visual system, which has higher acuity at the center of gaze and texture-like encoding in the periphery. To do so, we introduce models consisting of a first-stage \textit{fixed} image transform followed by a second-stage \textit{learnable} convolutional neural network, and we varied the first stage component. The primary model has a foveated-textural input stage, which we compare to a model with foveated-blurred input and a model with spatially-uniform blurred input (both matched for perceptual compression), and a final reference model with minimal input-based compression. We find that: 1) the foveated-texture model shows similar scene classification accuracy as the reference model despite its compressed input, with greater i.i.d. generalization than the other models; 2) the foveated-texture model has greater sensitivity to high-spatial frequency information and greater robustness to occlusion, w.r.t the comparison models; 3) both the foveated systems, show a stronger center image-bias relative to the spatially-uniform systems even with a weight sharing constraint. Critically, these results are preserved over different classical CNN architectures throughout their learning dynamics. Altogether, this suggests that foveation with peripheral texture-based computations yields an efficient, distinct, and robust representational format of scene information, and provides symbiotic computational insight into the representational consequences that texture-based peripheral encoding may have for processing in the human visual system, while also potentially inspiring the next generation of computer vision models via spatially-adaptive computation. Code + Data available here: \href{https://github.com/ArturoDeza/EmergentProperties}{https://github.com/ArturoDeza/EmergentProperties}.
\end{abstract}

\vspace{-10pt}
\section{Introduction}
\label{intro}
In the human visual system, incoming light is sampled with different resolution across the retina, a stark contrast to machines that perceive images at uniform resolution. One account for the nature of this \textit{foveated} (spatially-varying) array in humans is related purely to sensory efficiency (biophysical constraints)~\citep{land2012animal,eckstein2011visual}, e.g., there is only a finite amount of retinal ganglion cells (RGC) that can relay information from the retina to the Lateral Geniculate Nucleus (LGN) constrained by the thickness of the optic nerve. Thus it is ``more efficient'' to have a moveable high-acuity fovea, rather than a non-moveable uniform resolution retina when given a limited number of photoreceptors as suggested in~\citet{akbas2017object}. Machines, however do not have such wiring/resource constraints -- and with their already proven success in computer vision~\citep{lecun2015deep} -- this raises the question if a foveated inductive bias is necessary for vision at all.

However, it is also possible that foveation plays a functional role at the \textit{representational level}, which may confer perceptual advantages -- as most computational approaches have mainly focused on saccade planning~\citep{geisler2006visual,mnih2014recurrent,elsayed2019saccader,dauce2020dual}. This idea has remained elusive in computer vision, but popular in vision science, and has been explored both psychophysically~\citep{loschky2019contributions} and computationally~\citep{poggio2014computational,cheung2017emergence,han2020scale}. 
Other works that have suggested representational advantages of foveation include the work of~\citet{pramod2018peripheral}, where blurring the image in the periphery gave an increase in object recognition performance of computer vision systems by reducing their false positive rate. In ~\citet{wu2018learning}'s GistNet, directly introducing a dual-stream foveal-peripheral pathway in a neural network boosted object detection performance via scene gist and contextual cueing. Relatedly, the most well known example of work that has directly shown the advantage of peripheral vision for scene processing in humans is~\citet{wang2017central}'s dual stream CNN that modelled the results of~\citet{larson2009contributions} with a log-polar transform and adaptive Gaussian blurring (RGC-convergence). Taken together, these studies present support for the idea that foveation has useful \textit{representational consequences} for perceptual systems. Further, these computational examples have symbiotic implications for understanding biological vision, indicating what the functional advantages of foveation in humans may be, via functional advantages in machine vision systems.

\begin{figure}[t]
\centering
\includegraphics[width=0.85\columnwidth,clip=true,draft=false,]{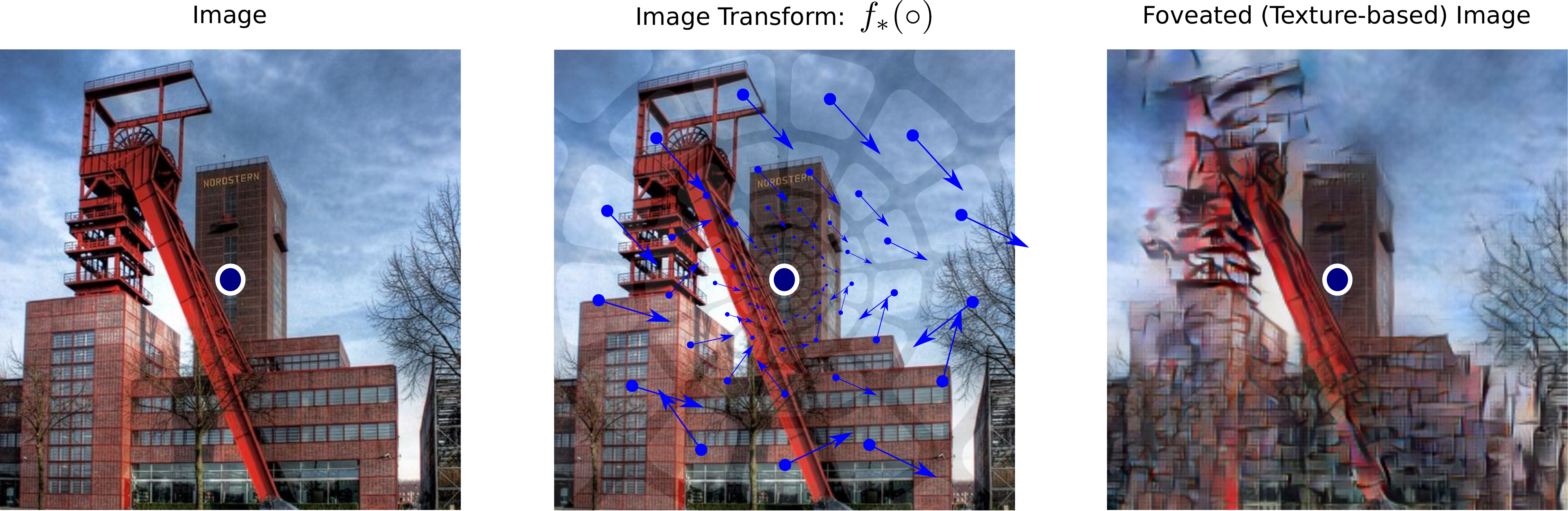}\caption{A cartoon illustrating how a biologically-inspired foveated image (texture-based) is rendered resembling a human visual \textit{metamer} 
via the foveated feed-forward style transfer model of~\citet{deza2018towards}. Here, each receptive field is locally perturbed with noise in its latent space in the direction of their equivalent texture representation (blue arrows) resulting in \textit{visual crowding} effects that warp the image locally in the periphery~\citep{balas2009summary,freeman2011metamers,rosenholtz2016capabilities}. These effects are most noticeable far away from the navy dot which is the simulated center of gaze (foveal region) of an observer under certain viewing conditions.}
\vspace{-15pt}
\label{fig:Metamer_Transform}
\end{figure}

Importantly, none of these studies introduce the notion of \textit{texture representation} in the periphery -- a key property of peripheral computation as posed in~\citet{rosenholtz2016capabilities}. What functional consequences does this well-known texture-based coding in the visual periphery have, if any, on the nature of later stage visual representation? Here we directly examine this question. Specifically, we introduce \textit{perceptual systems}: as two-stage models that have an image transform stage followed by a deep convolutional neural network. The primary model class of interest possesses a first stage image transform that mimics texture-based foveation via \textit{visual crowding}~\citep{levi2011visual,pelli2008crowding,doerig2019beyond,doerig2019crowding} in the periphery as shown in Figure~\ref{fig:Metamer_Transform}~\citep{deza2018towards}, rather than Gaussian
 blurring~\citep{wang2017central,pramod2018peripheral,malkin2020cudaoptimized} or compression~\citep{patney2016towards,kaplanyan2019deepfovea}. These rendered images capture image statistics akin to those preserved in human peripheral vision, and resembling texture computation at the stage of area V2, as argued in~\citet{freeman2011metamers,rosenholtz2016capabilities,wallis2019image}. 
 
Our strategy is thus to compare in terms of generalization, robustness and bias these \textit{foveation-texture models} to three other kinds of models. The first comparison model class -- \textit{foveation-blur models} -- uses the same spatially-varying foveation operations but uses blur rather than texture based input. The second class -- \textit{uniform-blur models} -- uses a blur operation uniformly over the input, with the level of blur set to match the perceptual compression rates of the foveation-texture nets. Finally, the last comparison model class is the \textit{reference}, which has minimal distortion, and serves as a perceptual upper bound from which to assess the impact of these different first-stage transforms. 

Note that our approach is different from the one taken by~\citet{wang2017central}, who have built foveated models that fit results to human behavioural data like those of ~\citet{larson2009contributions}. Rather, our goal is to explore the emergent properties in CNNs with \textit{texture-based foveation} on scene representation compared to their controls agnostic to any behavioural data or expected outcome. Naturally, the results of our experimental paradigm is symbiotic as it can shed light into both the importance of texture-based peripheral computation in humans, and could also suggest a new inductive bias for advanced machine perception in scenes.


\section{Perceptual Systems}
We define perceptual systems as \textit{two-stage} models with an image transform (stage 1,~$f(\circ):\mathbb{R}^D\rightarrow\mathbb{R}^D$), that is relayed to a deep convolutional neural network (stage 2,~$g(\circ):\mathbb{R}^D\rightarrow\mathbb{R}^d$). Note that the first transform stage is a \textit{fixed} operation over the input image,
while the second stage has \textit{learnable} parameters.
In general, the perceptual system $S(\circ)$, with retinal image input $I:\mathbb{R}^D$ is defined as: 
\begin{equation}
\label{eq:Perceptual_System}
    S(I) = g(f(I))
\end{equation}

\begin{figure}[!t]
\centering
\includegraphics[width=0.95\columnwidth,clip=true,draft=false,]{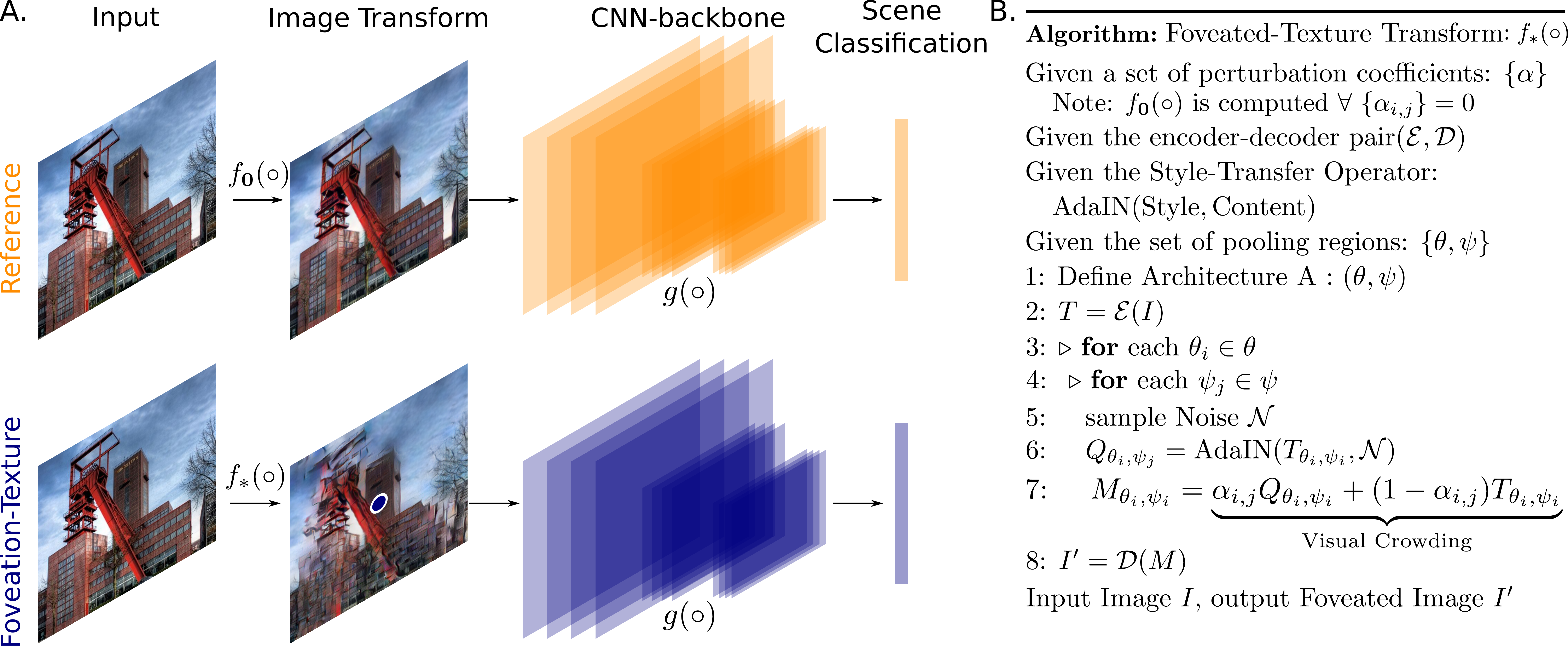}\caption{
\underline{A.} Two of the four perceptual systems: Reference (top row) and Foveation-Texture (bottom row), where each system receives an image as an input, applies an image transform $(f(\circ))$, which is then relayed to a CNN architecture $(g(\circ))$ for scene classification. Reference provides an undistorted baseline as a perceptual upper-bound, while Foveation-Texture uses a visual crowding model that distorts the image with spatially-varying texture computation (shown on right)
\underline{B.} The algorithm of how the biologically inspired \textit{Foveation-Texture} transform works which enables effects of \textit{visual crowding} in the periphery (mainly steps 5-7).}\label{fig:Methods_Figure}
\vspace{-10pt}
\end{figure}

\vspace{-5pt}
Such two-stage models have been growing in popularity, and the reasons these models are designed to \textit{not} be fully end-to-end differentiable is mainly to \textit{force} one type of computation into the first-stage of a system such that the second-stage $g(\circ)$ must figure out how to capitalize on such forced transformation and thus assess its $f(\circ)$ representational consequences (See Figure~\ref{fig:Methods_Figure}). For example,~\citet{parthasarathy2020self} successfully imposed V1-like computation in stage 1 to explore the learned role of texture representation in later stages with a self-supervised objective, and~\citet{dapello2020simulating} found that fixing V1-like computation also at stage 1 aided adversarial robustness. At a higher level, our objective is similar where we would like to force a texture-based peripheral coding mechanism (loosely inspired by V2; \citealp{ziemba2016selectivity}) at the first stage to check if the perceptual system (now foveated) will learn to pick-up on this newly made representation through $g(\circ)$ and make `good' use of it potentially shedding light on the \textit{functionality} hypothesis for machines and humans. 


\subsection{Stage 1: Image Transform}
\vspace{-5pt}
To model the computations of a texture-based foveated visual system, we employed the model of~\citet{deza2018towards}~(henceforth \textit{Foveated-Texture Transform}). This model is inspired by the metamer synthesis model of~\citet{freeman2011metamers}, where new images are rendered to have locally matching texture statistics~\citep{portilla2000parametric,balas2009summary} in greater size pooling regions of the visual periphery with structural constraints. Analogously, the~\citet{deza2018towards} Foveation Transform uses a foveated feed-forward style transfer~\citep{huang2017arbitrary} network to latently perturb the image in the direction of its locally matched texture (see Figure~\ref{fig:Metamer_Transform}). Altogether, $f:\mathbb{R}^D\rightarrow\mathbb{R}^D$ is a convolutional auto-encoder that is non-foveated when the latent space is un-perturbed: $f_\mathbf{0}(I) = \mathcal{D}(\mathcal{E}(I))$, but foveated $(\circ_{\Sigma})$ when the latent space is perturbed via localized style transfer: $f_\mathbf{*}(I) = \mathcal{D}(\mathcal{E}_{\Sigma}(I))$, for a given encoder-decoder~$(\mathcal{E},\mathcal{D})$ pair.

\begin{figure*}[!t]
\centering
\includegraphics[width=0.95\columnwidth,clip=false,draft=false,]{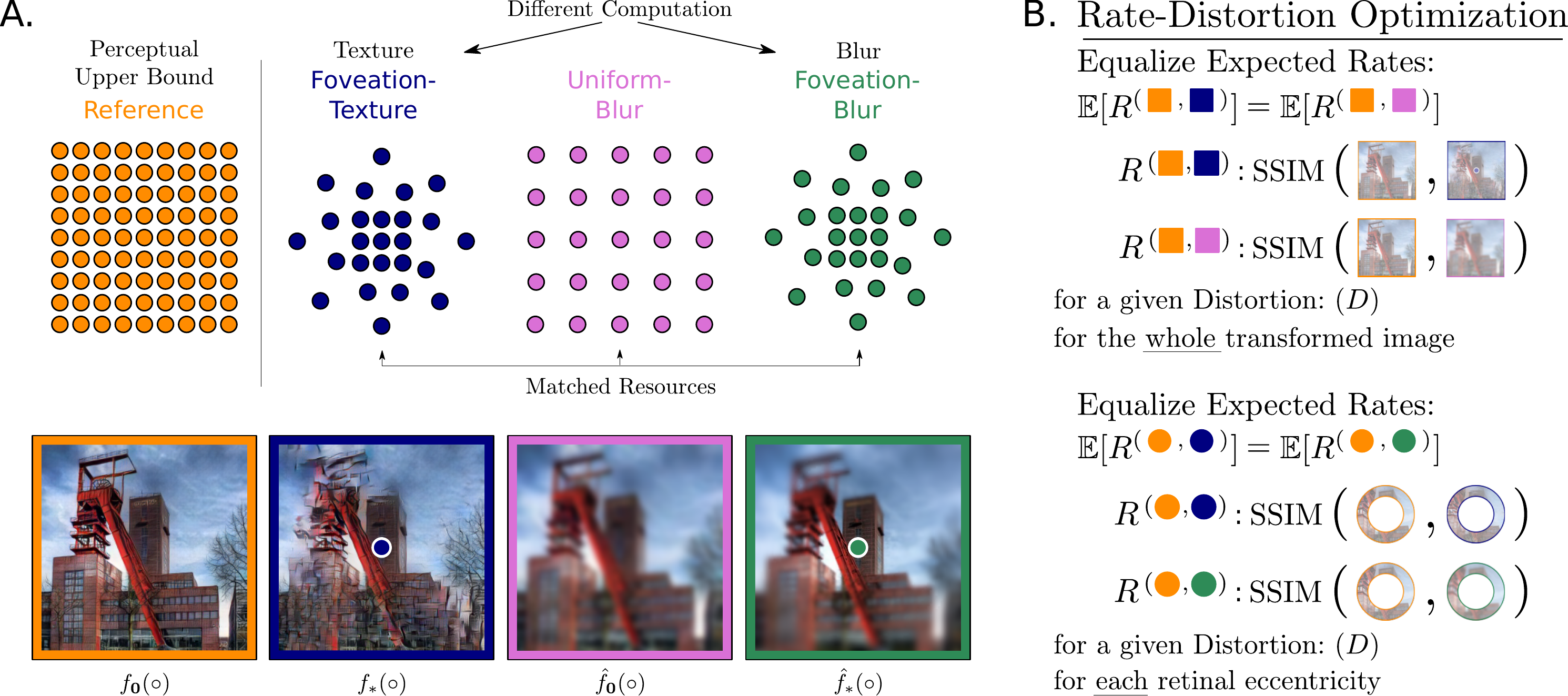}\caption{
\underline{A.} Two perceptually matched-resource controls to Foveation-Texture are introduced. Middle-Right, orchid: uniform blurring emulating a matched-resource non-foveated visual system (Uniform-Blur); Far-Right, seagreen: adaptive gaussian blurring (Foveation-Blur) emulating a matched resource blur-based foveated system. \underline{B.} A Rate-Distortion Optimization procedure is summarized where we find the hyper-parameters of the new matched-resource image transforms $\{(\hat{f}_\mathbf{0}(\circ),\hat{f}_\mathbf{*}(\circ))\}$ to Foveation-Texture via expected SSIM matching over the validation set.
}\label{fig:Critical_Manipulations}
\vspace{-10pt}
\end{figure*}

Note that with proper calibration, the resulting distorted image can be a visual metamer (for a human), which is a carefully perturbed image perceptually indistinguishable from its reference image~\citep{freeman2011metamers,rosenholtz2012summary,feather2019metamers,vacher2020texture}. However, importantly in the present work, we exaggerated the strength of these texture-driven distortions (beyond the metameric boundary), as our aim here is to understand the implications of this kind of texturized peripheral input  on later stage representations (e.g. following a similar approach as \citet{dapello2020simulating}). By having an extreme manipulation, we reasoned this would accentuate the consequences of these distortions, making them more detectable in our subsequent experiments.

\subsection{Stage 2: Convolutional Neural Network backbone}
\vspace{-5pt}
The transformed images (stage 1) are passed into a standard convolutional neural network architecture. 
Here we tested two different base architectures: AlexNet~\citep{krizhevsky2012imagenet}, and ResNet18~\citep{he2016deep}. The goal of running these experiments on two different hierarchically local architectures is to let us examine the consequences across all image transforms (with our main focus towards texture-based foveation) that are robust to these different network architectures.  
Further, this CNN backbone $(g:\mathbb{R}^D\rightarrow\mathbb{R}^d)$ should not be viewed in the traditional way of an end-to-end input/output system where the input is the retinal image $(I)$, and the output is a one-hot vector encoding a $d$-class-label in $\mathbb{R}^d$. Rather, the CNN $(g)$ acts as a loose proxy of higher stages of visual processing (as it receives input from $f$), analogous to the 2-stage model of~\citet{lindsey2019unified}.

\subsection{Critical Manipulations: Foveated vs Non-Foveated Perceptual Systems}
\vspace{-5pt}
Now, we can define the first two of the four perceptual systems that will perform 20-way scene categorization: \textit{Foveation-Texture}, receives an image input, applies the foveation-texture transform $f_{\mathbf{*}}(\circ)$, and relays it through the CNN $g(\circ)$. Similarly, \textit{Reference} performs a non-foveated transform $f_{\mathbf{0}}(\circ)$, where images are sent through the same convolutional auto-encoder $\mathcal{D}(\mathcal{E}(I))$ of $f_{\mathbf{*}}(\circ)$, but with the parameter that determines the degree of texture style transfer set to $0$ -- producing an upper-bounded, compressed and non-foveated reference image -- then relayed through the CNN $g(\circ)$. Both of these systems are depicted in Figure~\ref{fig:Methods_Figure}~(\underline{A}).
As the foveation-texture model has less information from the input, relative to the reference networks, we next designed two further comparison models which have a comparable amount of information after the input stage, but with different amounts of blurring in the stage 1 operations. To create matched-resources systems, our broad approach was to use a Rate-Distortion (RD) optimization procedure ~\citep{balle2016end} to match information between the stage 1 operations, given the SSIM~\citep{wang2004image} image quality assessment (IQA) metric.  

Specifically, to create matched-resource \textit{Uniform-Blur},  we identified the standard deviation of the Gaussian blurring kernel (the `distortion' $\mathcal{D}$), such that we could render a perceptually resource-matched Gaussian blurred image -- w.r.t Reference -- that matches the perceptual transmission `rate' $\mathcal{R}$ of Foveation-Texture via the SSIM perceptual metric~\citep{wang2004image}. This procedure yields a model class with uniform blur across the image, but with matched stage 1 information content as the Foveation-Texture. And, to create matched-resource \textit{Foveation-Blur}, we carried our this same RD optimization pipeline per eccentricity ring (assuming homogeneity across pooling regions at the same eccentricity), thus finding a set of blurring coefficients that vary as a function of eccentricity. This procedures yielded a different matched-resource model class, this time with spatially-varying blur.  Figure~\ref{fig:Critical_Manipulations} (B) summarizes our solution to this problem. Details of the RD Optimization are presented in Appendix~\ref{sec:Supplementary_Foveated}. 

Ultimately, it is important to note that the selection of the perceptual metric (SSIM in our case), plays a role in this optimization procedure, and sets the context in which we can call a network ``resource-matched''. We selected SSIM given its monotonic relationship of distortions to human perceptual judgements, symmetric upper-bounded nature, sensitivity to contrast, local structure and spatial frequency, and popularity in the Image Quality Assessment (IQA) community. However to anticipate any possible discrepancy in the interpretability of our future results, we additionally computed the Mean Square Error (MSE), MS-SSIM, and 11 other IQA metrics as recently explored in~\cite{2020arXiv200501338D} to compare all other image transforms to the Reference on the testing set. Our logic is the following: if the MSE is \textit{greater}($\uparrow$) for Foveation-Texture compared to Foveation-Blur and Uniform-Blur, then the current distortion levels place Foveation-Texture at a resource `disadvantage' relative to the other transforms, and any interesting results would not only hold but also be \textit{strengthened}. This same logic applies to the other IQA metrics contingent on their direction of \textit{greater} distortion. Indeed, these patterns of results were evident across IQA metrics -- except those tolerant to texture such as DISTS~\citep{ding2020image} -- as shown in Table~\ref{table:Main_IQA_Comparison}, and Appendix~\ref{sec:IQA_Supplement}.

\begin{table}[h]
\centering
\tiny
{
\centering
\begin{tabular}{|c|c|c|c|c|c|c|}
\hline
 (mean$\pm$std) & SSIM \textit{(Matched)} & MS-SSIM $(\downarrow)$ & MSE $(\uparrow)$ & Mutual Information $(\downarrow)$ & NLPD $(\uparrow)$ & DISTS $(\uparrow)$\\
 \hline
Reference & 1.0 & 1.0 & 0.0 & $7.39\pm0.52$ & 0 & 0 \\
\hline
Foveation-Texture & $\mathbf{0.58\pm0.11}$ & $\mathbf{0.20\pm0.03}$ & $\mathbf{976.78\pm522.22}$ & $\mathbf{1.40\pm0.42}$ & $\mathbf{0.75\pm0.16}$ & $0.20\pm0.03$\\
Uniform-Blur & $\mathbf{0.57\pm0.15}$ & $0.36\pm0.03$ & $458.67\pm277.13$ & $1.86\pm0.58$ & $0.40\pm0.09$ & $\mathbf{0.36\pm0.03}$\\
Foveation-Blur & $\mathbf{0.58\pm0.15}$ & $0.36\pm0.03$ & $507.35\pm302.71$ & $1.84\pm0.56$ & $0.45\pm0.11$ & $\mathbf{0.35\pm0.03}$\\
\hline
\end{tabular}
\vspace{5pt}
\caption{Comparing Image Transforms \textit{wrt} Reference. Arrows indicate direction of \textit{greater} distortion.}
\label{table:Main_IQA_Comparison}
}
\end{table}

\vspace{-20pt}
\section{Experiments}
\label{sec:Experiments}

\begin{wrapfigure}{r}{0.4\textwidth}
\vspace{-15pt}
\centering
\includegraphics[width=0.4\columnwidth,clip=true,draft=false,]{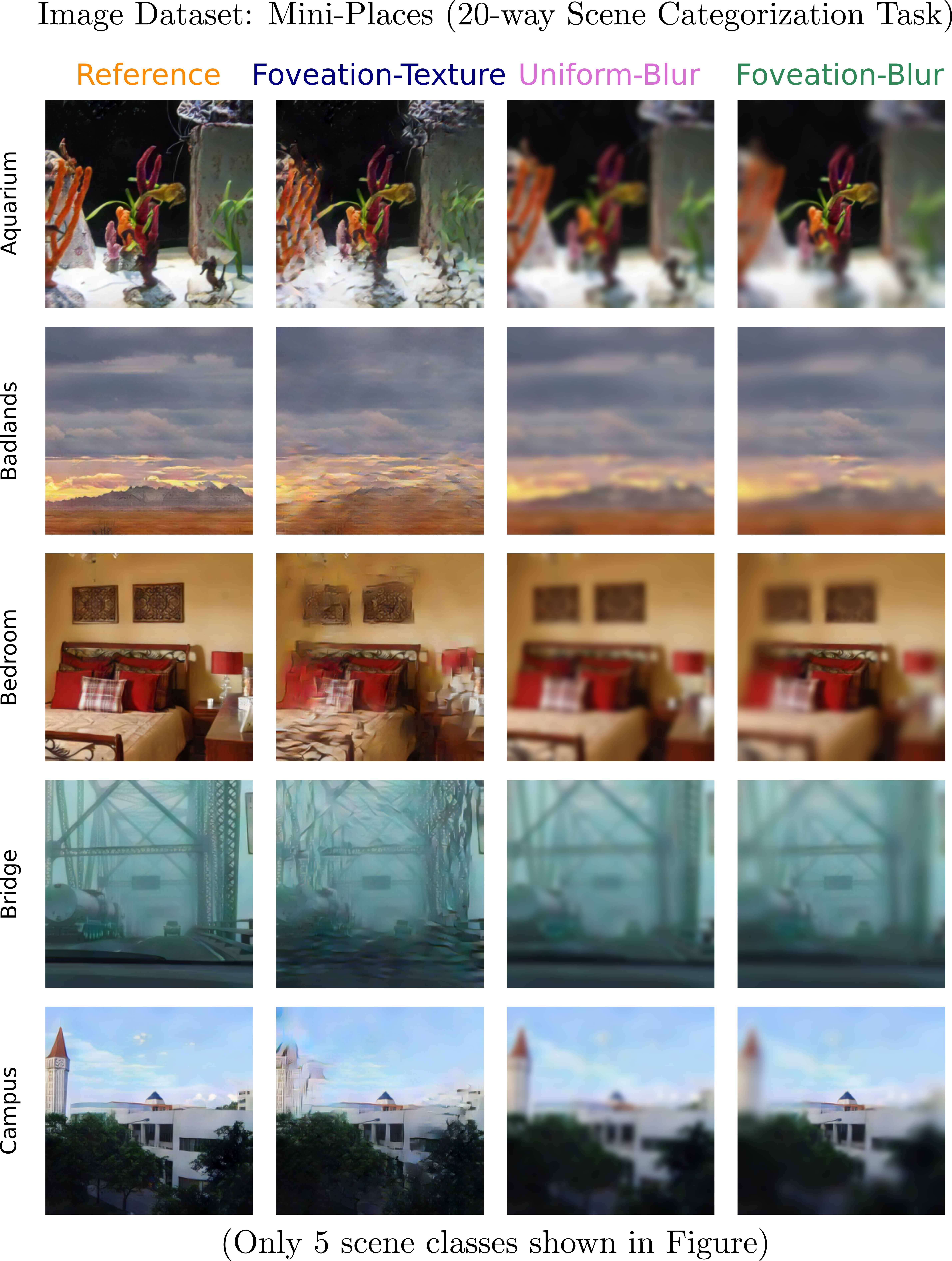}\caption{Five example images from the 20 scene categories are shown, after being passed through the first stage of each perceptual system.}\label{fig:Samples_and_Methods}
\vspace{-10pt}
\end{wrapfigure}
Altogether, the 4 previously introduced perceptual systems help us answer three key questions that we should have in mind throughout the rest of the paper: \underline{1)} Foveation-Texture vs Reference will tell us how a texture-based foveation mechanism will compare to its perceptual upper-bound -- shedding light into arguments about computational efficiency. \underline{2)} Foveation-Texture vs Foveation-Blur will tell us if any potentially interesting pattern of results is due to the \textit{type/stage} of foveation. This will help us measure the contributions of the adaptive texture coding vs adaptive gaussian blurring; \underline{3)} Foveation-Texture vs Uniform-Blur will tell us how do these perceptual systems (one foveated, and the other one not) behave when allocated with a fixed number of perceptual resources under certain assumptions -- potentially shedding light on why biological organisms like humans have foveated texture-based computation in the visual field instead of uniform spatial processing like modern machines.


\textbf{Dataset:} All previously introduced models were trained to perform 20-way scene categorization. Scene categories were selected from the Places2 dataset~\citep{zhou2017places}, and were re-partitioned into a new 4500 images per category for training, 250 per category for validation, and 250 per category for testing. The categories included were: aquarium, badlands, bedroom, bridge, campus, corridor, forest path, highway, hospital, industrial area, japanese garden, kitchen, mansion, mountain, ocean, office, restaurant, skyscraper, train interior, waterfall. Samples of these scenes coupled with their image transforms can be seen in Figure~\ref{fig:Samples_and_Methods}.

\textbf{Networks:} \underline{Training:} Convolutional neural networks of the stage 2 of each perceptual system were trained which resulted in 40 image-transform based networks~\textit{per architecture} (AlexNet/ResNet18): 10 Foveation-Texture, 10 Reference, 10 Uniform-Blur, 10 Foveation-Blur; totalling 80 trained networks to compute relevant error bars shown in all figures (standard deviations, not standard errors) and to reduce effects of randomness driven by the particular network initialization. All systems were paired such that their stage 2 architectures $g(\circ)$ started with the \textit{same random weight initialization} prior to training. \underline{Testing:} The networks of each perceptual system were tested on \textit{the same} type of image distribution they were trained on. \underline{Learning Dynamics}: Available in Appendix~\ref{sec:Appendix_Train_Test_Dyn}.

\begin{figure*}[!t]
\centering
\includegraphics[width=1.0\columnwidth,clip=false,draft=false,]{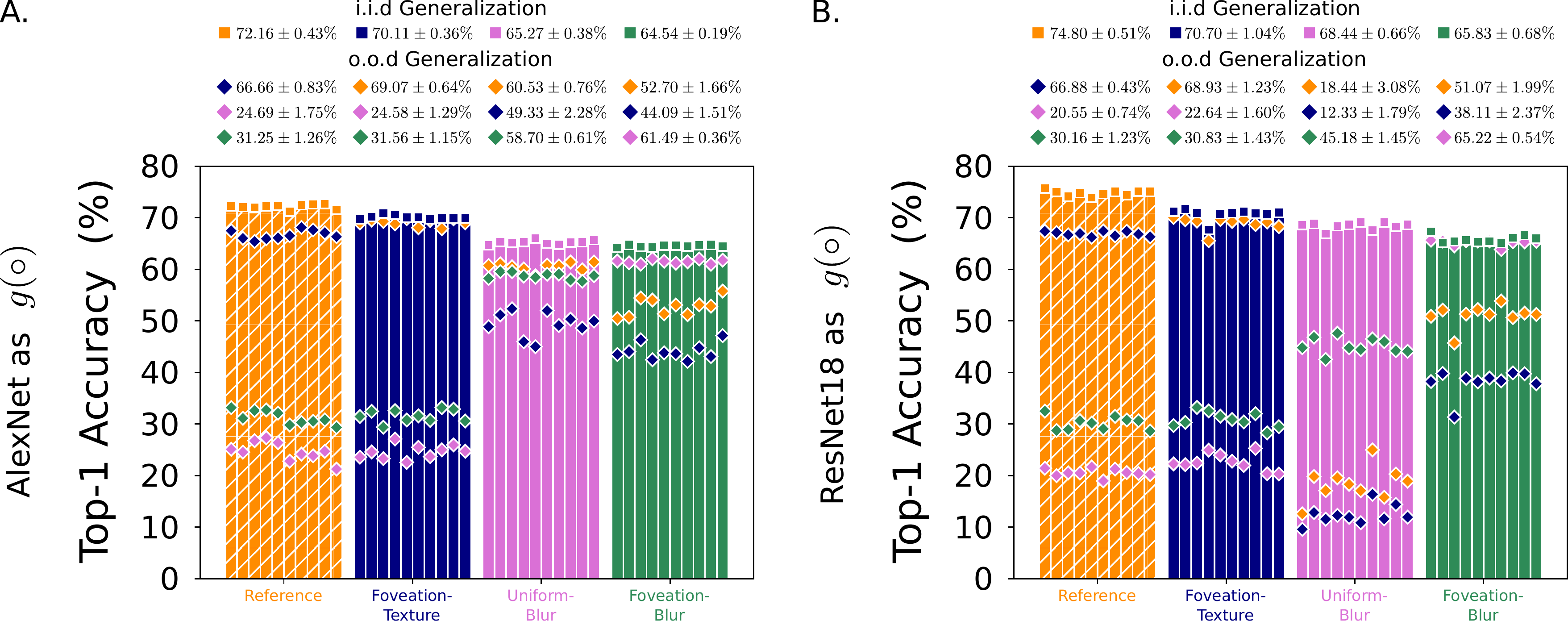}\caption{Scene Categorization Accuracy of AlexNet and ResNet18 as $g(\circ)$. We observe the following: Foveation-Texture has greater i.i.d. generalization than other matched-resource systems across both network architectures; Uniform-Blur's o.o.d generalization interacts with the architecture (performing worse for ResNet18, but highest for AlexNet); Foveation-Blur maintains high o.o.d. generalization independent of network architecture. Confusion Matrices can be seen in Appendix~\ref{sec:Appendix_Generalization}.}\label{fig:Results_Generalization}
\vspace{-10pt}
\end{figure*}

\vspace{-5pt}
\subsection{Texture-based foveation provides greater \textit{i.i.d.} generalization than Blur-based foveation}
\label{sec:Generalization}
\vspace{-5pt}

How well does the foveation-texture stage classify scene images (i.i.d. generalization) compared to the other matched-resource models that use blurring and the reference? The results can be seen in Figure~\ref{fig:Results_Generalization}. Each bars' height reflects overall accuracy for each of the 10 neural network backbone runs $(g(\circ))$ per system, with a \textit{square} marker at the top indicating the i.i.d. accuracy.
We found that Foveation-Texture had similar i.i.d. performance to the Reference -- which is the the undistorted perceptual upper bound, and \textit{greater} performance than both Uniform-Blur and Foveation-Blur. Thus the compression induced by foveated-texture generally maintains scene category information.

We next performed a contrived experiment where we tested how well each perceptual system could classify the stage 1 outputs of the other models. For example, we showed a set of foveated blurred images to a network trained on foveated texture images. This experiment is in essence a test of out-of-distribution \textit{(o.o.d.)} generalization. The results of these tests are also shown in Figure~\ref{fig:Results_Generalization}. For each model, the classification accuracy for the inputs from the other stage 1 images is indicated by the height of the different colored \textit{diamonds}, where the color corresponds to the stage 1 operation. 

This experiment yielded a rather complex set of patterns, that even differed depending on the architecture (AlexNet vs ResNet18 as $g(\circ)$). Generally, the Foveation-Texture model had a similar profile of generalization as the Reference model. However, the networks trained with different types of blur (Uniform-Blur \& Foveated-Blur) in some cases showed very high o.o.d. generalization -- though once again this is contingent on $g(\circ)$.


Unraveling the underlying causes to understand this last set of results sets the stage for our experiments in the rest of this section. So far it seems like Foveation-Texture has learned to properly capitalize the texture information in the periphery and still out-perform all other matched-resource systems even if heavily penalized under several IQA metrics (Table~\ref{table:Main_IQA_Comparison}) -- highlighting the critical differences in texture vs blur for scene processing. As for the interaction of Uniform-Blur with $g(\circ)$, is is likely that the residual connections are counter-productive to o.o.d. generalization (or it has overfit). Interestingly, humans have a combination of texture and adaptive-gaussian based peripheral computation~\citep{ehinger2016general}, so future work should look into the effects of continual learning, joint-training or a combined image transform (Texture + Blur) to merge gains of both i.i.d and o.o.d generalization.


\subsection{Texture-based foveated systems preserve greater high-spatial frequency sensitivity}
\label{sec:Frequency}

\begin{figure}[!t]
\centering
\includegraphics[width=1.0\columnwidth,clip=false,draft=false,]{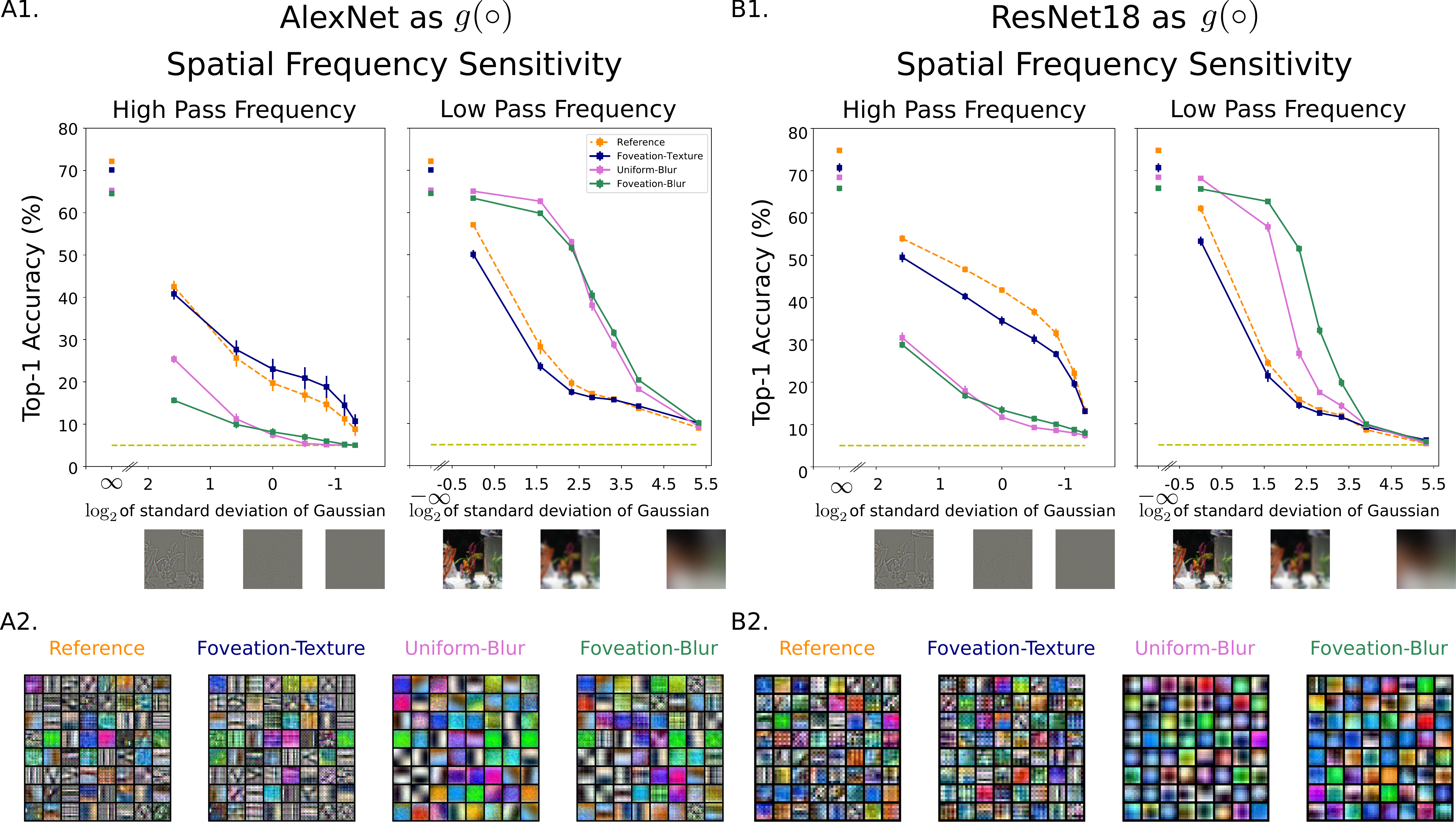}\caption{Foveation-Texture has greater sensitivity to high pass spatial frequency filtered stimuli than the Reference (contingent on the architecture for $g(\circ)$ -- See A1.,B1.), though both of these systems present notably higher sensitivity to high spatial frequencies than Uniform-Blur and Foveation-Blur. This pattern is reversed for low pass frequency stimuli applied to both color and grayscale filtered images (Appendix~\ref{sec:Appendix_Spatial_Frequency}). Visualizations of the first convolutional layer of AlexNet and ResNet18 as $g(\circ)$ (A2.,B2.) shows strong similarities of learned filters despite texture-distortion for Foveation-Texture to Reference preserving high spatial frequency Gabors; Uniform-Blur shows a strong predominance of low spatial frequency Gabors for AlexNet and low spatial frequency center-surround filters for ResNet18, and Foveation-Blur a mixture of high-low spatial frequency tuned filters.}\label{fig:High_Low_Freq}
\vspace{-15pt}
\end{figure}

We next examined whether the learned feature representations of these models are more reliant on low or high pass spatial frequency information. To do so, we filtered the testing image set at multiple levels to create both high pass and low pass frequency stimuli and assessed scene-classification performance over these images for all models, as shown in Figure~\ref{fig:High_Low_Freq}. Low pass frequency stimuli were rendered by convolving a Gaussian filter of standard deviation $\sigma=[0,1,3,5,7,10,15,40]$ pixels on the foveation transform $(f_\mathbf{0},\hat{f}_\mathbf{0},f_\mathbf{*},\hat{f}_\mathbf{*})$ outputs. Similarly, the high pass stimuli was computed by subtracting the reference image from its low pass filtered version with $\sigma=[\infty,3,1.5,1,0.7,0.55,0.45,0.4]$ pixels and adding a residual. These are the same values used in the experiments of~\citet{geirhos2018imagenettrained}.

We found that Foveation-Texture and Reference trained networks were more sensitive to High Pass Frequency information, while Foveation-Blur and Uniform-Blur were selective to Low Pass Frequency stimuli. Although one may naively assume that this is an expected result -- as both Foveation-Blur and Uniform-Blur networks are exposed to a blurring procedure -- it is important to note that: 1) the foveal resolution has been \textit{preserved} between Foveation-Texture and Foveation-Blur (See Fig.~\ref{fig:Samples_and_Methods}), thus high spatial frequency sensitivity could have still predominated in Foveation-Blur but it did not (though see Fig.~\ref{fig:High_Low_Freq} A2/B2 where these high pass Gabors are still learned, implying that higher layers in $g(\circ)$ overshadow their computation); and 2) Foveation-Texture could have also learned to develop low spatial frequency sensitivity given the crowding/texture-like peripheral distortion, but this was not the case (likely due to the weight sharing constraint embedded in the CNN architecture \citealp{elsayed2020revisiting}). Finally, the robustness to low-pass filtering of Foveation-Blur suggests that foveation via adaptive gaussian blurring may implicitly contribute to scale-invariance as also shown in~\cite{poggio2014computational,cheung2017emergence,han2020scale}.

\subsection{Texture-based foveation develops greater robustness to occlusion}
\label{sec:Occlusion}
\begin{figure}[!t]
\centering
\includegraphics[width=1.0\columnwidth,clip=true,draft=false,]{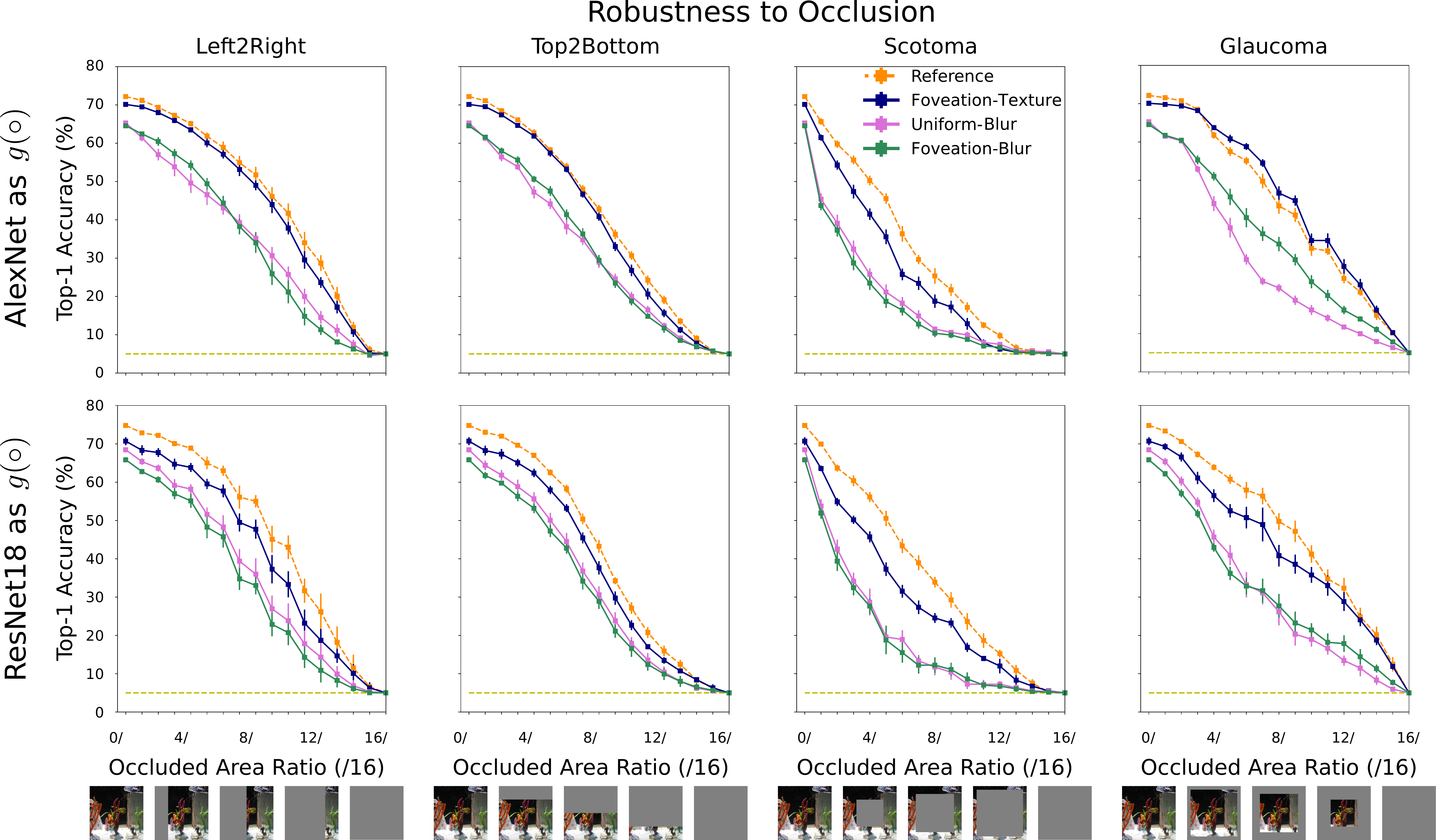}\caption{
Foveation-Texture has greater robustness than both Foveation-Blur and Uniform-Blur while roughly preserving a performance similarity to Reference (the upper bound) beyond the \textit{i.i.d.} regime. The asymmetry in performance of the Scotoma vs Glaucoma conditions for foveated models also suggests they have learned to weigh spatial information differently in the fovea vs the periphery despite a weight sharing constraint imposed through $g(\circ)$.}
\label{fig:Robustness_to_Occlusion}
\vspace{-10pt}
\end{figure}

\vspace{-5pt}
We next examined how all perceptual systems could classify scene information under conditions of visual field loss, either from left to right (left2right), top to bottom (top2bottom), center part of the image (scotoma), or the periphery (glaucoma). This manipulation lets us examine the degree to which learned representations relying on different parts of the image to classify scene categories. Critically, here we apply the occlusion \textit{after} the stage 1 operation. The results are shown in Figure~\ref{fig:Robustness_to_Occlusion}.

Overall we found that, across all types of occlusion the Foveation-Texture modules have greater robustness to occlusion than both the Foveation-Blur and Uniform-Blur models. Further, the Foveation-Texture models have nearly equivalent performance to the Reference. In contrast, both models with blurring, whether uniformly or in a spatially-varying way, were far worse at classifying scenes under conditions of visual field loss. These results highlight that the texture-based information content captured by the foveation-texture nets preserves scene category content in dramatically different way than simple lower-resolution sampling -- perhaps using the texture-bias~\citep{geirhos2018imagenettrained} in their favor; as humans too use texture as their classification strategy for scenes~\citep{renninger2004scene}. 

In addition, the Foveation-Texture model is not overfitting. As recent work has suggested an Accuracy vs Robustness trade-off where networks trained to outperform under the \textit{i.i.d.} generalization condition will do worse under other perceptual tasks -- mainly adversarial~\citep{zhang2019theoretically} -- we did not observe such trade-off and a greater accuracy did not imply lower robustness to occlusion.

\vspace{-10pt}
\subsection{Foveated systems learn a stronger center image bias than non-foveated systems}
\label{sec:Window}

\begin{figure}[!t]
\centering
\includegraphics[width=1.0\columnwidth,clip=false,draft=false,]{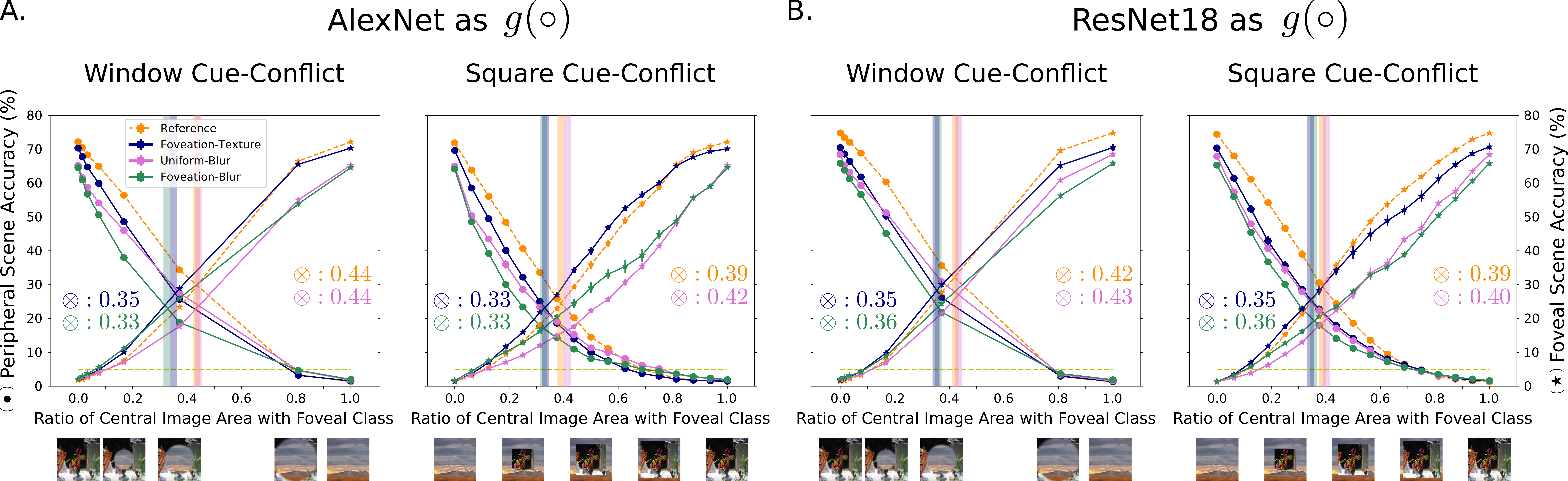}\caption{Foveated Perceptual Systems -- independent of the computation type (Foveation-Texture, Foveation-Blur) -- show stronger biases to classify hybrid scenes with the foveal region; a result also observed in humans~\citep{larson2009contributions}.}
\label{fig:Window_Cue_Conflict}
\vspace{-10pt}
\end{figure}

It is possible that foveated systems weight visual information strongly in the foveal region than the peripheral region as hinted by our occlusion results (the different rate of decay for the accuracy curves in the Scotoma and Glaucoma conditions). To resolve this question, we conducted an experiment where we created a windowed cue-conflict stimuli where we re-rendered our set of testing images with one image category in the fovea, and another one in the periphery (all aligned with a different class systematically; \textit{ex:} aquarium with badlands). We also had an additional condition where the conflicting cue was now square-like and uniformly and randomly paired with a conflicting scene class and more finely sampled. We then systematically varied the fovea-periphery visual area ratios \& re-examined classification accuracy for both the foveal and peripheral scenes (Figure~\ref{fig:Window_Cue_Conflict}). 

We found that the Foveation-Texture and Foveation-Blur transform imposed the networks $g(\circ)$ to learn to weigh information in the center of the image stronger than Reference \& Uniform-Blur for scene categorization.  A qualitative way of seeing this foveal-bias is by checking the foveal/peripheral ratio where these two accuracy lines cross. The more leftward the cross-over point $(\otimes)$, the higher the foveal bias (highlighted through the vertical bars). This result was unexpected as we initially predicted that $g(\circ)$ would weigh the peripheral information stronger as it has been implicitly regularized through a distortion. However this was not the case and our findings are similar to~\citet{wang2017central} who showed this foveal bias on a foveated system with adaptive blur with a dual-stream neural network. Thus, these results indicate that the \textit{spatially varying computation from center to periphery} is mainly responsible for the development of a center image bias \textit{even with a weight sharing constraint}. Furthermore, it is possible that one of the functions of any spatially-varying coding mechanisms in the visual field is to \textit{enforce} the perceptual system to \textit{attend} on the foveal region -- avoiding the shortcut of learning to attend the entire visual field if unnecessary ~\citep{geirhos2020shortcut}.

\vspace{-10pt}
\section{Discussion}
\label{sec:Discussion}
\vspace{-5pt}
The present work was designed to probe the impact of foveated texture-based input representations in machine vision systems. To do this we specifically compared the learned perceptual signatures in the second-stage of visual processing across a set of of networks trained on other image transforms. We found that when comparing Foveation-Texture to their matched-resource models that differed in computation: Foveation-Blur (foveated w/ adaptive gaussian blur) and Uniform-Blur (non-foveated w/ uniform blur) -- that peripheral texture encoding did lead to specific representational signatures, particularly greater i.i.d generalization, preservation of high-spatial frequency sensitivity, and robustness to occlusion -- even as high as its perceptual upper bound (Reference). We also found that foveation (in general) seems to induce a \textit{focusing mechanism}, servicing the foveal/central region -- whereas neither a perceptually upper-bounded system (Reference) or a non-foveated compressed system (Uniform-Blur) did \textit{not} develop as strongly. 

The particular consequences of our foveation stage raises interesting future directions about what computational advantages could arise when trained on object categorization~\citep{pramod2018peripheral} coupled with eye-movements~\citep{akbas2017object,deza2017attention}, as objects are typically centered in view and have different hierarchical/compositional priors than scenes~(\citet{zhou2014object,deza2020hierarchically}) in addition to different processing mechanisms~(\citet{renninger2004scene,ehinger2016general}). We are currently exploring the impact of these \textit{foveated texture-based} representational signatures on shape vs texture bias for object recognition similar to~\citet{geirhos2018imagenettrained} and \citet{hermann2020origins}, and assessing their interaction with scene representation. 

Further, a future direction is investigating the effects of texture-based foveation to \textit{adversarial robustness}. Motivated by the recent work of~\citet{dapello2020simulating} which has shown promise of adversarial robustness via enforcing stochasticity and V1-like computation by obeying the Nyquist sampling frequency of these filters w.r.t the image~\citep{serre2007robust} in addition to a natural gamut of orientations and frequencies as studied in~\cite{de1982orientation}, it raises the question of how much we can further push for robustness in hybrid perceptual systems like these, drawing on even \textit{more} biological mechanisms. Works such as ~\citet{luo2015foveation} and recently~\citet{reddy2020biologically,kiritani2020recurrent} have already taken steps in this direction by coupling fixations with a spatially-varying retina. However, the representational impact of texture-based foveation on adversarial robustness, and its symbiotic implication for human vision still remains an open question.

\newpage

\bibliography{neurips_2021}
\bibliographystyle{neurips_2021}

\newpage

\section*{Checklist}

\begin{enumerate}

\item For all authors...
\begin{enumerate}
  \item Do the main claims made in the abstract and introduction accurately reflect the paper's contributions and scope?
    \answerYes{We have focused our experiments on implementing a two-stage model that has a texture-based foveation transform and compared it to a reference model (a perceptual upper bound), and two matched resource systems: one foveated with blur and another one uniformly blurred. }
  \item Did you describe the limitations of your work?
    \answerYes{At the end of each Experiments Sub-Section we provide a mini-discussion of our work and how it fits or does not fit the literature. Mainly we provide limitations in the Discussion at the end (See Section~\ref{sec:Discussion})}
  \item Did you discuss any potential negative societal impacts of your work?
    \answerNo{To our knowledge, there are none.}
  \item Have you read the ethics review guidelines and ensured that your paper conforms to them?
    \answerYes{}
\end{enumerate}

\item If you are including theoretical results...
\begin{enumerate}
  \item Did you state the full set of assumptions of all theoretical results?
    \answerYes{We include only one supplementary theoretical result and proof in the Appendix\ref{sec:Appendix_Reference_Bound}}
	\item Did you include complete proofs of all theoretical results?
    \answerYes{See above.}
\end{enumerate}

\item If you ran experiments...
\begin{enumerate}
  \item Did you include the code, data, and instructions needed to reproduce the main experimental results (either in the supplemental material or as a URL)?
    \answerYes{See Supplementary Material (that provides access to a URL)}
  \item Did you specify all the training details (e.g., data splits, hyperparameters, how they were chosen)?
    \answerYes{These are reported brielfy in Section~\ref{sec:Experiments}, and in more detail through-out the Appendix.}
	\item Did you report error bars (e.g., with respect to the random seed after running experiments multiple times)?
    \answerYes{All experiments were ran with paired initial noise seeds to control for matched initial conditions derived from SGD (though the order in which the networks were exposed to images was different). All errorbars report 1 standard deviation, and these can be seen throughout Sections~\ref{sec:Frequency},\ref{sec:Occlusion},\ref{sec:Window}}
	\item Did you include the total amount of compute and the type of resources used (e.g., type of GPUs, internal cluster, or cloud provider)?
    \answerYes{These are specified in the Appendix.}
\end{enumerate}

\item If you are using existing assets (e.g., code, data, models) or curating/releasing new assets...
\begin{enumerate}
  \item If your work uses existing assets, did you cite the creators?
    \answerYes{We use a re-partition of the Places2 Dataset which is cited.}
  \item Did you mention the license of the assets?
    \answerNo{Given that to our knowledge the Places2 dataset is widely known and free to use.}
  \item Did you include any new assets either in the supplemental material or as a URL?
    \answerNo{As everything in the Supplementary Material/URL has been created/derived by us.}
  \item Did you discuss whether and how consent was obtained from people whose data you're using/curating?
    \answerNA{We did not run any experiments with humans.}
  \item Did you discuss whether the data you are using/curating contains personally identifiable information or offensive content?
    \answerNA{We did not run any experiments with humans, and the scene classes we used were all publicly known and non-offensive places: \textit{e.g.} ocean.}
\end{enumerate}

\item If you used crowdsourcing or conducted research with human subjects...
\begin{enumerate}
  \item Did you include the full text of instructions given to participants and screenshots, if applicable?
    \answerNA{No human subjects were used.}
  \item Did you describe any potential participant risks, with links to Institutional Review Board (IRB) approvals, if applicable?
    \answerNA{No human subjects were used.}
  \item Did you include the estimated hourly wage paid to participants and the total amount spent on participant compensation?
    \answerNA{No human subjects were used.}
\end{enumerate}

\end{enumerate}

\newpage

\appendix

\begin{figure}[!t]
\centering{\centering
\includegraphics[width=1.0\columnwidth,clip=true,draft=false,]{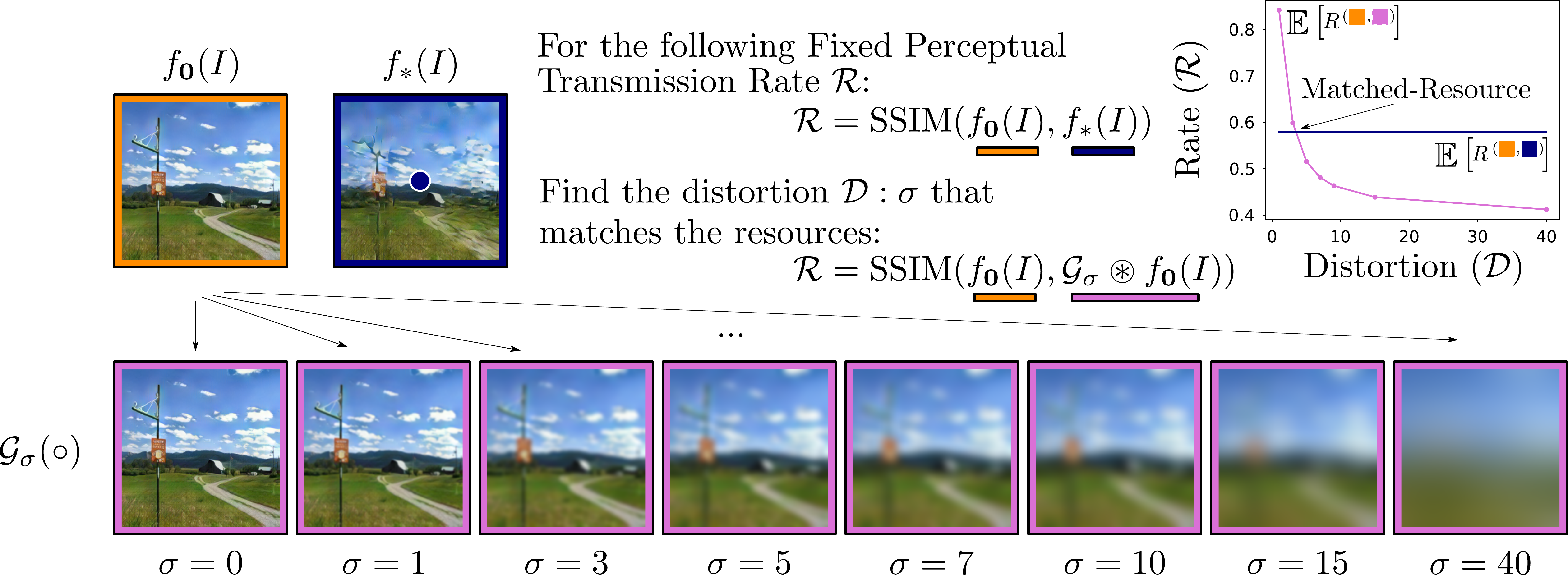}\caption{A full explanatory diagram of the Rate-Distortion Optimization Procedure inspired from both~\citet{balle2016end}~and~\citet{deza2018towards}. The goal is to find the equivalent `perceptual transmission rate' for a given distortion $\sigma$ to find a matched-resource perceptual input for Foveation-Texture that is non-foveated. This optimization pipeline produces Uniform-Blur, a perceptual system that receives as input uniformly blurred images as a way to loosely mimic uniform retinal ganglion cell re-distribution in as if it were to occur in humans. We now have a proper control to evaluate how a foveated texture based model (Foveation-Texture) compares to a non-foveated model (Uniform-Blur) when restricted with the \textit{same} amount of perceptual resources under the aggregate SSIM matching constraint.}}
\label{fig:Rate_Distortion_Uniform}
\end{figure}

\section{Description of All Perceptual Systems}
\label{sec:Supplementary_Foveated}

\textbf{Foveation-Texture}: We adjusted the parameters of the foveation texture transform to have stronger distortions in the periphery that can consequently amplify the differences between a foveated and non-foveated system. This was done setting the rate of growth of the receptive field size (scaling factor) $s=0.4$. 

This value $(s=0.4)$ was used instead of $s=0.5$, given that experiments of~\citet{freeman2011metamers,deza2018towards} have shown that this scaling factor yields a match with physiology but only when human observers are psychophysically tested \textit{between} pairs of synthesized/rendered image metamers. Works of~\citet{wallis_parametric_2017,wallis2019image,deza2018towards,shumikhin2020quantitative} have suggested that the when comparing a non-foveated \textit{reference image} to it's foveated texturized version, the scaling factor is actually much smaller than $0.5$ (0.24, or in some cases as small as 0.20; See Table~\ref{table:Metamer_Comparison}). We thus selected a smaller factor of $s=0.4$ (that is still metameric to a human observer between synthesized pairs), as smaller scaling factors significantly reduced the crowding effects. Ultimately, the selection of this value is not critical in our studies as: 1) we are interested in grossly exagerrating the distortions beyond the human metameric boundary to test if the perceptual system will learn something new or different from the highly manipulated images that use a new family of transformations; 2) we are not making any comparative measurements to human psychophysical experiments where matching such scaling factors would be critical \textit{e.g.}~\citet{deza2016can,eckstein2017humans,geirhos2018generalisation}.

\vspace*{12pt}
\textbf{Reference}:  We use the same image transform at the foveation stage for Reference but set the scaling factor set to $s=0$. In this way, any potential effects of the compression/expansion operations of the image transform stage in the perceptual system is tightly upper-bounded by Reference over Foveation-Texture. Thus, the only difference after stage 1 is whether the image statistics were texturized in increasingly large pooling windows (Foveation-Texture), or not (Reference) -- however note that the texturization procedure comes at a computational cost and modifies the amount of resources allocated in the image. 

Indeed, the Reference system does not provide a matched-resource non-foveated control -- the Reference model only provides a non-foveated \textit{upper bound} that removes the effects of crowding that Foveation-Texture does have (See Theorem~\ref{theo:Standard_Bound}). In fact, the matched-resource control -- under certain constraints (See Table~\ref{table:IQA_Comparison}) -- that is also non-foveated is the Uniform-Blur system as described earlier in the paper, and in more detail as follows.

\textbf{Uniform-Blur}: Uniform-Blur provides a non-foveated resource matched control with respect to Foveation-Texture. This perceptual system is essentially computed via finding the optimal standard deviation $\sigma$ of the Gaussian filtering kernel $\mathcal{G}_\sigma$ as shown in Figure~\ref{fig:Rate_Distortion_Uniform}.
This distortion image is computed via the convolution $(\circledast)$ of the Gaussian filter $\mathcal{G}_\sigma$ with the image $f_\mathbf{0}(I)$. Here,~\citet{wang2004image}'s SSIM is our candidate perceptual metric as it will take into consideration the luminance, contrast and structural changes locally for the entire image and pool them together for an aggregate perceptual score (and also the rate $\mathcal{R}$) that is upper bounded by 1 and correlated with human perceptual judgments. As SSIM operates on the luminance of the image, all validation images over which the RD curve (right) was computed were transformed to grayscale to find the optimal  standard deviation ($\sigma=3.4737$). 

It is also worth emphasizing that the previous matching procedure is done over an aggregate family of images in the validation set (hence the use of the expected value $(\mathbb{E}[\circ])$ in Figure~\ref{fig:Rate_Distortion_Uniform}). This gives us a single standard deviation that will be used to filter \textit{all} the images corresponding to the Uniform-Blur transform the same way.

\begin{figure}[!t]
\centering
{\centering\includegraphics[width=1.0\columnwidth,clip=true,draft=false,]{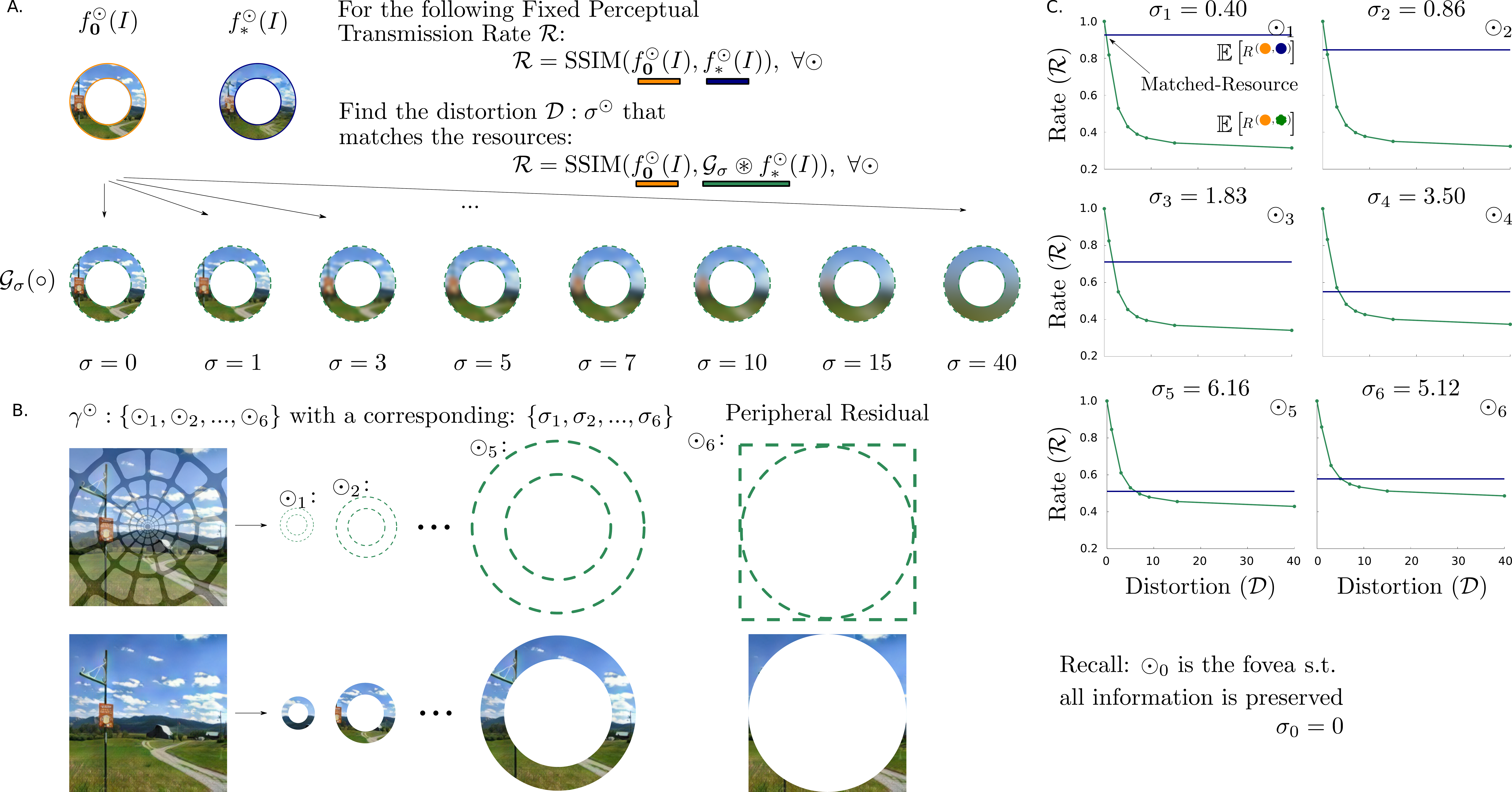}
\caption{\underline{A.} The full explanatory diagram of the Rate-Distortion Optimization Procedure adapted for Foveation-Blur. \underline{B.} The goal is to find the equivalent `perceptual transmission rate' for a given distortion $\sigma$ to find a matched-resource perceptual input for Foveation-Texture that is foveated but with adaptive Gaussian blurring, \textit{i.e.} we must find the standard deviation of the Gaussian blurring kernel which is computed over a set of eccentricity rings that have been windowed with cosine functions. \underline{C.} The full Rate-Distortion curves as a function of retinal eccentricity rings.}
\label{fig:Rate_Distortion_Full_Cartoon_Matched_Ada_Gauss}
}
\end{figure}

\textbf{Foveation-Blur}: Is a foveated perceptual system that receives Rate-Distortion optimized images that have been blurred with different standard deviations of the gaussian kernel $\mathcal{G}_\sigma$ as a function of retinal eccentricity. We picked the same eccentricity rings (collection of pooling regions that lie along the same retinal eccentricity) as Foveation-Texture given that we did not want to include a potential effect that is driven by differences in receptive field sizes rather than differences in type of computation. Figure~\ref{fig:Rate_Distortion_Full_Cartoon_Matched_Ada_Gauss} shows the full set of distortion strengths $(\sigma)$ of each receptive field ring to match the perceptual transmission rate of the Foveation Texture Transform $(f_\mathbf{*}(\circ))$.

There are other alternatives to potentially find the set of standard deviation coefficients that are not driven by a rate-distortion optimization procedure. One possibility could have been to find a mapping between pixels and degrees of visual angle as done in~\citet{pramod2018peripheral} and derive the coefficients by fitting a contrast sensitivity function given the visual field. While this approach is appealing, the coefficients for object recognition such as in ImageNet~\cite{russakovsky2015imagenet} can not be extended to scenes such as Places~\cite{zhou2017places}. In addition, the coupling of the RD-optimization with SSIM provides a perceptual guarantee to compare Foveation-Blur-Net to either Foveation-Texture or Uniform-Blur.

\section{Reference as a perceptual Upper Bound}
\label{sec:Appendix_Reference_Bound}

\vspace{8pt}
\hrule
\begin{theorem}
\label{theo:Standard_Bound}
Reference is a perceptual upper bound, and it's generalization performance can be matched, but can not be exceeded (due to possession of maximum image information).
\end{theorem}
\vspace{-10pt}
\begin{proof}
Let $I'=\mathcal{D}(M)$ be the decoded image to be received by the second stage $g(\circ)$ of any perceptual system, where $M_{\theta_i,\psi_i}=\alpha_i Q_{\theta_i,\psi_i}+(1-\alpha_{i,j})T_{\theta_i,\psi_i}$ is the convex combination between structure and texture for the collection of pooling regions $i$ (Figure~\ref{fig:Methods_Figure} B.). It can be observed that for Reference the values of $\alpha$ yield $\alpha_i=0,\forall i$, thus any other system that has at least 1 value of $\alpha_i\neq 0$ will render a decoded image with a non-zero distortion in pixel space, thus making the resources (amount of information) of Reference greater or equal than any other system with non-zero coefficients (\textit{e.g.} Foveation-Texture).
\end{proof}
\begin{remark}
An example of a theoretically matched generalization performance system to Reference from another non-zero distortion network is possible if the family of pre-distorted images were based on textures (also see~\cite{gatys2015texture} Figure 5).
\end{remark}

\begin{remark}
The resulting transformed images from $f_\mathbf{0}(\circ)$ and $f_\mathbf{*}(\circ)$ are not diffeomorphic to each other. 
\end{remark}
\hrule

\section{Full set of IQA Metrics}
\label{sec:IQA_Supplement}

\begin{table}[h]
\centering
\small
{
\centering
\begin{adjustwidth}{-1.0in}{-.5in}  
\begin{tabular}{|c|c|c|c|}
\hline
 (mean$\pm$std) & SSIM \textit{(Matched)} & MSE $(\uparrow)$ & Mutual Information $(\downarrow)$ \\
 \hline
Reference & 1.0 & 0.0 & $7.39\pm0.52$ \\
Foveation-Texture & $\mathbf{0.58\pm0.11}$ & $\mathbf{976.78\pm522.22}$ & $\mathbf{1.40\pm0.42}$  \\
Uniform-Blur & $\mathbf{0.57\pm0.15}$ & $458.67\pm277.13$ & $1.86\pm0.58$ \\
Foveation-Blur & $\mathbf{0.58\pm0.15}$ & $507.35\pm302.71$ & $1.84\pm0.56$ \\
\hhline{|=|=|=|=|}
 (mean$\pm$std) & MS-SSIM~\citep{wang2003multiscale}$(\downarrow)$ & CW-SSIM~\citep{wang2005translation} $(\downarrow)$ & FSIM~\citep{zhang2011fsim}$(\downarrow)$ \\
 \hline
Reference & 1.0 & 1.0 & 1.0 \\
Foveation-Texture & $\mathbf{0.20\pm0.03}$ & $\mathbf{0.74\pm0.05}$ & $\mathbf{0.76\pm0.05}$  \\
Uniform-Blur & $0.36\pm0.03$ & $0.98\pm0.01$ & $\mathbf{0.69\pm0.09}$ \\
Foveation-Blur & $0.36\pm0.03$ & $0.98\pm0.01$ & $\mathbf{0.67\pm0.10}$ \\
\hhline{|=|=|=|=|}
 (mean$\pm$std) & VSI~\citep{zhang2014vsi} $(\downarrow)$ & GMSD~\citep{xue2013gradient} $(\uparrow)$ & NLPD~\citep{laparra2016perceptual} $(\uparrow)$\\
 \hline
Reference & 1.0 & 0.0 & 0.0 \\
Foveation-Texture & $\mathbf{0.93\pm0.02}$ & $\mathbf{0.19\pm0.03}$ & $\mathbf{0.75\pm0.16}$ \\
Uniform-Blur & $\mathbf{0.91\pm0.04}$ & $\mathbf{0.19\pm0.03}$ & $0.40\pm0.09$ \\
Foveation-Blur & $\mathbf{0.91\pm0.04}$ & $\mathbf{0.22\pm0.04}$ & $0.45\pm0.11$\\
\hhline{|=|=|=|=|}
 (mean$\pm$std) & MAD~\citep{larson2010most} * $(\uparrow)$ & VIF~\citep{sheikh2006image} $(\downarrow)$ & LPIPSvgg~\citep{zhang2018unreasonable} * $(\uparrow)$ \\
 \hline
Reference & 0.0 & 1.0 & 0.0 \\
Foveation-Texture & $166.77\pm19.46$ & $\mathbf{0.12\pm0.03}$ & $0.35\pm0.05$ \\
Uniform-Blur & $\mathbf{182.19\pm16.50}$ & $\mathbf{0.12\pm0.03}$ & $\mathbf{0.52\pm0.07}$ \\
Foveation-Blur & $\mathbf{185.90\pm18.60}$ & $\mathbf{0.16\pm0.03}$ & $\mathbf{0.54\pm0.08}$ \\
\hhline{|=|=|=|=|}
 (mean$\pm$std) & DISTS~\citep{ding2020image} * $(\uparrow)$ & & \\
 \hline
Reference & 0.0 &  &  \\
Foveation-Texture & $0.20\pm0.03$ & &\\
Uniform-Blur & $\mathbf{0.36\pm0.03}$ & & \\
Foveation-Blur & $\mathbf{0.35\pm0.03}$ & & \\
\hhline{|=|=|=|=|}
\end{tabular}
\end{adjustwidth}
\vspace{5pt}
\caption{List of Full IQA Metrics from~\cite{2020arXiv200501338D} where we compare Image Transforms $f(\circ)$ w.r.t. Reference for the \textit{testing} set. Arrows ($\uparrow/\downarrow$) indicate the direction of the \textit{greatest} distortion according to the metric thus values further away from the Reference place a specific transform at a resource disadvantage. We observe matched distortion via virtual ties for SSIM (matched and optimized in the \textit{validation} set), VSI, GMSD FSIM, and VIF; greater distortion (Foveation-Texture at a disadvantage) for MSE, Mutual Information, MS-SSIM, CW-SSIM, NLPD; and lower distortion (Foveation-Texture at an advantage) for MAD, and texture-based tolerance methods such as DISTS and LPIPSvgg -- hence implicitly proving that our transform does indeed preserve local texture. Scores were computed over 5000 images. Numbers in bold represent highest/lowest IQA scores; virtual ties were declared if highly overlapping standard deviations are noticeable \textit{e.g.}: FSIM, VIF.}
\label{table:IQA_Comparison}
}
\end{table}

\newpage
\clearpage

\section{Image Transform Samples}

\begin{figure}[!t]
\centering\includegraphics[width=1.0\columnwidth,clip=true,draft=false,]{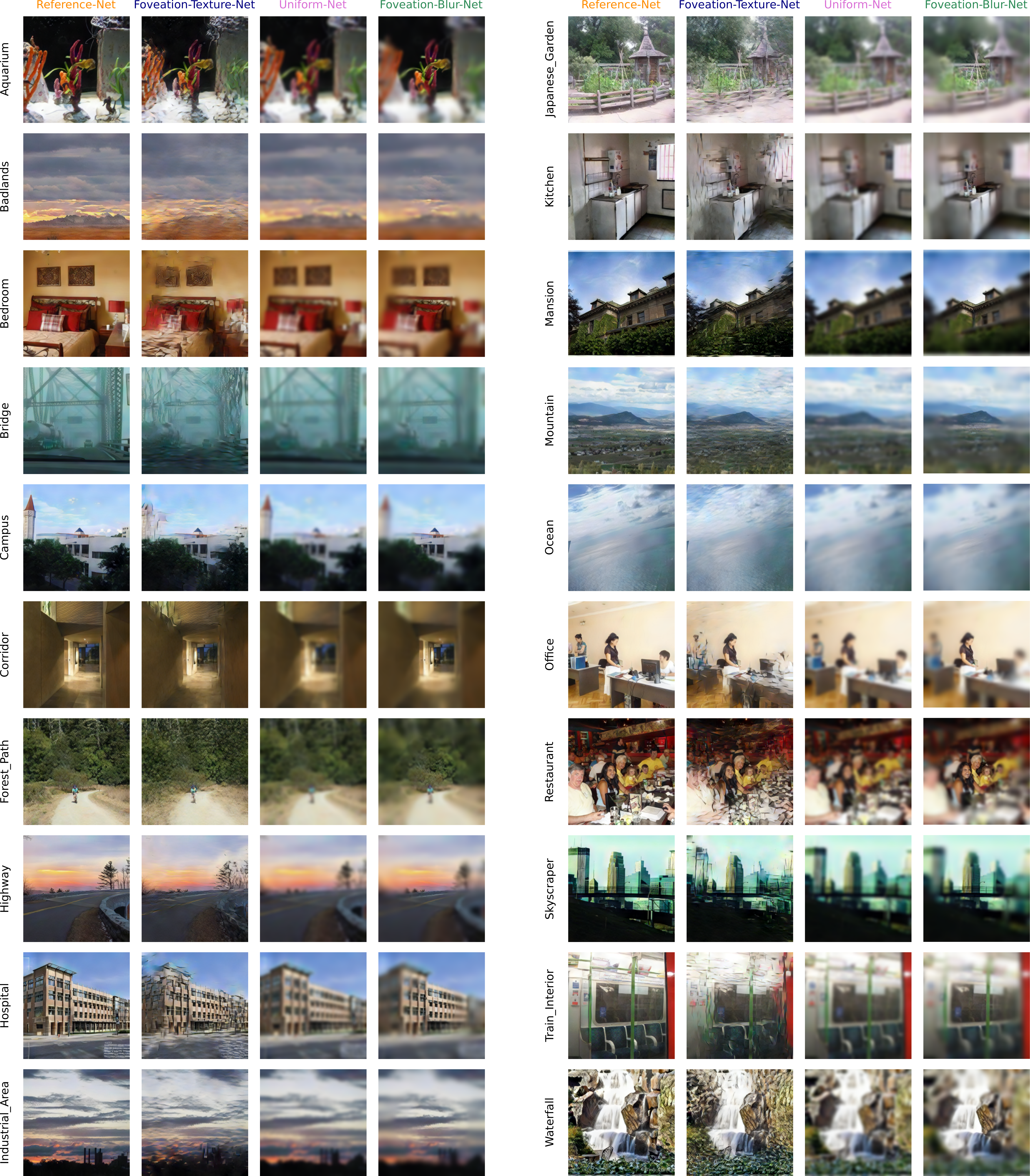}\caption{Sample Testing Image Mosaics.}
\label{fig:Sample_Testing_Images}
\end{figure}

Figure~\ref{fig:Sample_Testing_Images} is an extension of Figure~\ref{fig:Samples_and_Methods} which shows a collection of randomly sampled images from each one of the 20 scene classes and how they look under each image transform before being fed to each network. Details worth noticing include: 1) Reference images are not full high resolution, and are slightly compressed given the encoder/decoder pipeline of the transform to operate as a tighter upper bound (observable when zooming in); 2) The foveal area is preserved and \textit{identical} for Reference, Foveation-Texture and Foveation-Blur; 3) The peripheral distortions are more or less apparent contingent on the image structure; 4) All images used in our experiments were rendered at $256\times256$ px.

\label{sec:Visualization}

\newpage
\clearpage

\begin{table}[!t]
\centering
\scriptsize
\centering
\begin{adjustwidth}{-0.65in}{-.5in}  
\begin{tabular}{|c|c|c|c|c|}
 \hline
Model & \cite{freeman2011metamers} & \cite{wallis2019image} & \cite{fridman2017sideeye} & \cite{deza2018towards} \\
 \hline
Feed-Forward       & - & - & \checkmark & \checkmark  \\
Input & Noise & Noise & Image & Image \\
Multi-Resolution & \checkmark & \checkmark & - & - \\
Texture Statistics & Steerable Pyramid & VGG19 \textit{conv-}$1_1,2_1,3_1,4_1,5_1$ & Steerable Pyramid & VGG19 \textit{relu}$4_1$ \\
Style Transfer &~\cite{portilla2000parametric} & \cite{gatys2016image} & \cite{rosenholtz2012summary} & \cite{huang2017arbitrary}\\
Foveated Pooling & \checkmark & \checkmark & (Implicit via FCN) & \checkmark \\
Decoder (trained on) & - & - & metamers/mongrels & images \\ 
Moveable Fovea & \checkmark & \checkmark & \checkmark & \checkmark \\
\hline
Use of Noise & Initialization & Initialization & -  & Perturbation \\
Non-Deterministic & \checkmark & \checkmark & -  & \checkmark \\
Direct Computable Inverse & - & - & (Implicit via FCN) & \checkmark \\
\hline
Rendering Time & hours & minutes & miliseconds & seconds \\
\hline
Image type & scenes & scenes/texture & scenes & scenes\\ 
Critical Scaling (\textit{vs} Synth) & 0.46 & $\sim\{0.39/0.41\}$ & Not Required & 0.5 \\
Critical Scaling (\textit{vs} Reference) & Not Available & $\sim\{0.2/0.35\}$ & Not Required & 0.24 \\
Experimental design & ABX & Oddball & - & ABX \\ 
Reference Image in Exp. & Metamer & Original & - & Compressed via Decoder \\
Number of Images tested & 4 & 400 & - & 10 \\
Trials per observers & $\sim 1000$ & $\sim 1000$ & - & $\sim 3000$ \\
\hline
\end{tabular}
\end{adjustwidth}
\caption{Foveated Texture-based transform comparison. Redrawn from~\cite{deza2018towards}.}
\label{table:Metamer_Comparison}
\end{table}

\section{Differences across other Foveation models}
There are currently 4 foveation models that implement texture-like computation in the peripheral field of view as shown in Table~\ref{table:Metamer_Comparison}. We selected the Foveation Texture Transform model of~\cite{deza2018towards} given that it is computationally tractable to render a foveated image dataset (100'000) at a rate of 1 image/second (rather than hours~\cite{freeman2011metamers} or minutes~\cite{wallis_parametric_2017}). We did not use the highly accelerated model of~\citet{fridman2017sideeye} (order of miliseconds, that was based on the Texture-Tiling Model of~\citet{rosenholtz2012summary}) because it was: 1) Not psychophysically tested with human observers thus there is no guarantee of visual metamerism via the choice of texture statistics (although see the recent work of~\citet{shumikhin2020quantitative}); 2) But most importantly, it does not provide an upper-bound computational baseline (similar to Reference). 

Altogether, we think that re-running our experiments and testing them with all other foveated models such as the before-mentioned is a direction of future work as we would be curious to see the replicability of our pattern of results across other texture-based peripheral models. Naturally, the type of texture-based foveation used will also yield different matched resource systems (Uniform-Blur and Foveation-Blur), as different models rely on texture computation in different ways -- and thus will affect the IQA metric scores when performing the perceptual optimization.

\clearpage
\newpage

\begin{table}[!t]
\tiny
\centering
\begin{adjustwidth}{-0.95in}{-.5in}  
\begin{tabular}{|c|c|c|c|c|}
 \hline
Model & \cite{wang2017central}) & \cite{wu2018learning} & \cite{pramod2018peripheral} & (Ours) \\
 \hline
Image input type & scenes & objects & objects & scenes \\
Single/Dual Stream       & Dual + Gating & Dual + Concatenation & Single & Single\\
Role of Single/Dual Stream & Coupling the fovea + periphery & Contextual modulation (scene gist) & \multicolumn{2}{|c|}{Serializing the (single) two-stage model} \\ 
Foveated Transform (F.T.) & log-polar + adaptive gaussian blurring & Region Selection & adaptive gaussian blurring & Visual Metamer w/ texture-distortion  \\
Stochastic F.T. & - & - & - &\checkmark~\cite{deza2018towards} \\
Representational Stage of F.T. & retinal~\citep{geisler1998real} & "Overt Attention" & retinal~\citep{geisler1998real} & V2~\citep{freeman2011metamers}\\
Moveable Fovea & \checkmark & \checkmark & \checkmark & \checkmark \\
\hline
Accounts for pooling regions & Implicit via adaptive gaussian blurring & - & Implicit via adaptive gaussian blurring & \checkmark\\
Accounts for visual crowding & - & - & - & \checkmark\\
Accounts for retinal eccentricity & \checkmark & Implicit via cropping & \checkmark & \checkmark \\
Accounts for loss of visual acuity & \checkmark & - & \checkmark & Implicit via visual crowding\\
\hline
Critical Radius~\citep{larson2009contributions} & $8\deg$ & \multicolumn{2}{|c|}{Not Applicable (Objects)} & $\sim8.67\deg$ (Estimated from Fig.~\ref{fig:Window_Cue_Conflict})  \\
\hline
Out of Distribution Generalization & - & - & - & \checkmark \\
Robustness to Distortion Type & - & Blurring & Blurring & Occlusion \\
Spatial Frequency Preference & High (Fovea), Low (Periphery) & Low (Global) & High (Fovea), Low (Periphery) & High (Global) \\ 
Weighted Bias Emerges & Center/Fovea & Center/Fovea & Center/Fovea & Center/Fovea \\
\hline
Goal of Foveal-Peripheral Architecture & Fit Behavioural Results & \multicolumn{2}{|c|}{Increase Recognition Accuracy} & Explore Perceptual Properties \\
Model System Focus & Human & Machine & Human & Hybrid\\
\hline
\end{tabular}
\end{adjustwidth}
\caption{A summary set of Foveal-Peripheral CNN model characteristics.}
\label{table:Foveation_Comparison}
\end{table}

\section{Differences to other Relevant Work}
There are several works that have used foveation to show a type of representational advantage over non-foveated systems. Mainly~\citet{pramod2018peripheral} with adaptive gaussian blur, and~\citet{wu2018learning} with scene gist, that have been targeted towards a computational goal in increasing object recognition performance. For scene recognition, only~\citet{wang2017central} has successfully modelled known behavioural results of~\citet{larson2009contributions} via a dual-stream neural network that uses adaptive gaussian blurring and a log-polar transform. One key difference however is that we are interested in exploring the effects of peripheral texture-base computation that give rise to~\textit{visual crowding} and that is also linked to area V2 in the primate ventral stream  -- rather than retinal as in~\citet{wang2017central} which resembles our control condition: Foveation-Blur. 

In general, we are taking a complimentary approach to \citet{wang2017central}~\&~\citet{wu2018learning}, and a similar one to~\citet{pramod2018peripheral} where we \textit{a priori do not know of a functional role of texture-based computation or prime ourselves to fit our model to a reference behavioural result}. Thus we explore what perceptual properties it may have in comparison to a non-foveated system (Uniform-Blur, Reference) or a foveated system that only implements adaptive gaussian blurring (Foveation-Blur). Table~\ref{table:Foveation_Comparison} highlights key similarities~\&~differences between these papers and ours.

\newpage
\clearpage

\section{Training, Testing and Learning Dynamics}
\label{sec:Appendix_Train_Test_Dyn}

\begin{figure*}[!h]
\centering
\includegraphics[width=1.0\columnwidth,clip=false,draft=false,]{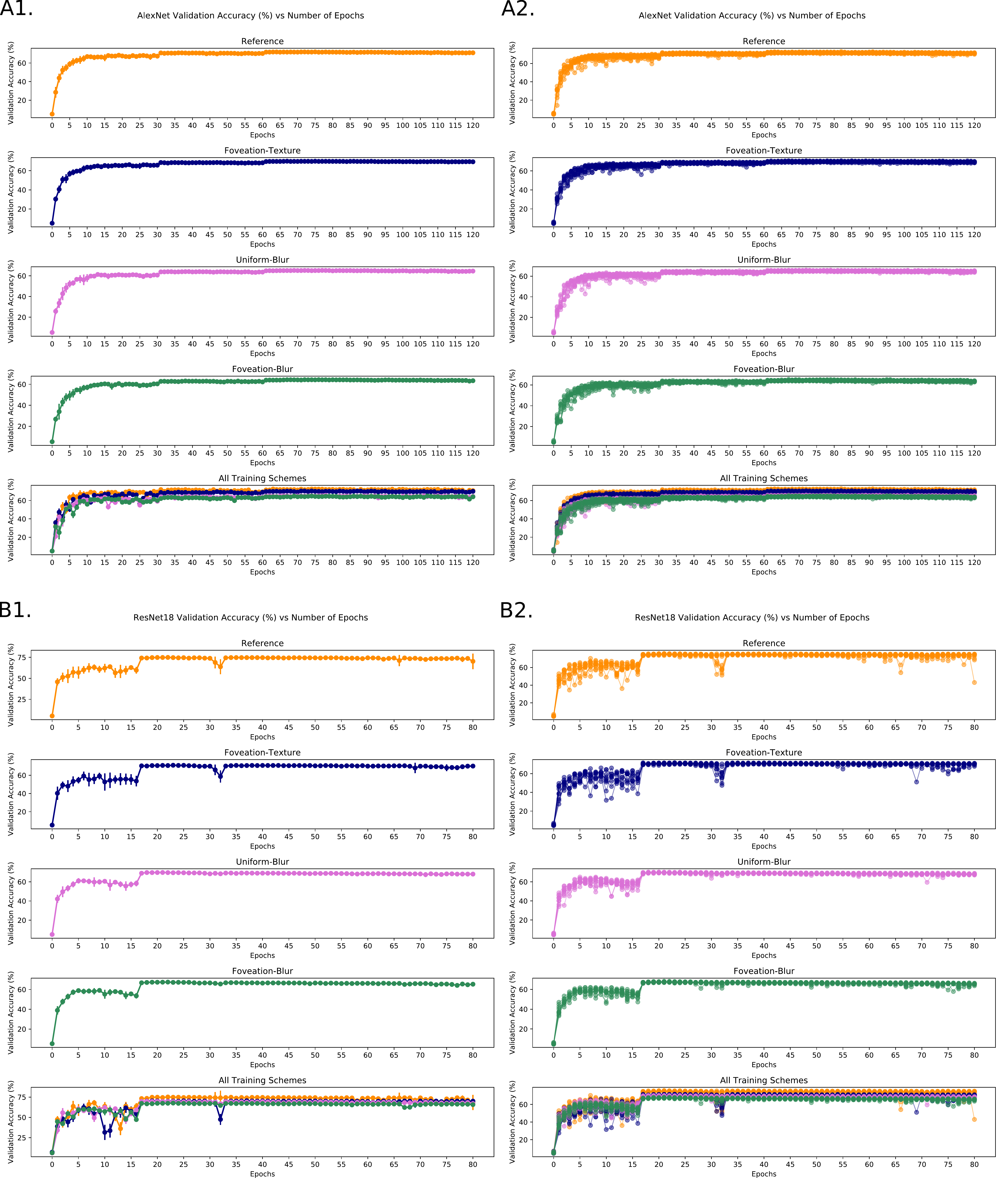}\caption{Learning Dynamics visualized via the Validation Accuracy over all epochs for AlexNet and ResNet18 as $g(\circ)$. Left: A1/B1 we see the aggregate Validation Accuracy. Right: A2/B2 the individual Validation Accuracies are shown for each network.}\label{fig:Validation_Accuracy}
\end{figure*}

\newpage
\clearpage

\begin{figure*}[!t]
\centering
\includegraphics[width=1.0\columnwidth,clip=false,draft=false,]{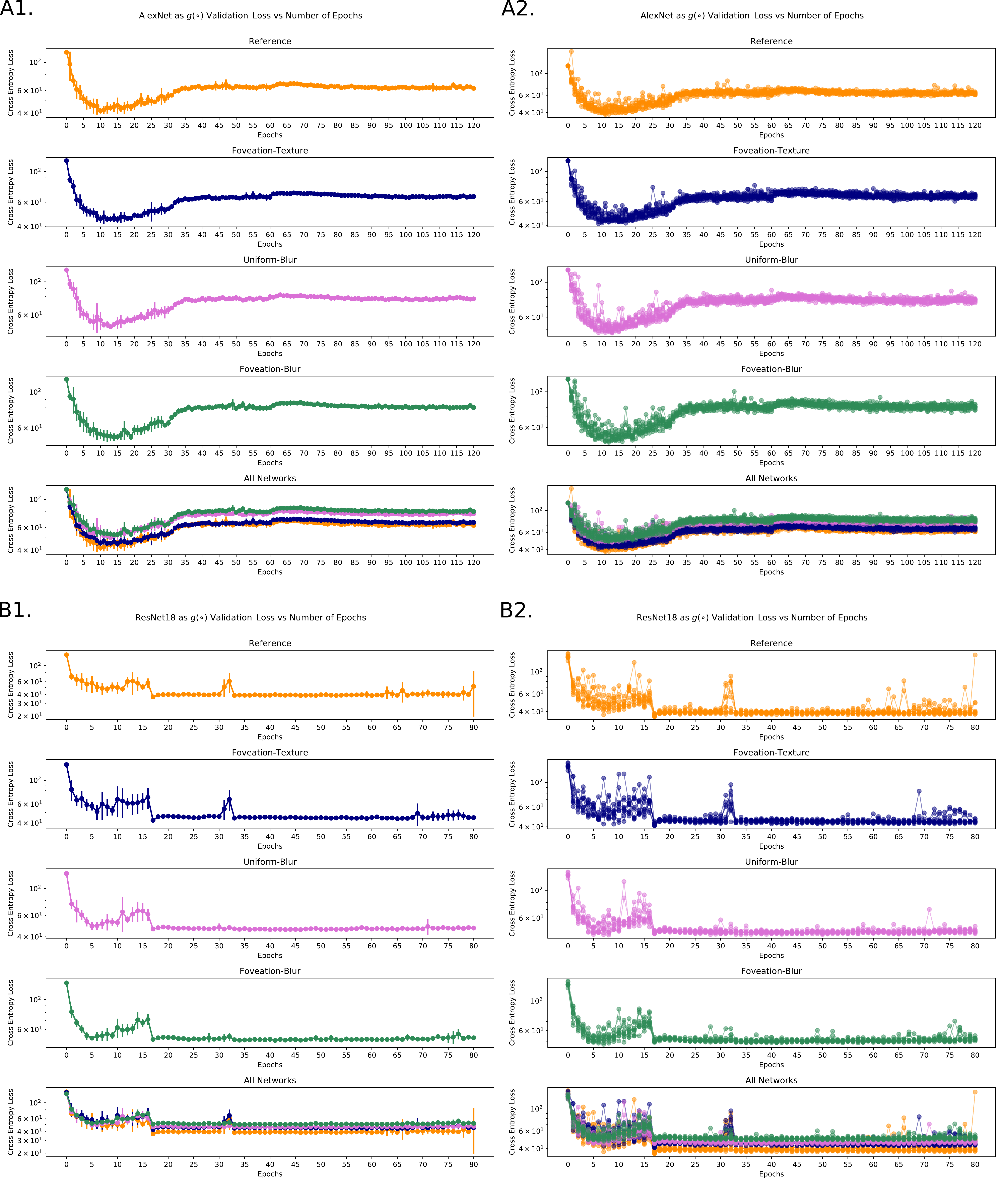}\caption{Validation Loss (Cross Entropy) over all epochs for AlexNet and ResNet18 as $g(\circ)$. Left: A1/B1 we see the aggregate Validation Loss. Right: A2/B2 the individual Validation Losses for each network. It is interesting to see that despite re-bound effects in the validation loss, that the validation \textit{accuracy} continues to increase (See Figure~\ref{fig:Validation_Accuracy}). }\label{fig:Validation_Loss}
\end{figure*}

\newpage
\clearpage

\begin{figure*}[!t]
\centering
\includegraphics[width=1.0\columnwidth,clip=false,draft=false,]{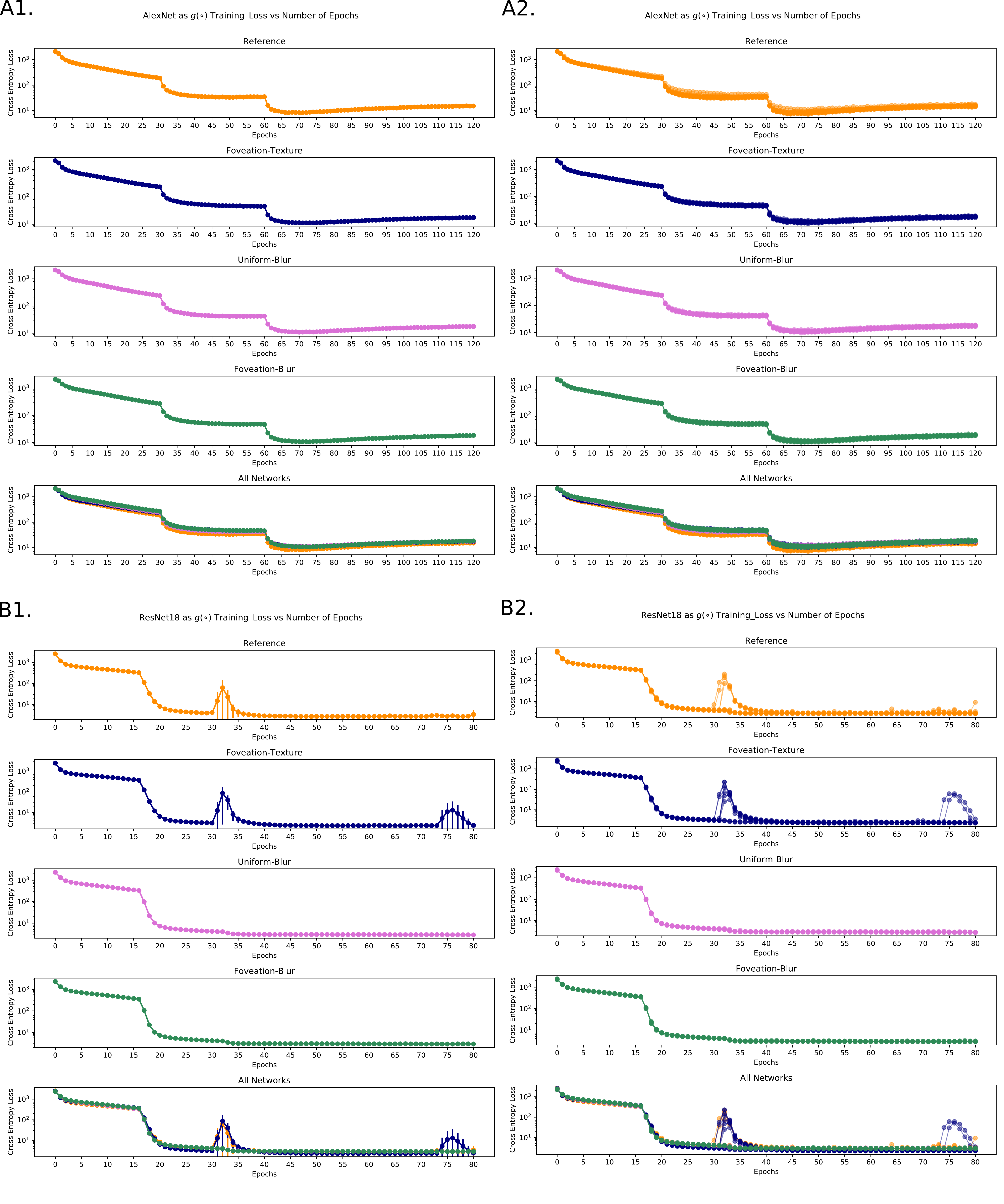}\caption{Training Loss (Cross Entropy) over all epochs for AlexNet and ResNet18 as $g(\circ)$. Left: A1/B1 we see the aggregate training loss. Right: A2/B2 the individual training losses for each network.}\label{fig:Training_Loss}
\end{figure*}

Perceptual Systems were trained with SGD, nestorov momentum, no dampening, weight decay = $0.0005$, momentum = 0.9, a batch size of 128, Color Normalization of mean = $(0.485, 0.456, 0.406)$, and std = $(0.229,0.224,0.225)$. Systems that used AlexNet as $g(\circ)$ were trained for 120 epochs with a scheduled learning rate, where the initial learning rate of $0.01$ was halved after the 30th epoch, and halved again after 60th epoch. Systems that used ResNet18 as $g(\circ)$ were trained for 80 epochs and with an initial learning rate of $0.05$, which was multiplied by $0.25$ after the first 16 epochs, and then multiplied again by $0.25$ after the 32nd epoch. All systems were trained with a cross-entropy loss and received images size of $256\times256\times3$. No data-augmentation or cropping was used at training or testing.

\newpage
\clearpage

\section{Generalization}
\label{sec:Appendix_Generalization}

\begin{figure*}[!h]
\centering
\includegraphics[width=1.0\columnwidth,clip=false,draft=false,]{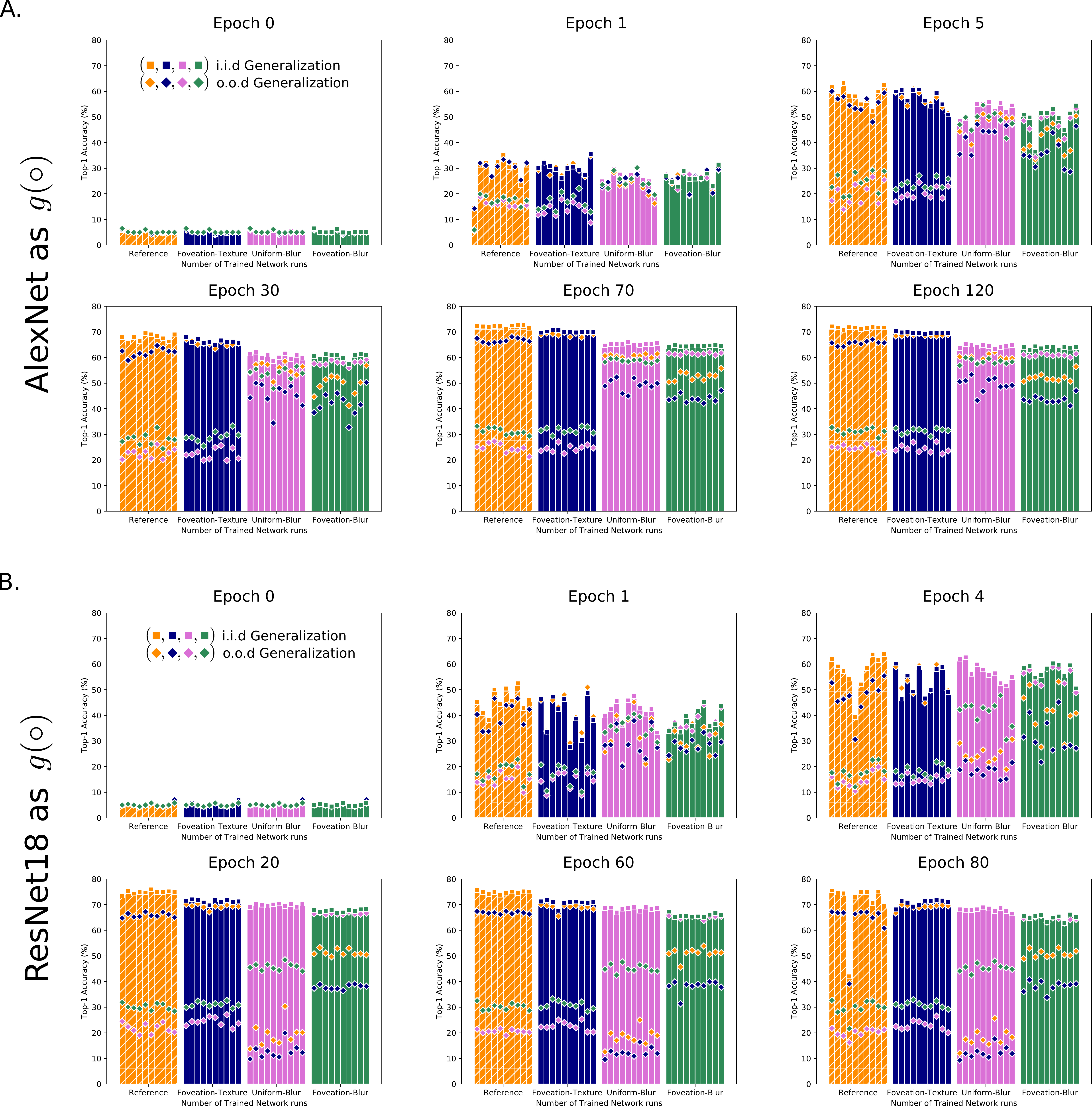}\caption{Generalization Dynamics over a set of multiple epochs for AlexNet and ResNet18 as $g(\circ)$.}\label{fig:Generalization_AlexNet}
\end{figure*}

\begin{figure*}[!h]
\centering
\includegraphics[width=0.9\columnwidth,clip=false,draft=false,]{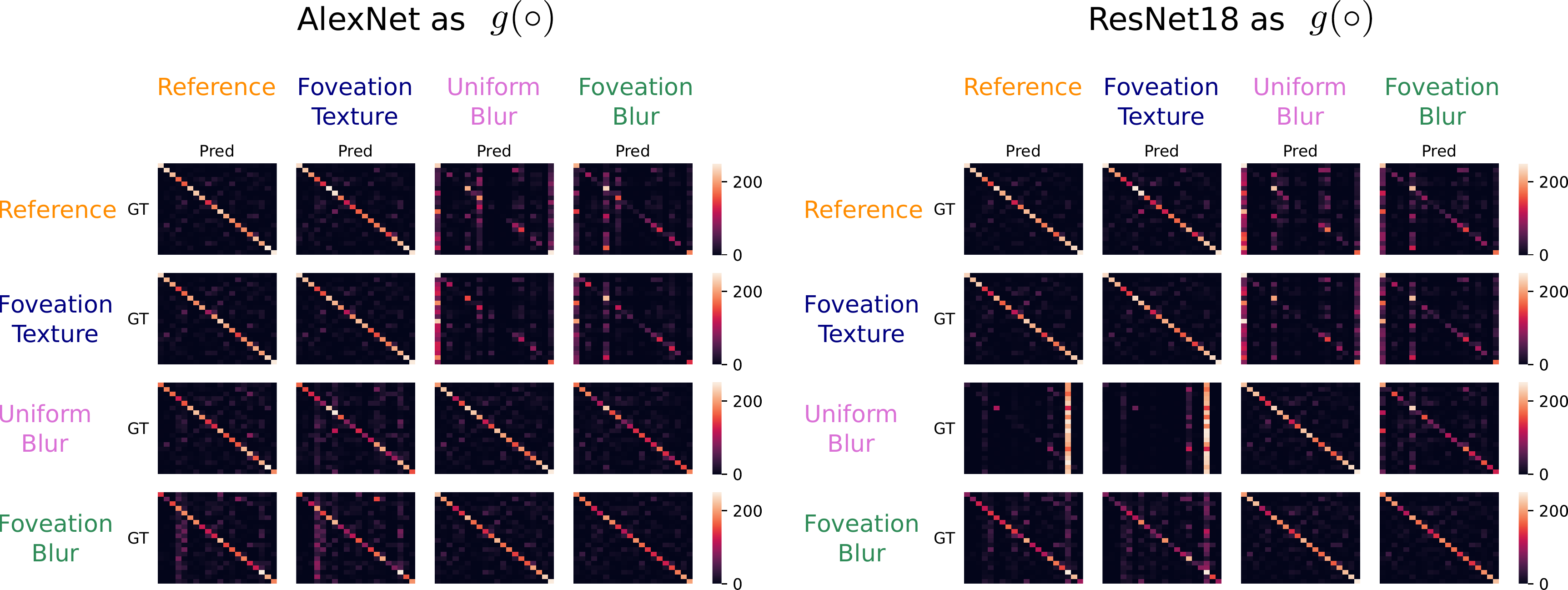}\caption{A sample collection of Confusion Matrices for the first of the 10 randomly initialized networks for each of the 4 perceptual systems with their transforms for both AlexNet and ResNet18 as $g(\circ)$. We see similar classification patterns between Foveation-Texture and the Reference, and also similar classification strategies between the Foveation-Blur and Uniform-Blur system. The asymmetry in the upper and lower off-diagonal quadrants highlight the differences between Foveation-Texture \& Reference \textit{vs} Foveation-Blur \& Uniform-Blur. Each row/column per confusion matrix represents each of the scene classes in alphabetical order. These classes are: aquarium, badlands, bedroom, bridge, campus, corridor, forest path, highway, hospital, industrial area, japanese garden, kitchen, mansion, mountain, ocean, office, restaurant, skyscraper, train interior, waterfall.}\label{fig:Supplement_Confusion_Matrices}
\vspace{-10pt}
\end{figure*}

\newpage
\clearpage

\begin{figure}[!t]
\centering
\includegraphics[width=1.0\columnwidth,clip=true,draft=false,]{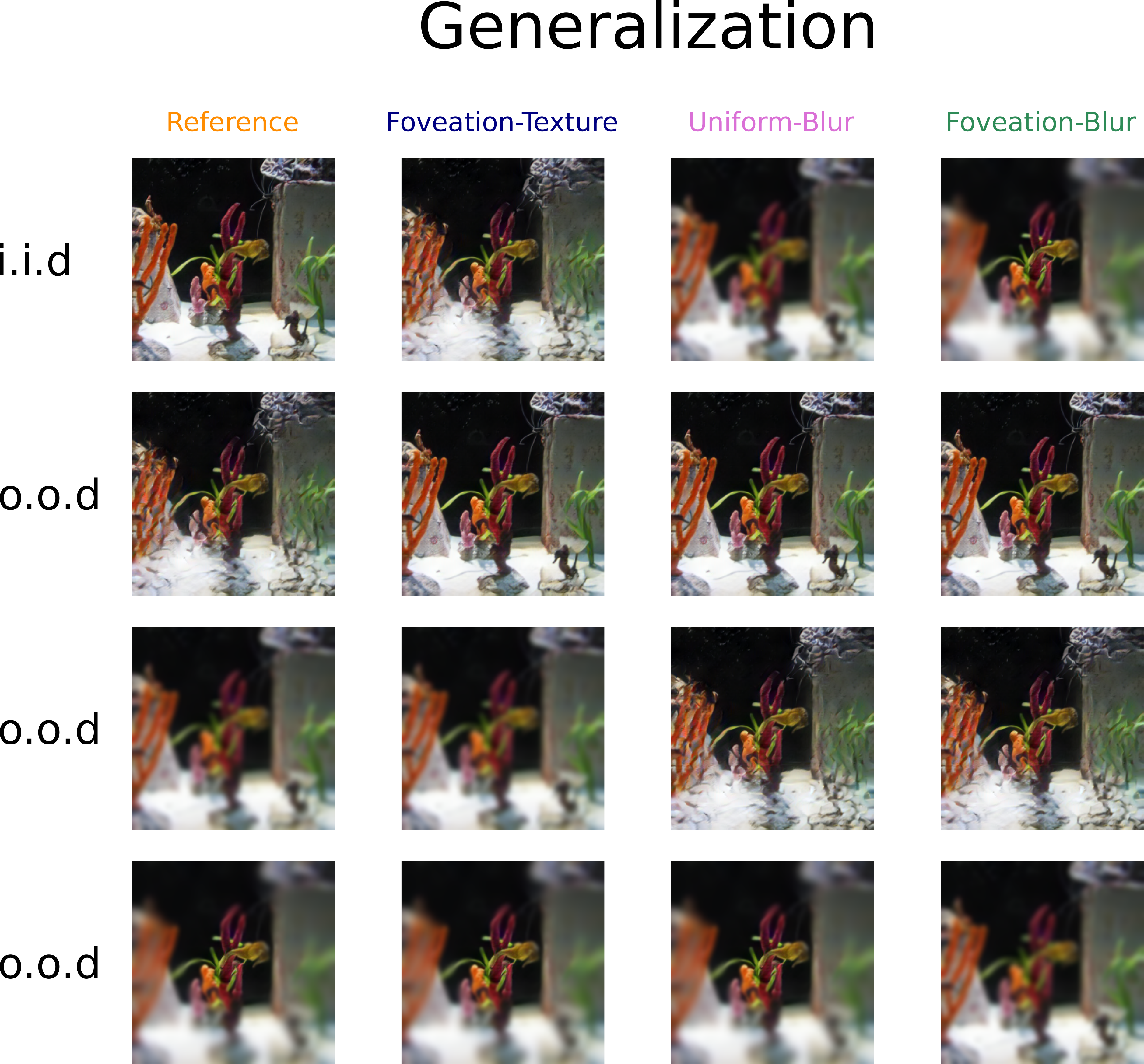}\caption{Sample image used in a full i.i.d and o.o.d evaluation.}\label{fig:Generalization_Schematic}
\end{figure}

\newpage
\clearpage

\section{Filter Visualization \& Spatial Frequency Sensitivity}
\label{sec:Appendix_Spatial_Frequency}

\begin{figure*}[!h]
\centering
\includegraphics[width=1.0\columnwidth,clip=false,draft=false,]{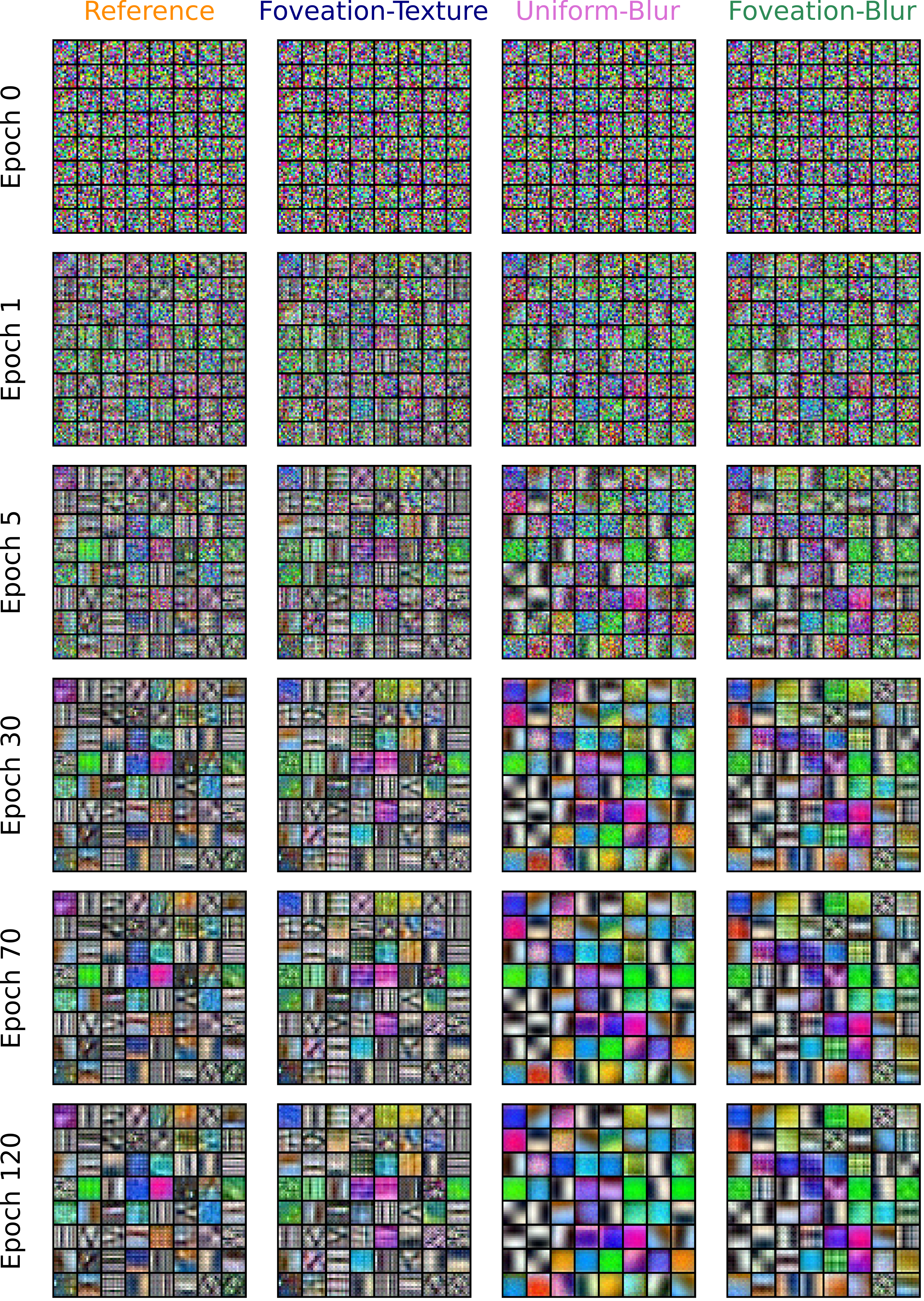}\caption{Evolution of AlexNet as $g(\circ)$ Conv-1 Filters from 1st Random Weight Initialization.}\label{fig:Filters_AlexNet1}
\end{figure*}

\newpage
\clearpage

\begin{figure*}[!t]
\centering
\includegraphics[width=1.0\columnwidth,clip=false,draft=false,]{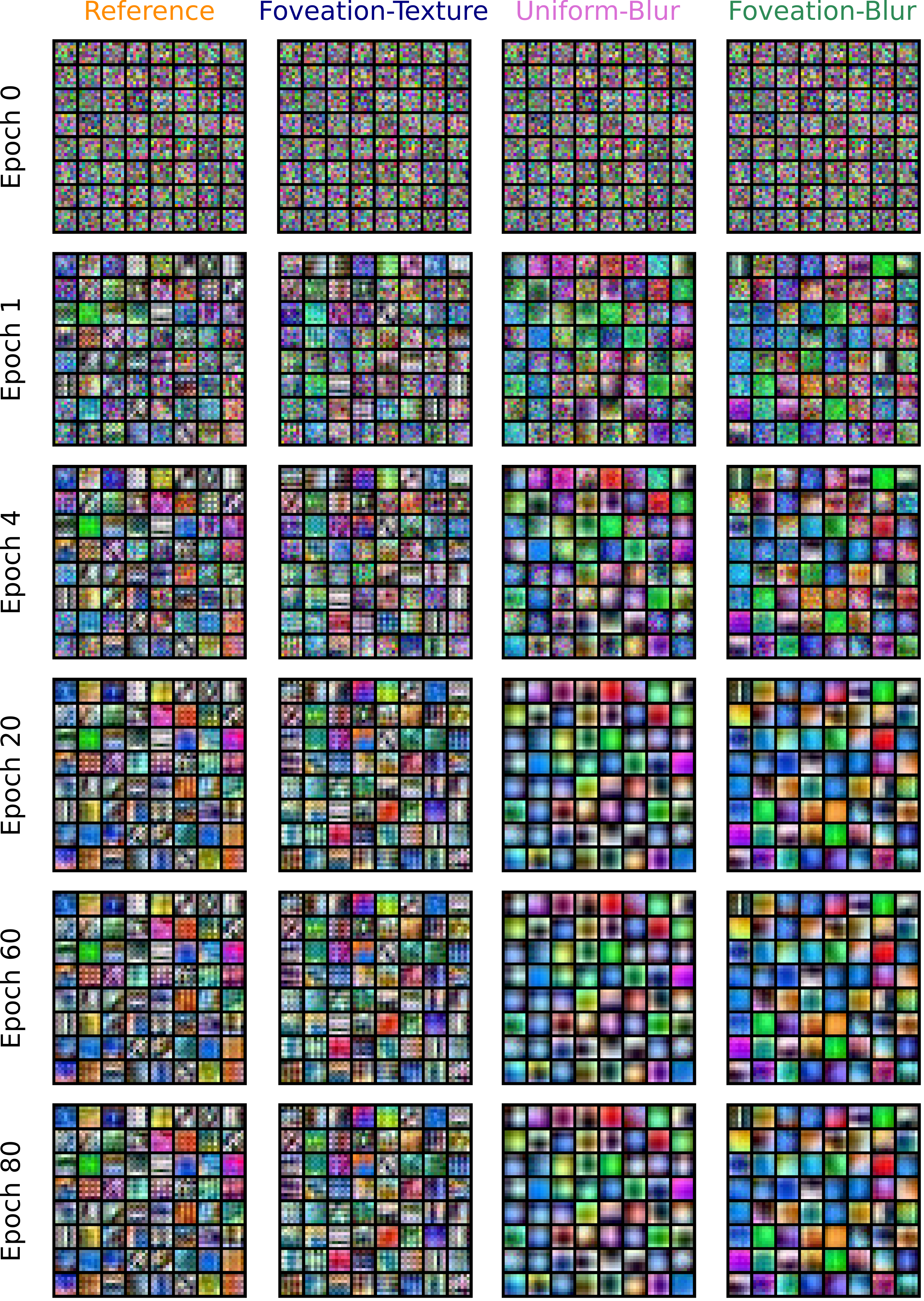}\caption{Evolution of ResNet18 as $g(\circ)$ Conv-1 Filters from 1st Random Weight Initialization.}\label{fig:Filters_ResNet181}
\end{figure*}

\clearpage
\newpage


The size of all shown images was $256\times256\times3$, thus the units of the gaussian filters specified from Section~\ref{sec:Frequency} are in pixels. For a given Gaussian filtering operation $\mathcal{G}_\sigma$ for a given standard deviation $\sigma$, low pass spatial frequency (LF) images were computed via:
\begin{equation}
    LF(I^C) = \mathcal{G}_\sigma \star I^C
\end{equation}
for each channel $C$. Similarly, High Pass Spatial Frequency (HF) image stimuli were computed via:
\begin{equation}
    HF(I^C) = I^C - \mathcal{G}_\sigma \star I^C + \text{mean}_{\text{val}}^C
\end{equation}
where $\text{mean}_{\text{val}}^C$ (which we call the residual in the main body of the paper) is the average of image intensity over the held-out validation set for each channel $C$, a small extension from~\citet{geirhos2018imagenettrained} as our image stimuli is in both color and grayscale. 


\clearpage
\newpage

\begin{figure*}[!h]
\centering
\includegraphics[width=0.8\columnwidth,clip=false,draft=false,]{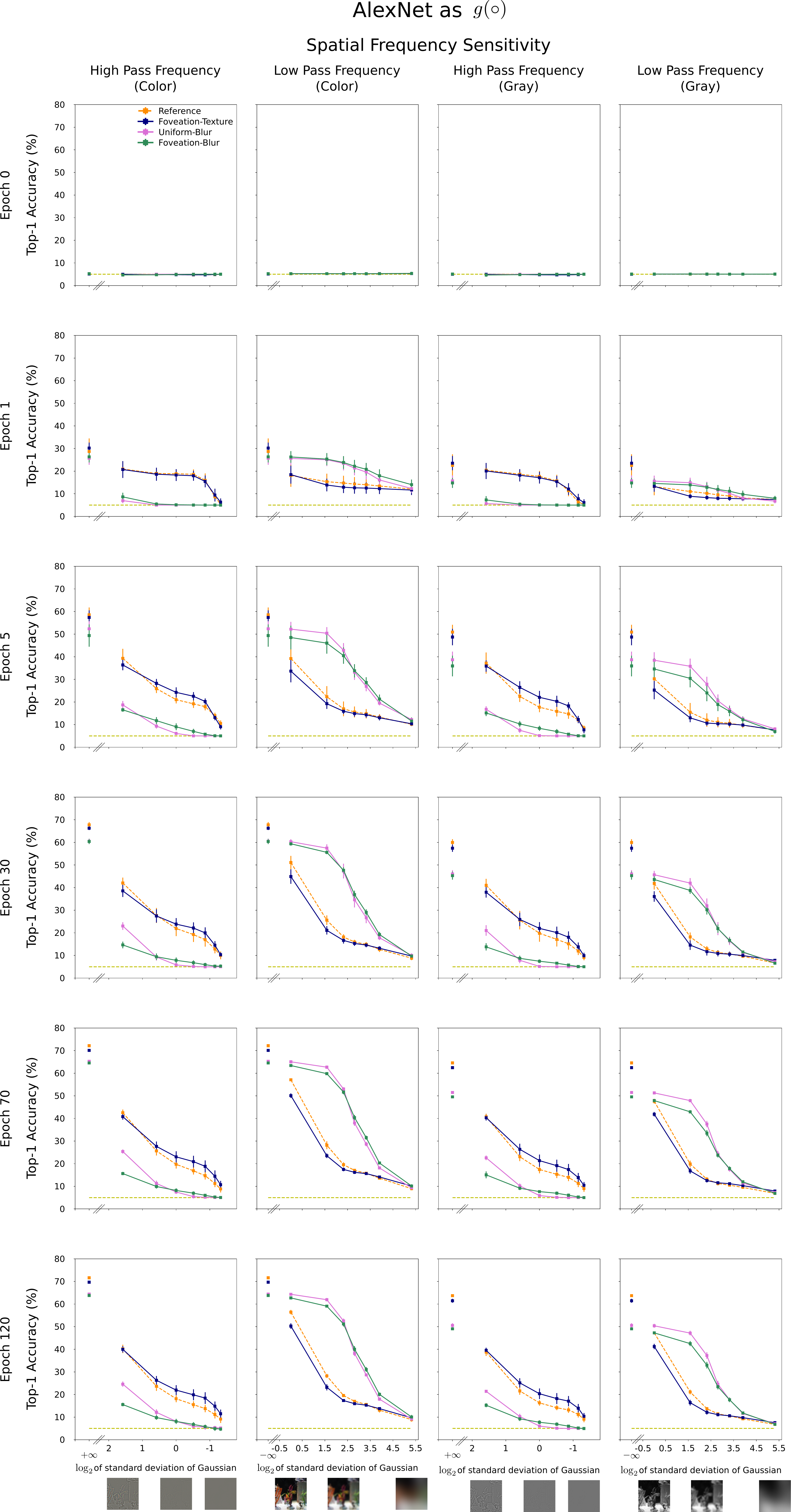}\caption{Aggregate Spatial Frequency Sensitivity for AlexNet as $g(\circ)$ after epochs 0, 1, 5, 30, 70, 120.}\label{fig:Spatial_Frequency_AlexNet_Average}
\end{figure*}

\clearpage
\newpage

\begin{figure*}[!t]
\centering
\includegraphics[width=0.8\columnwidth,clip=false,draft=false,]{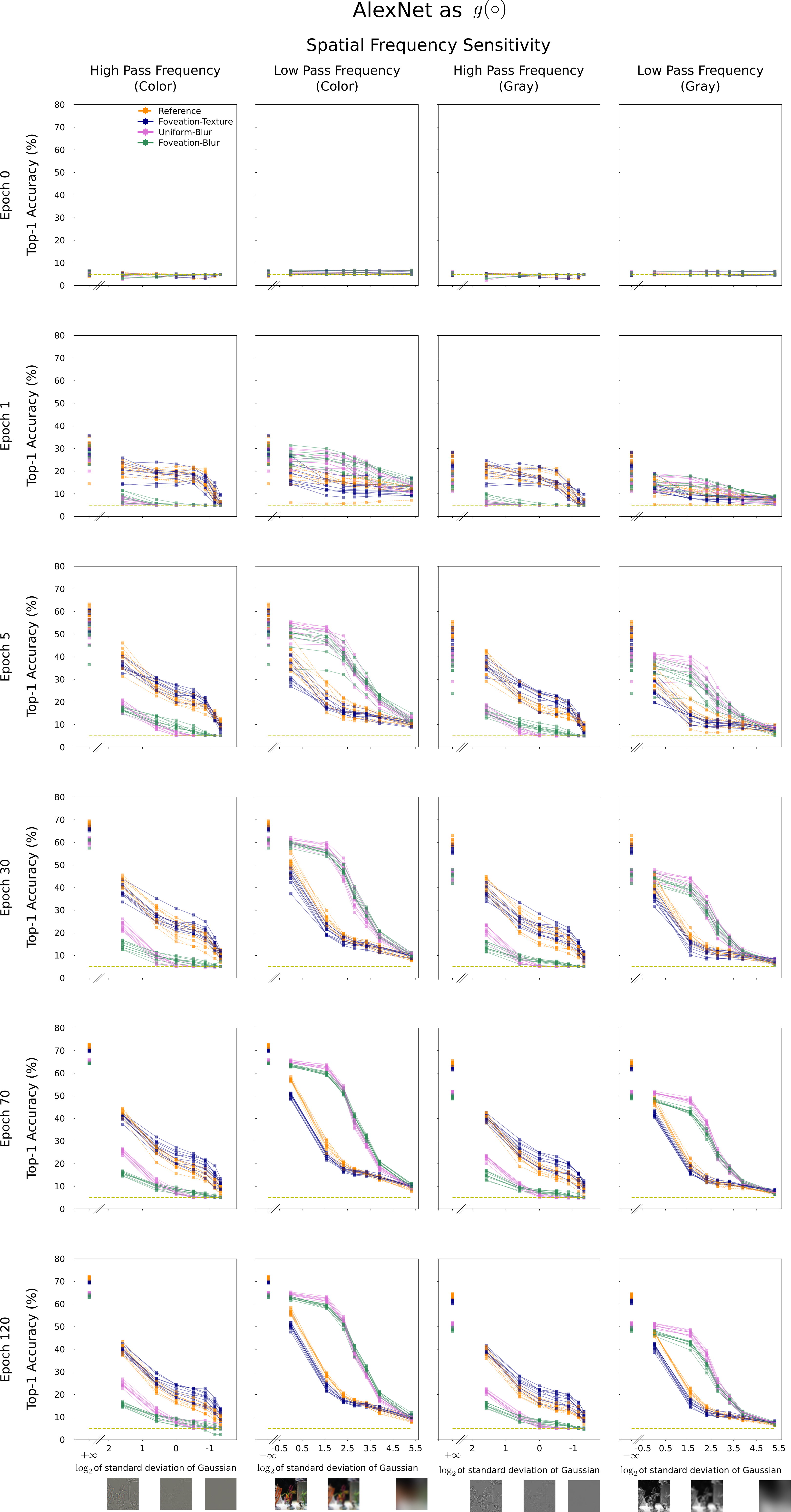}\caption{Individual Spatial Frequency Sensitivity for AlexNet as $g(\circ)$ after epochs 0, 1, 5, 30, 70, 120.}\label{fig:Spatial_Frequency_AlexNet_Individual}
\end{figure*}

\clearpage
\newpage

\begin{figure*}[!t]
\centering
\includegraphics[width=0.8\columnwidth,clip=false,draft=false,]{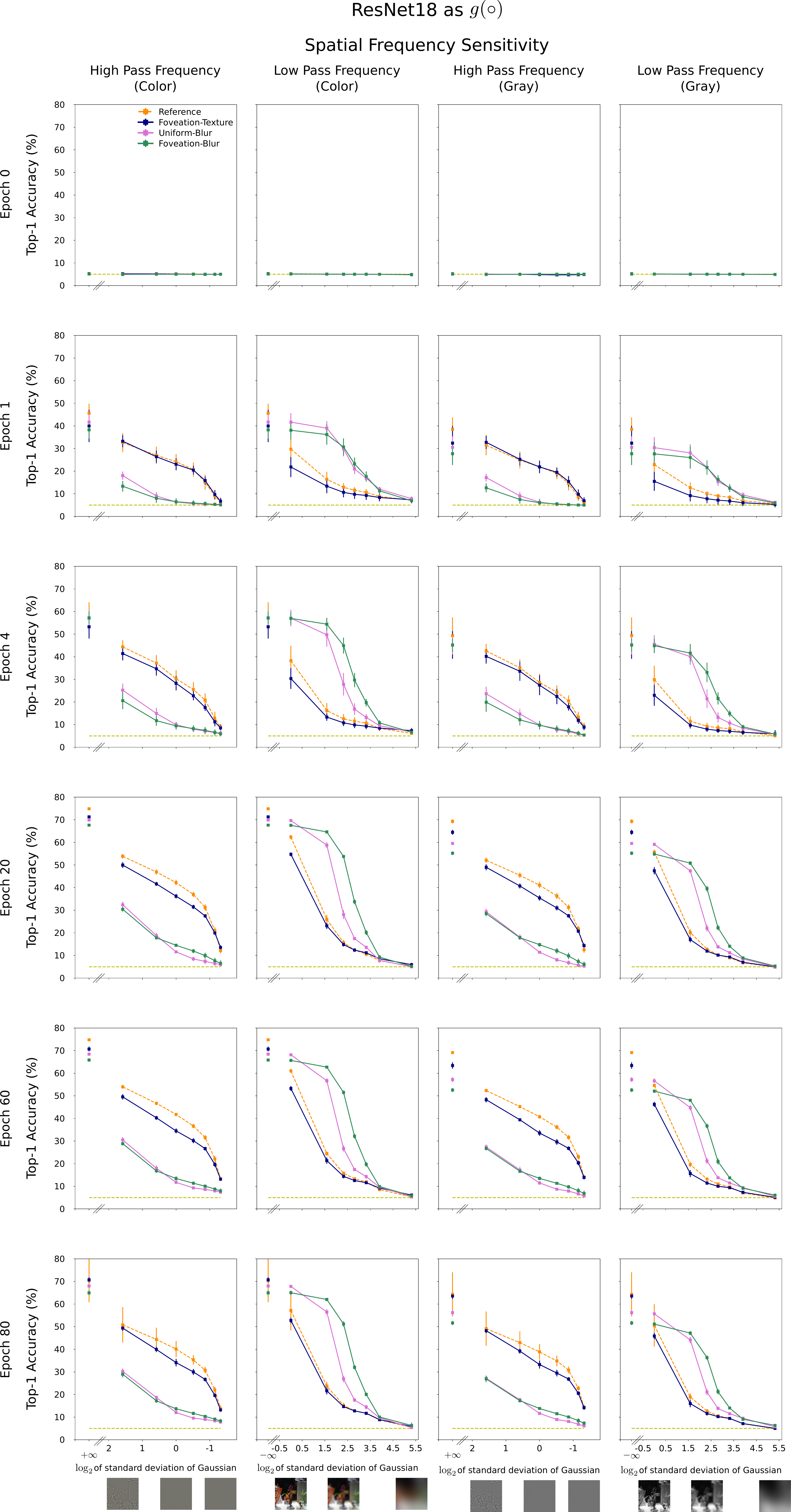}\caption{Aggregate Spatial Frequency Sensitivity for ResNet18 as $g(\circ)$ after epochs 0, 1, 4, 20, 60, 80.}\label{fig:Spatial_Frequency_ResNet18_Average}
\end{figure*}

\clearpage
\newpage

\begin{figure*}[!t]
\centering
\includegraphics[width=0.8\columnwidth,clip=false,draft=false,]{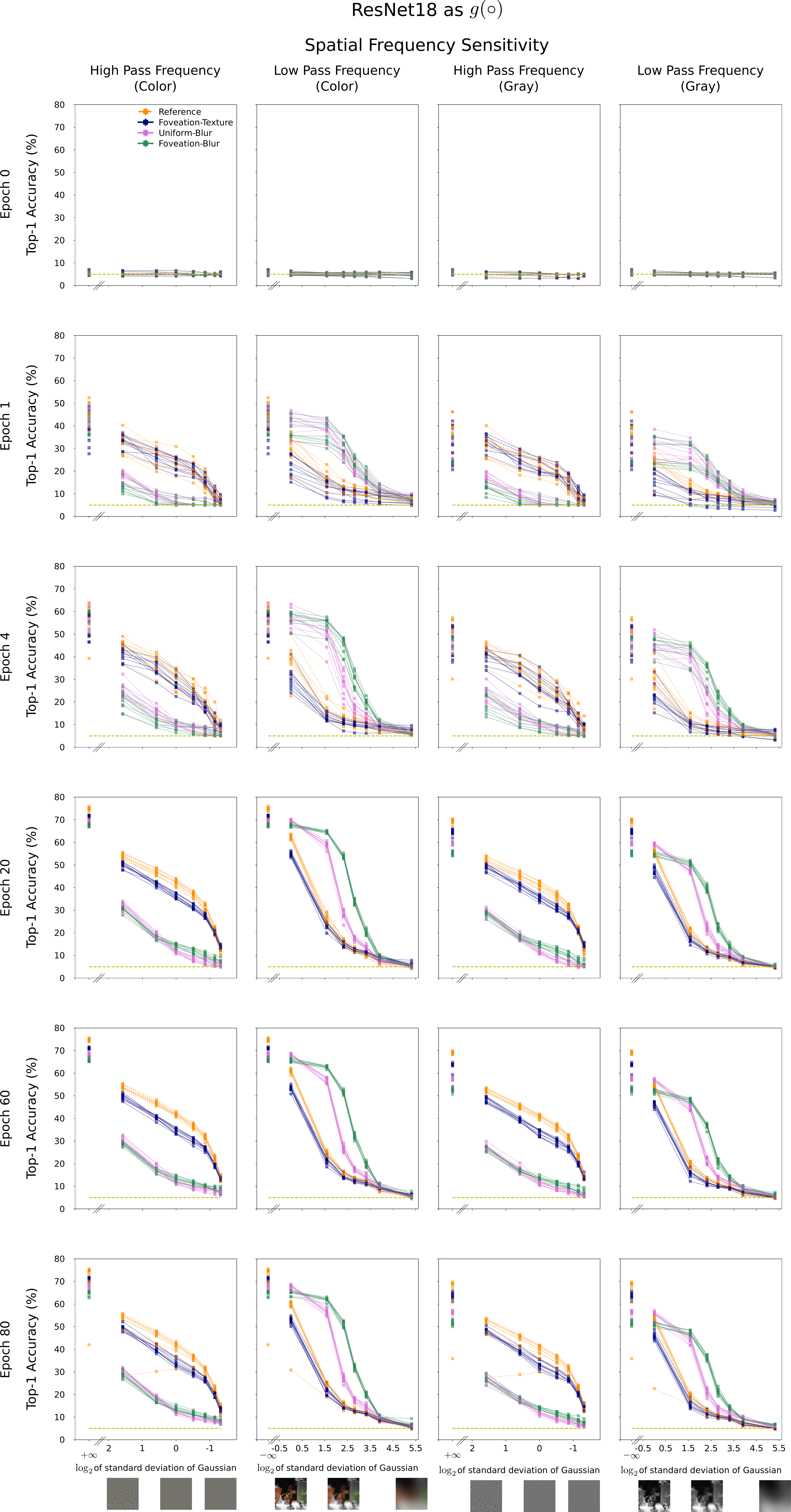}\caption{Individual Spatial Frequency Sensitivity for ResNet18 as $g(\circ)$ after epochs 0, 1, 4, 20, 60, 80.}\label{fig:Spatial_Frequency_ResNet18_Individual}
\end{figure*}

\clearpage
\newpage

\begin{figure}[!t]
\centering\includegraphics[width=0.8\columnwidth,clip=true,draft=false,]{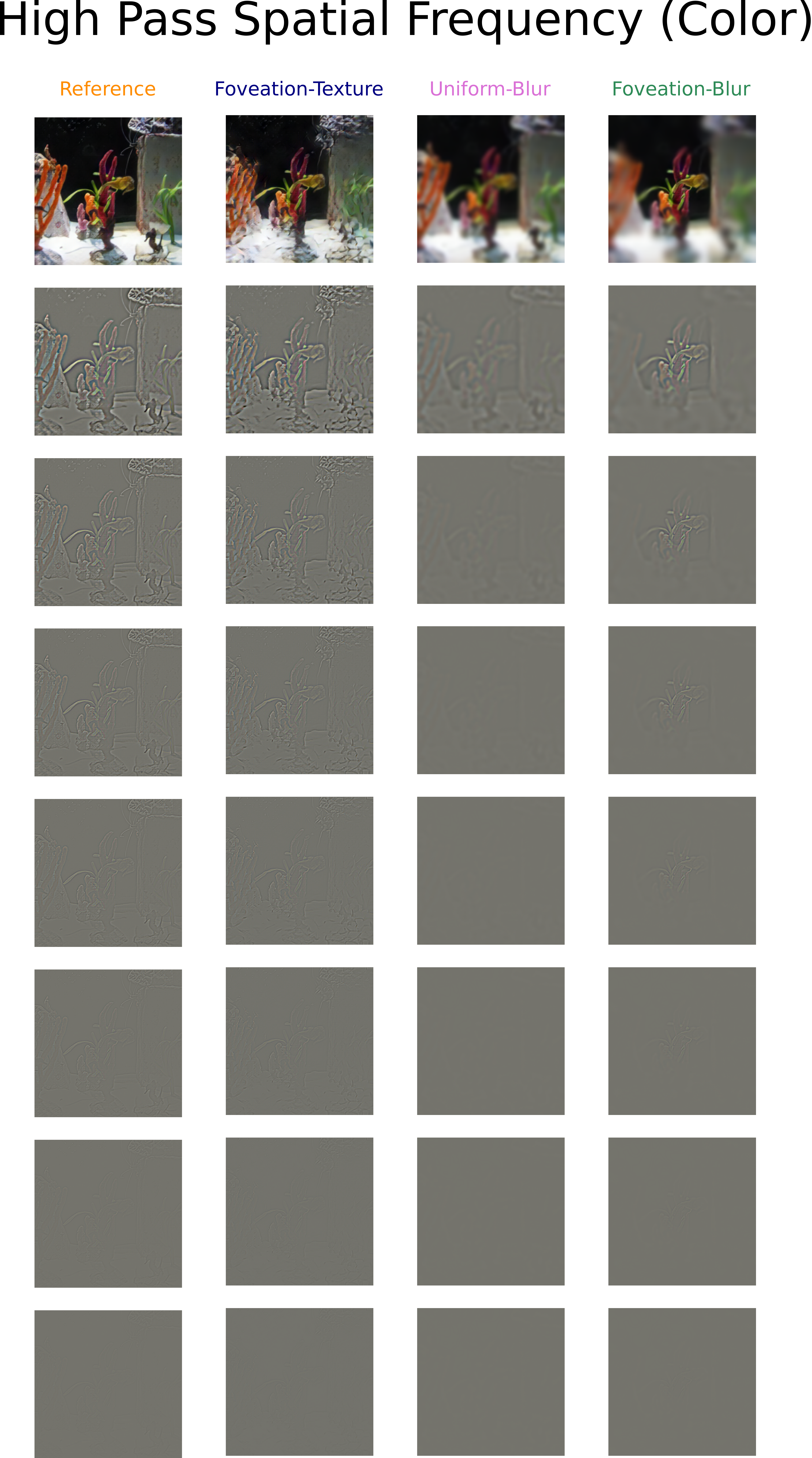}\caption{Sample High Pass Spatial Frequency Color Stimuli.
}\label{fig:HP_SF_Color_Stimuli}
\end{figure}

\clearpage
\newpage

\begin{figure}[!t]
\centering\includegraphics[width=0.8\columnwidth,clip=true,draft=false,]{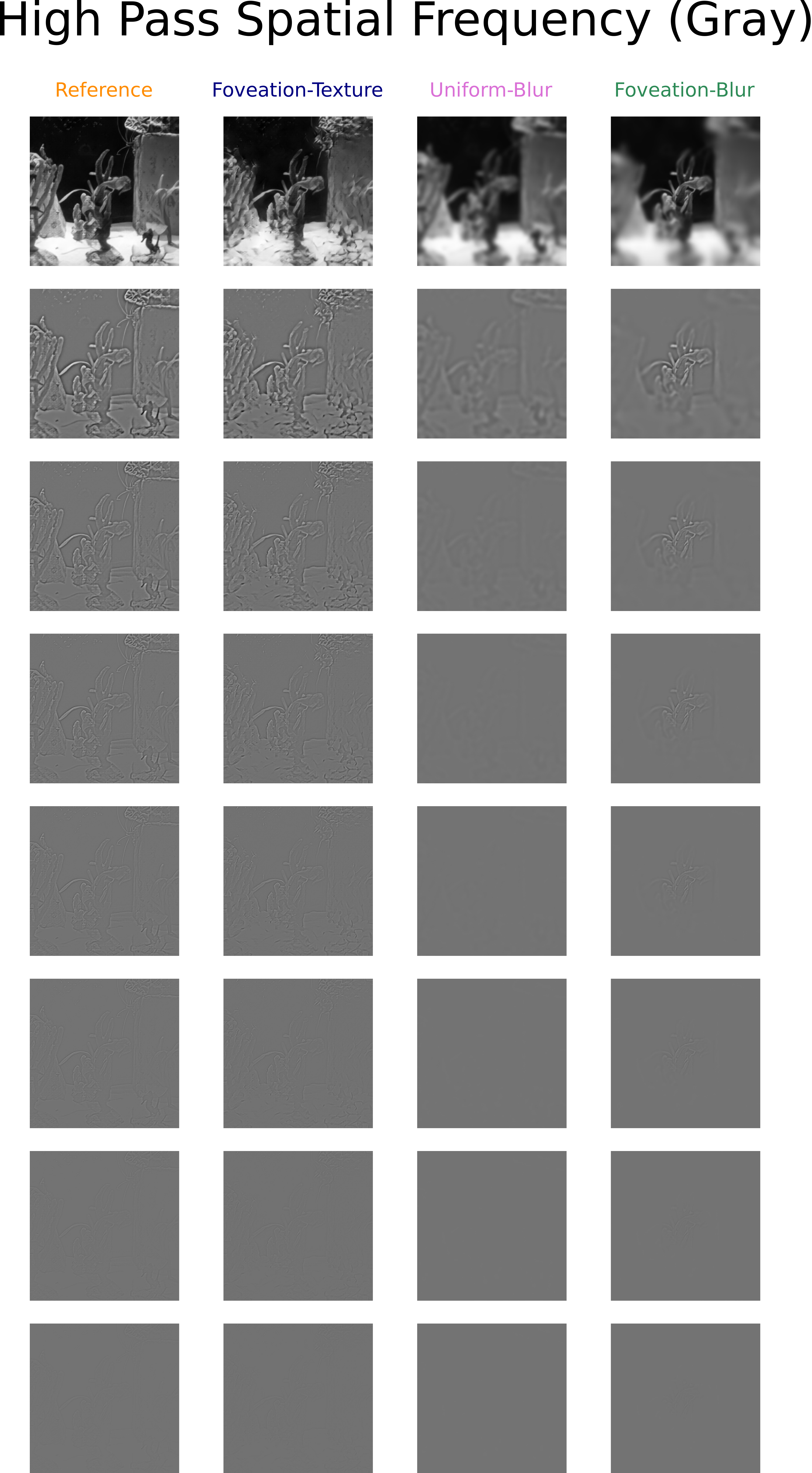}\caption{Sample High Pass Spatial Frequency Gray Stimuli.
}\label{fig:HP_SF_Gray_Stimuli}
\end{figure}

\clearpage
\newpage

\begin{figure}[!t]

\centering\includegraphics[width=0.8\columnwidth,clip=true,draft=false,]{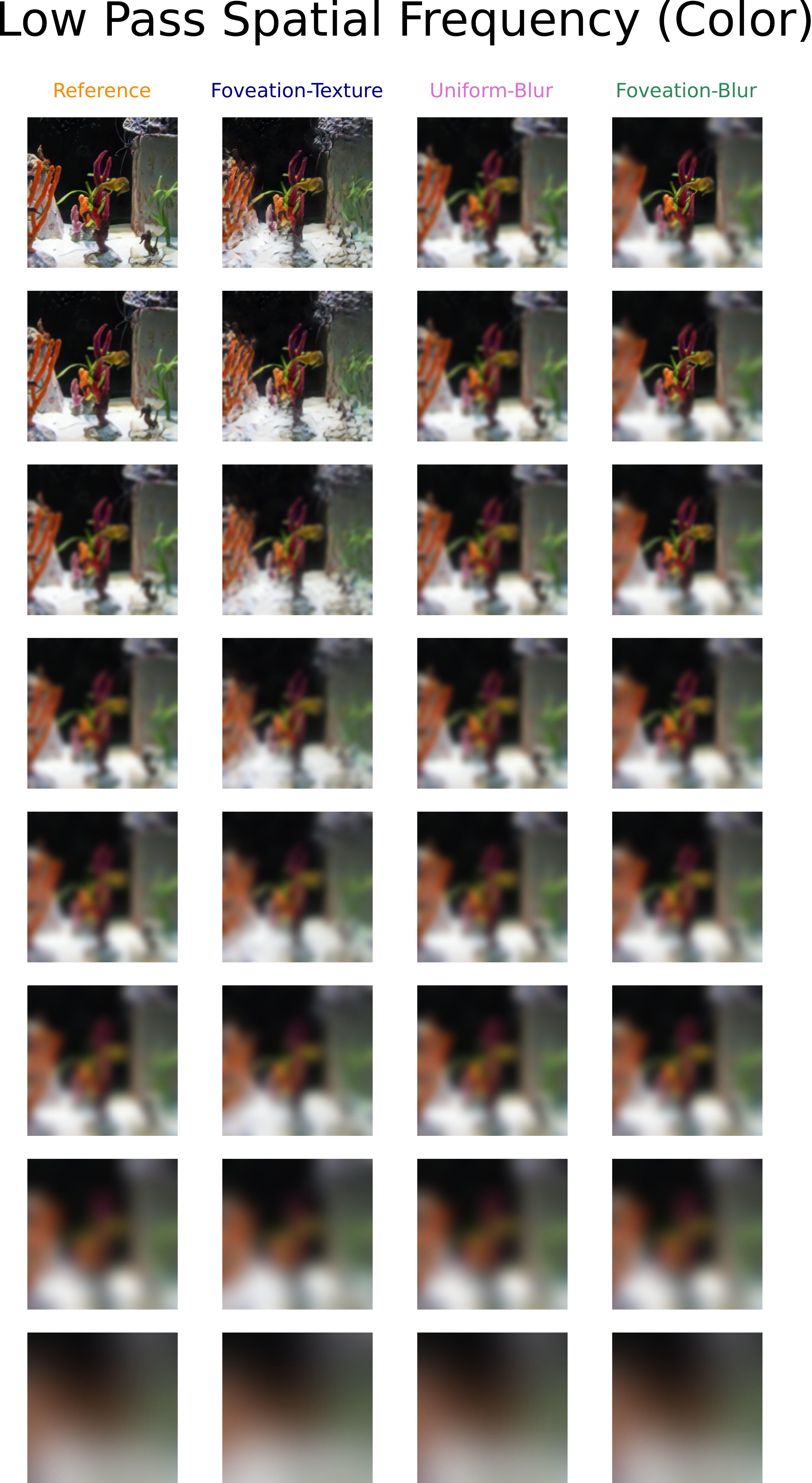}\caption{Sample Low Pass Spatial Frequency Color Stimuli.
}\label{fig:LP_SF_Color_Stimuli}
\end{figure}

\clearpage
\newpage

\begin{figure}[!t]

\centering\includegraphics[width=0.8\columnwidth,clip=true,draft=false,]{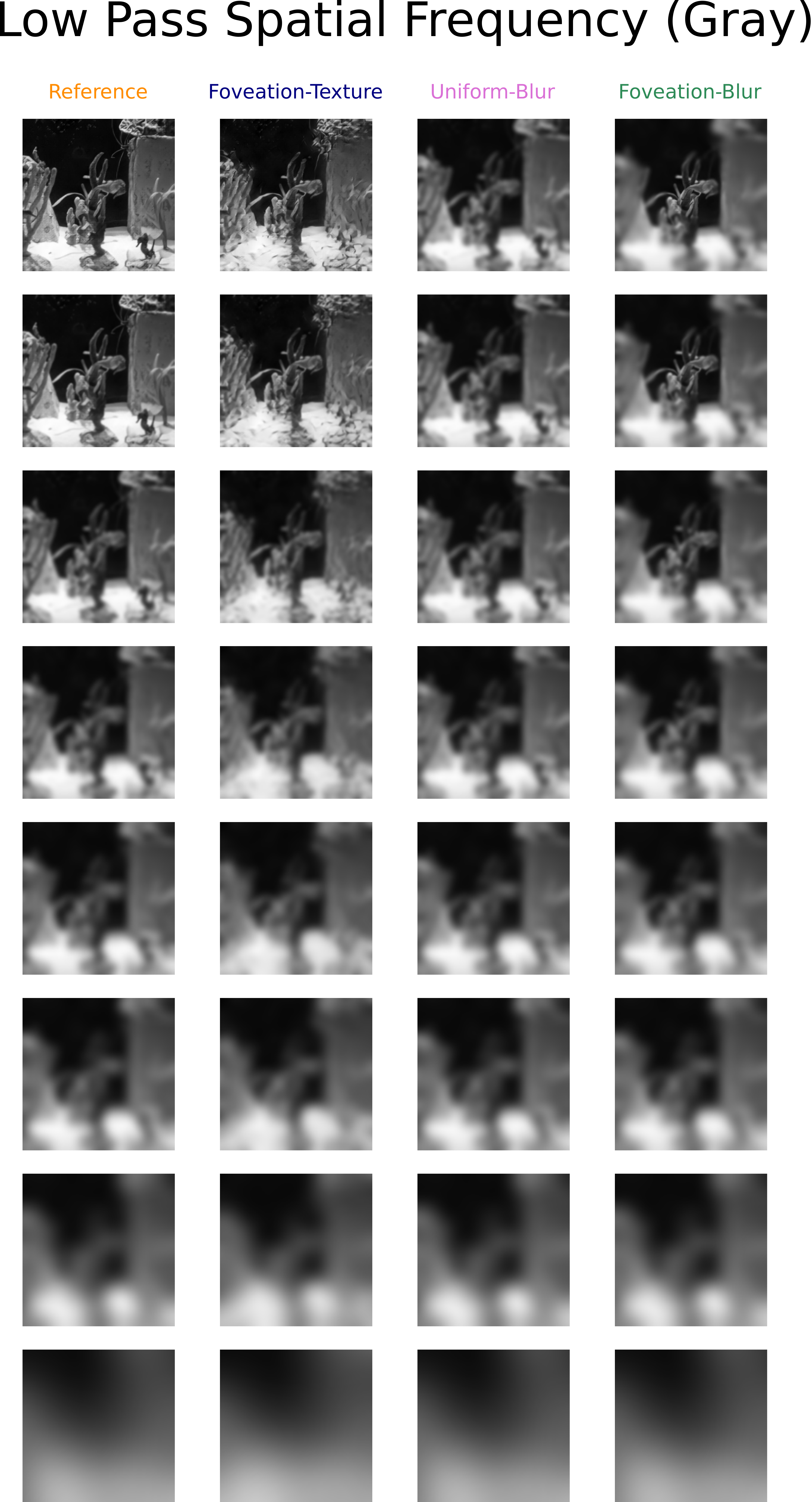}\caption{Sample Low Pass Spatial Frequency Gray Stimuli.
}\label{fig:LP_SF_Gray_Stimuli}
\end{figure}

\clearpage
\newpage

\section{Robustness to Occlusion}

\begin{figure*}[!h]
\centering
\includegraphics[width=0.75\columnwidth,clip=false,draft=false,]{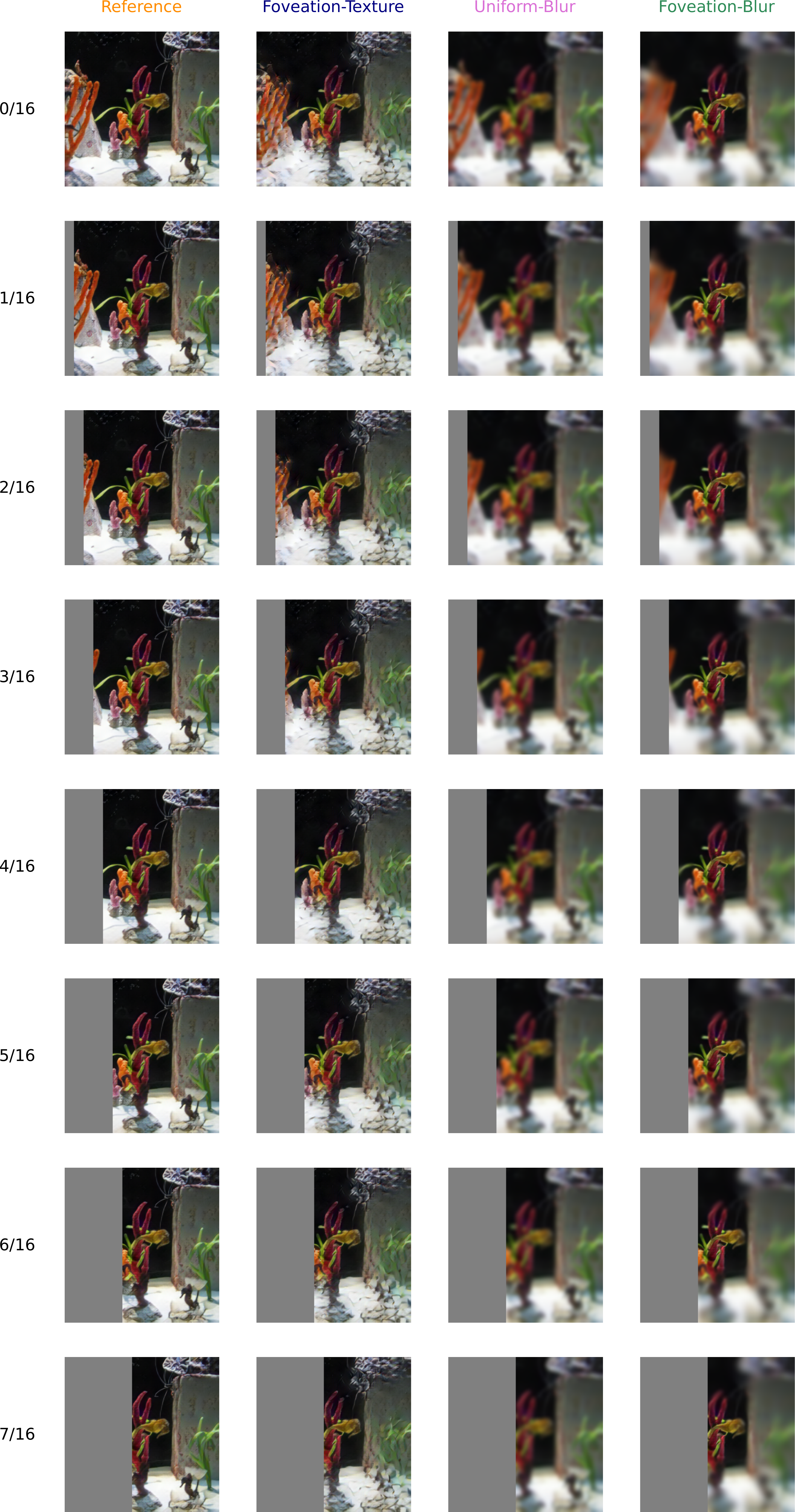}\caption{Left2Right Occlusion Sample Stimuli.}\label{fig:Left2Right_1}
\end{figure*}

\clearpage
\newpage

\begin{figure*}[!h]
\centering
\includegraphics[width=0.75\columnwidth,clip=false,draft=false,]{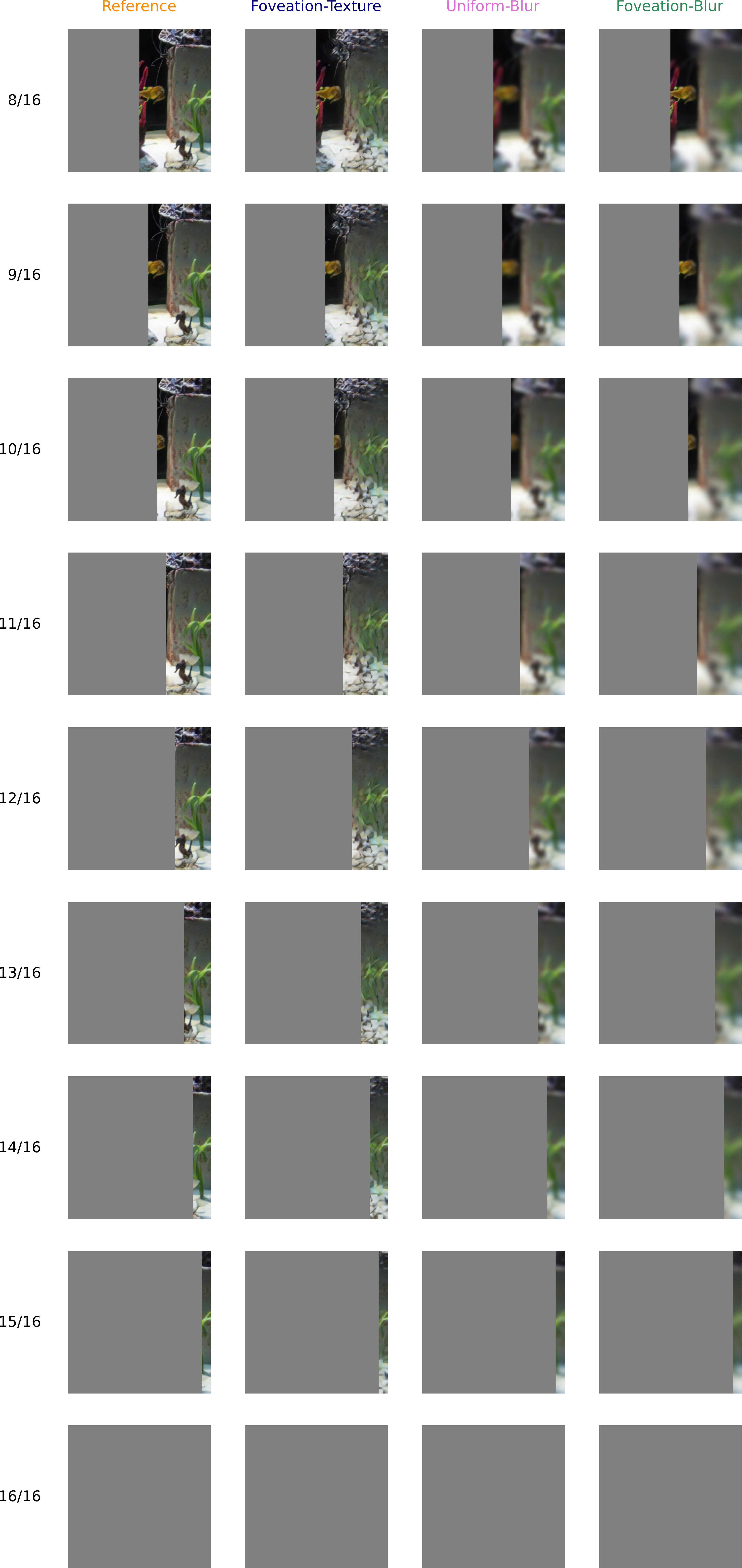}\caption{Left2Right Occlusion Sample Stimuli.}\label{fig:Left2Right_2}
\end{figure*}

\clearpage
\newpage

\begin{figure*}[!h]
\centering
\includegraphics[width=0.75\columnwidth,clip=false,draft=false,]{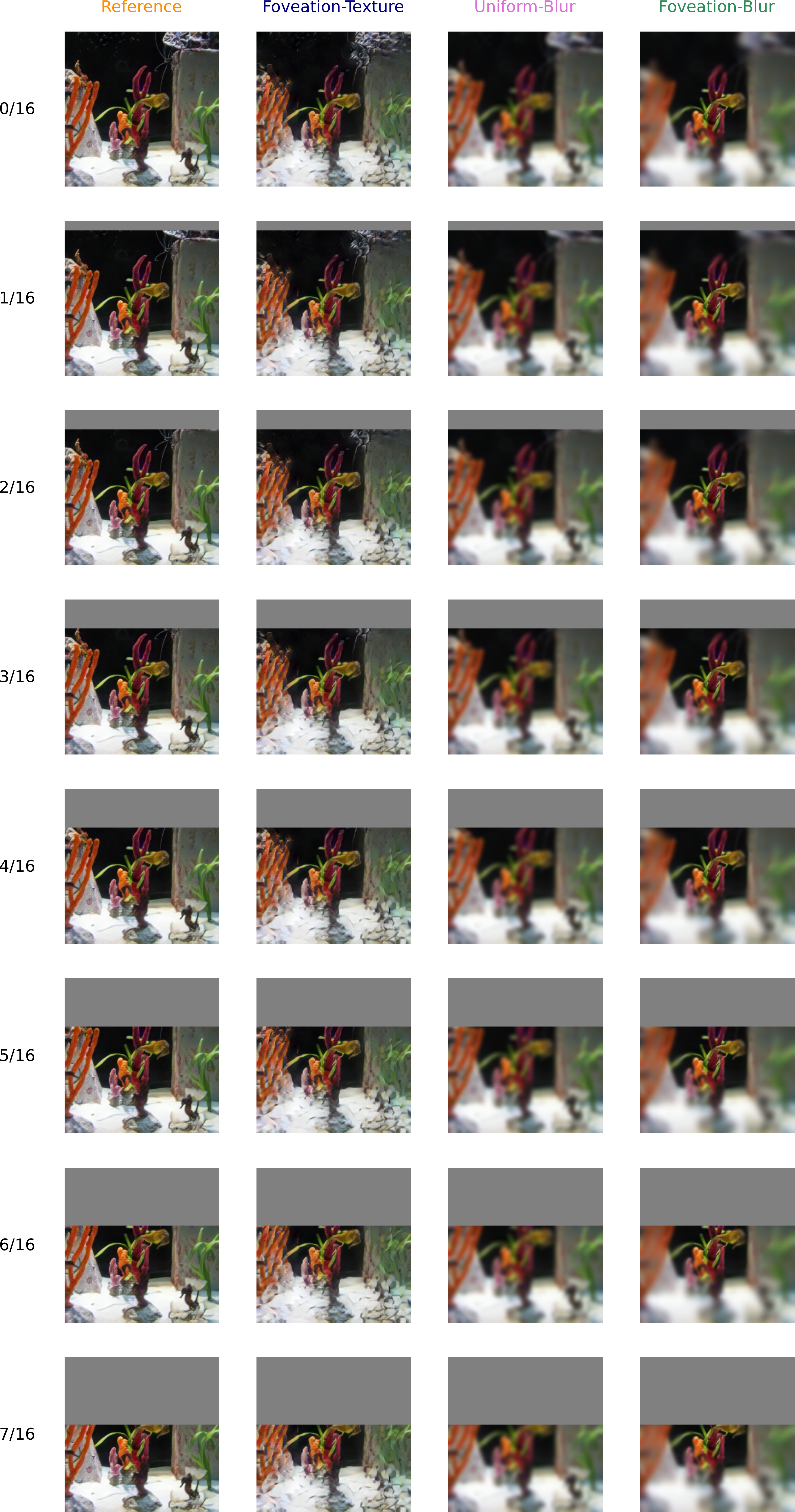}\caption{Top2Bottom Occlusion Sample Stimuli.}\label{fig:Top2Bottom_1}
\end{figure*}

\clearpage
\newpage

\begin{figure*}[!h]
\centering
\includegraphics[width=0.75\columnwidth,clip=false,draft=false,]{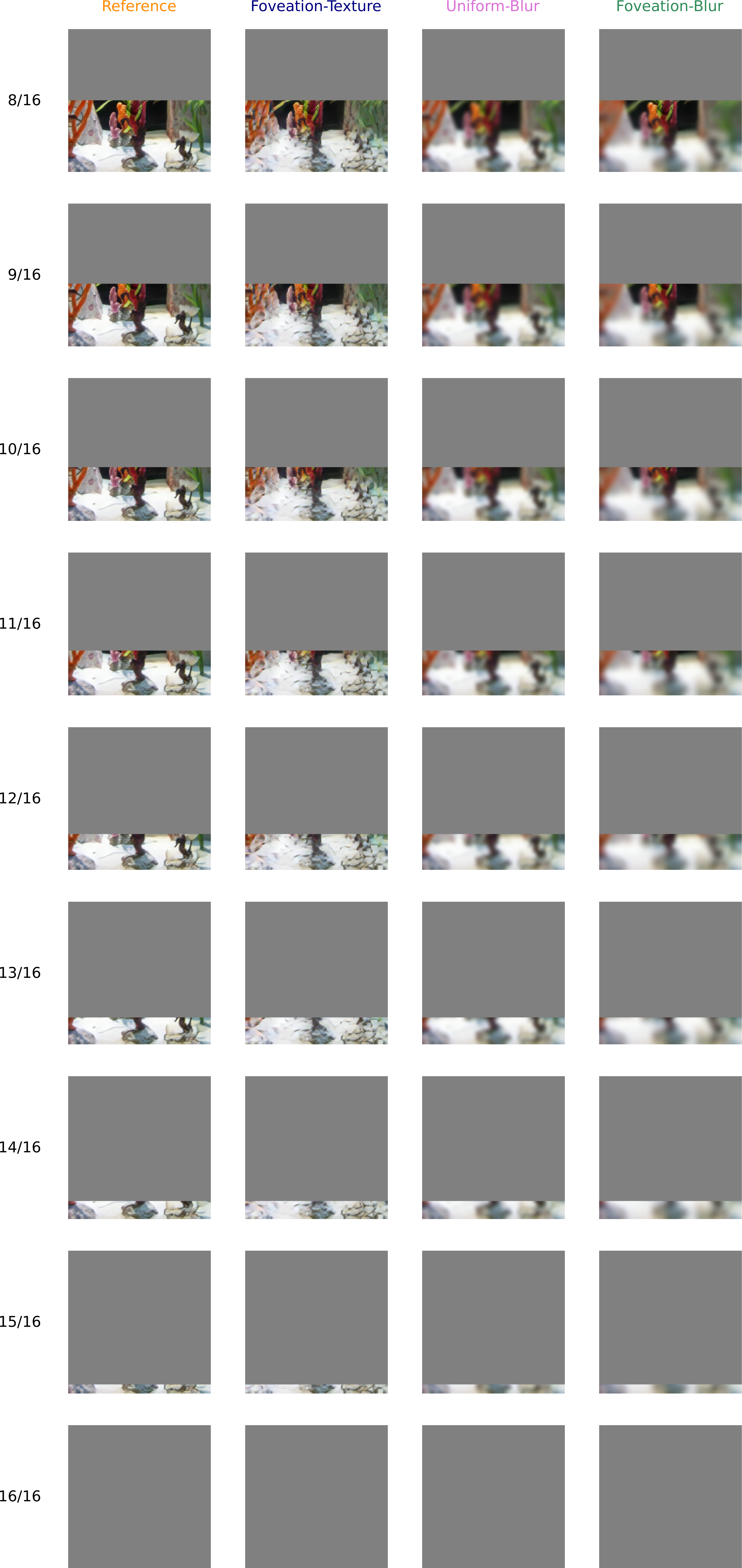}\caption{Top2Bottom Occlusion Sample Stimuli.}\label{fig:Top2Bottom_2}
\end{figure*}

\clearpage
\newpage

\begin{figure*}[!h]
\centering
\includegraphics[width=0.75\columnwidth,clip=false,draft=false,]{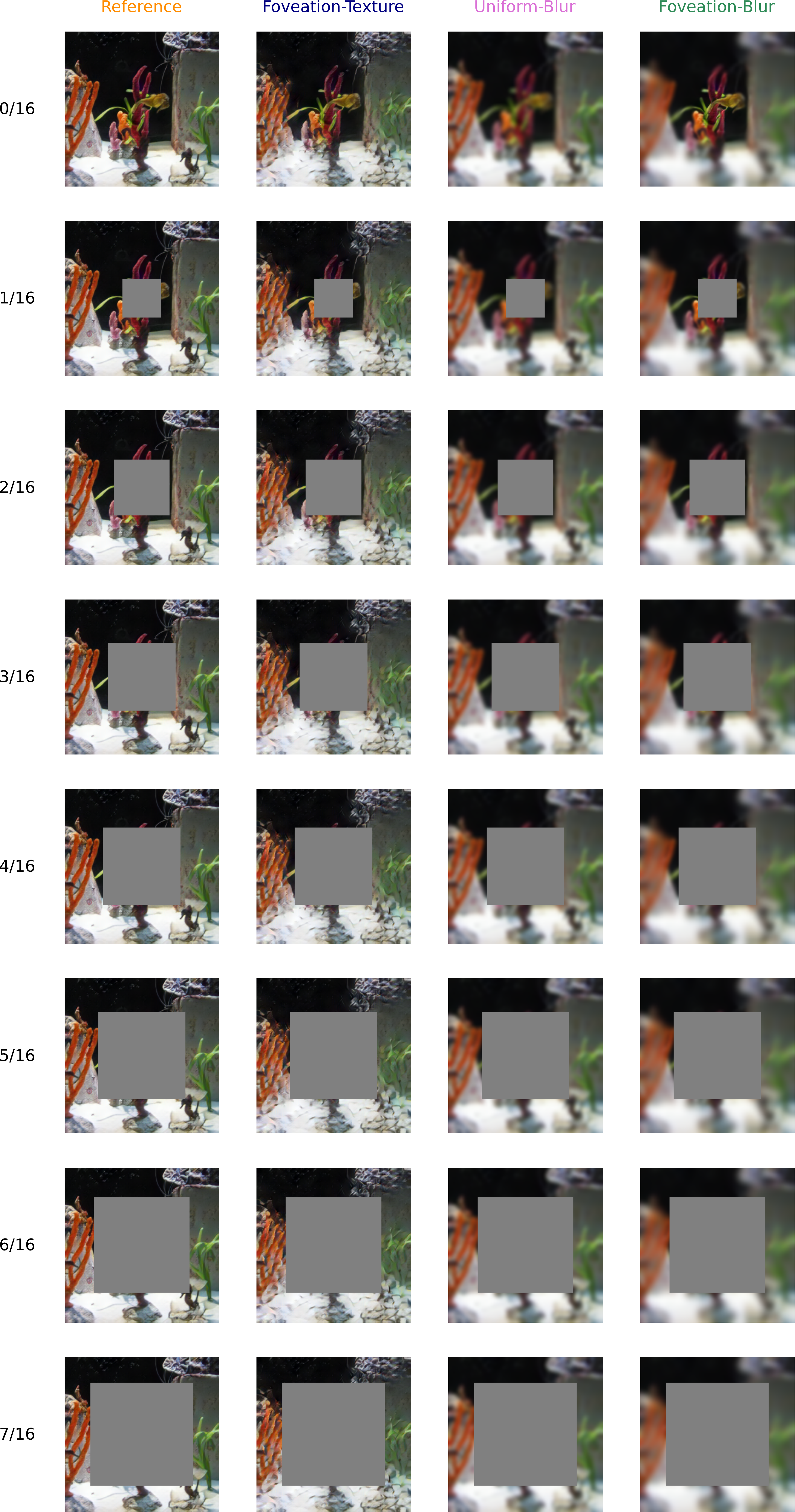}\caption{Scotoma Occlusion Sample Stimuli.}\label{fig:Scotoma_1}
\end{figure*}

\clearpage
\newpage

\begin{figure*}[!h]
\centering
\includegraphics[width=0.75\columnwidth,clip=false,draft=false,]{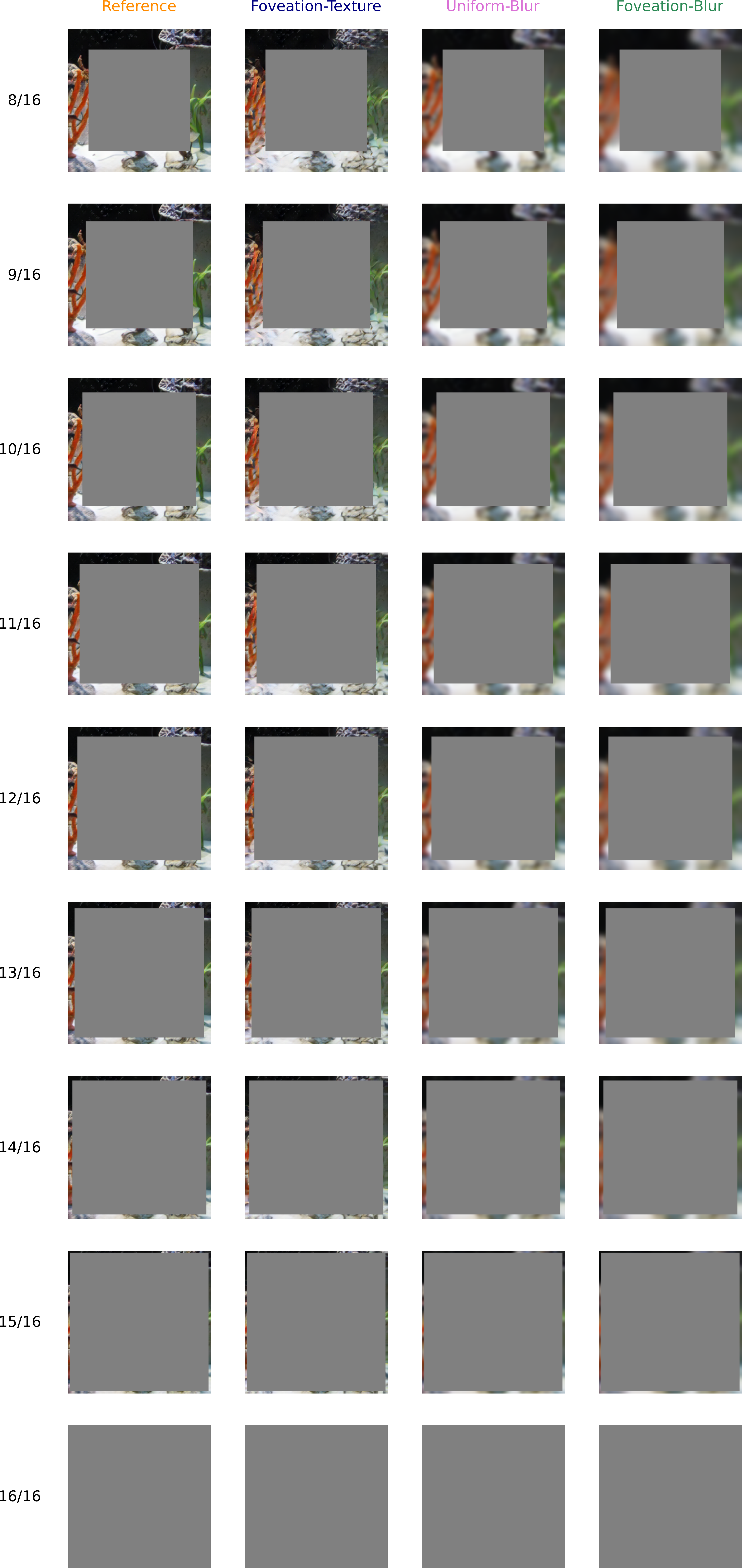}\caption{Scotoma Occlusion Sample Stimuli.}\label{fig:Scotoma_2}
\end{figure*}

\clearpage
\newpage

\begin{figure*}[!h]
\centering
\includegraphics[width=0.75\columnwidth,clip=false,draft=false,]{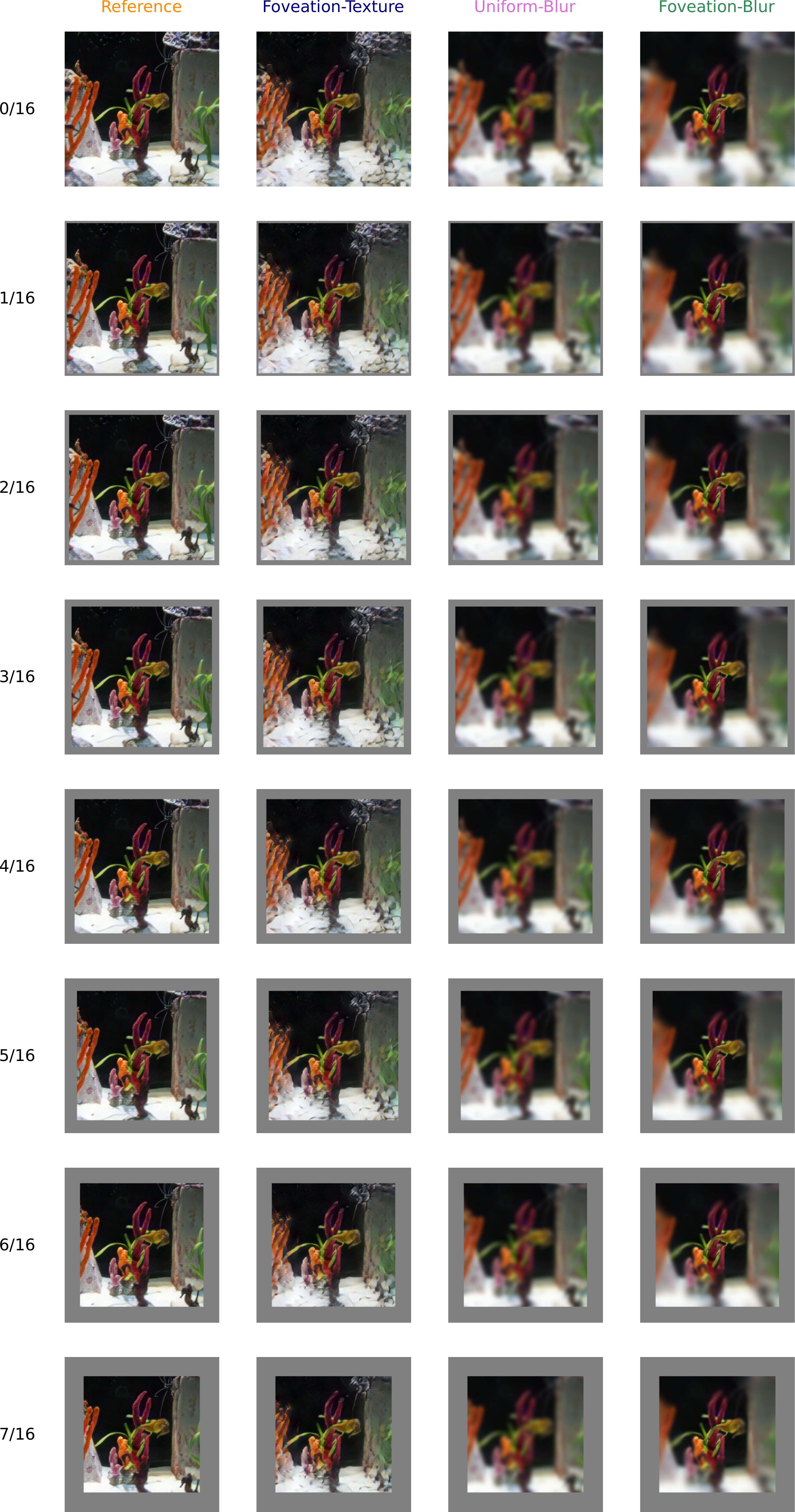}\caption{Glaucoma Occlusion Sample Stimuli.}\label{fig:Glaucoma_1}
\end{figure*}

\clearpage
\newpage

\begin{figure*}[!h]
\centering
\includegraphics[width=0.75\columnwidth,clip=false,draft=false,]{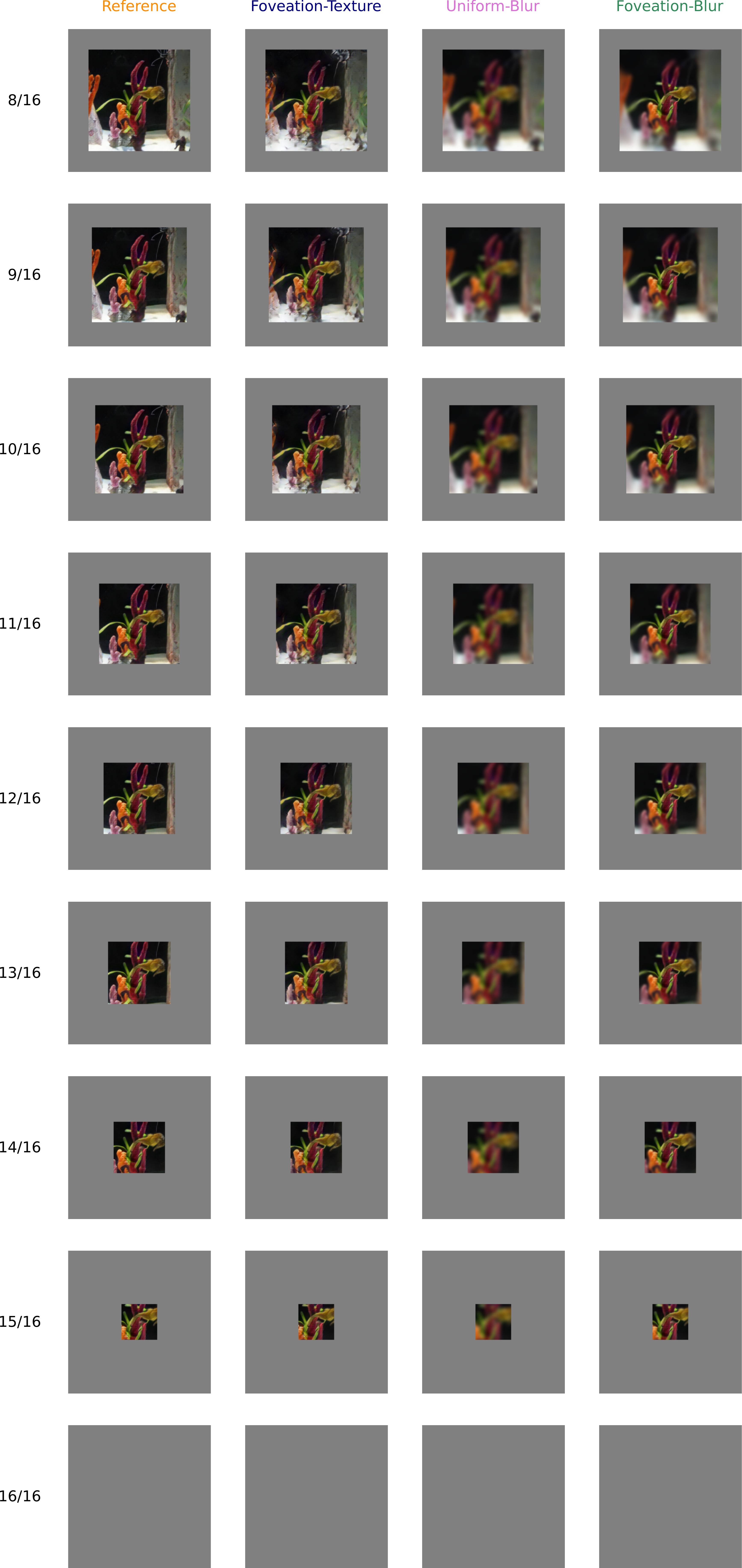}\caption{Glaucoma Occlusion Sample Stimuli.}\label{fig:Glaucoma_2}
\end{figure*}

\clearpage
\newpage

\begin{figure*}[!h]
\centering
\includegraphics[width=1.0\columnwidth,clip=false,draft=false,]{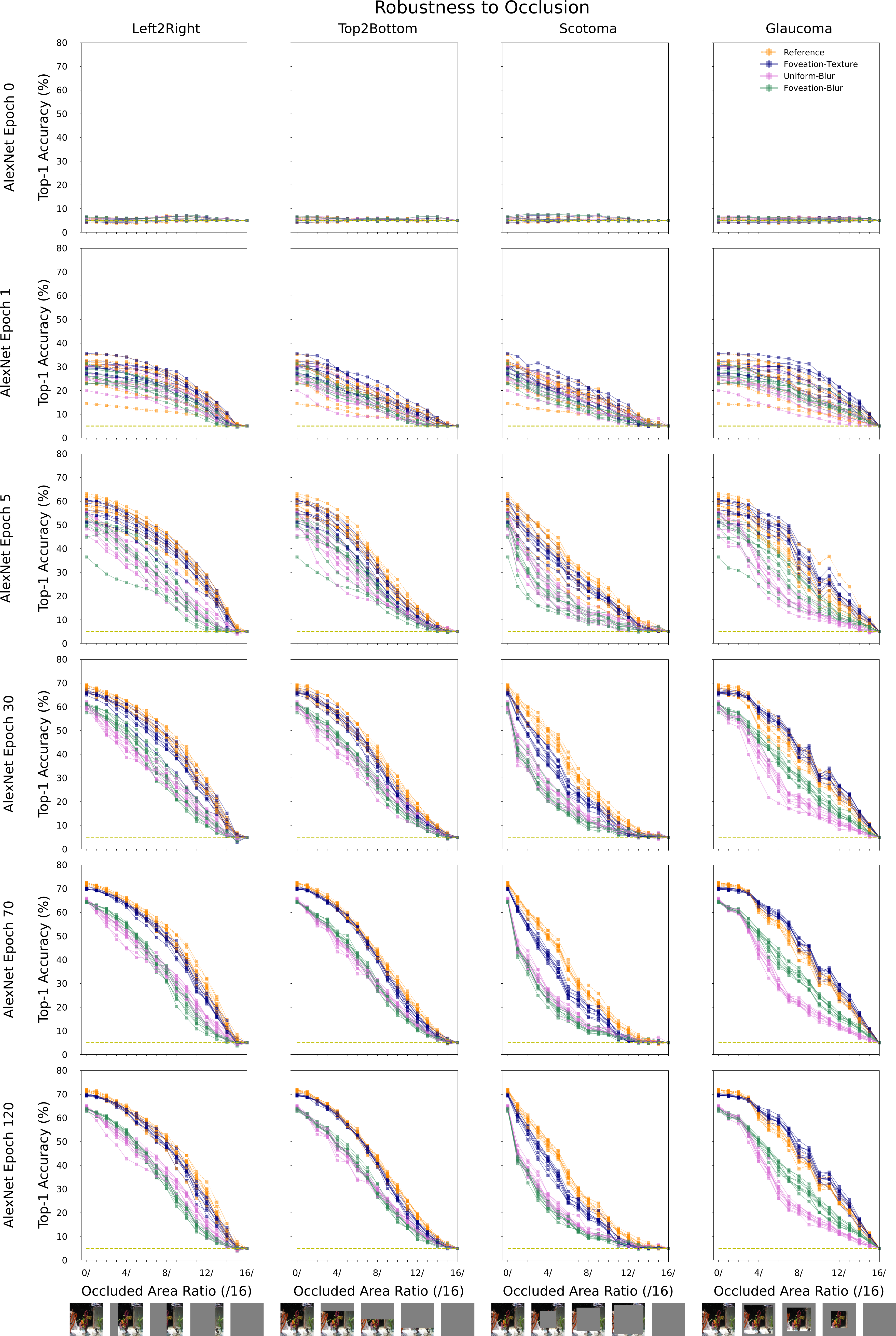}\caption{Individual Robustness to Occlusion plots for AlexNet as $g(\circ)$ after epochs 0, 1, 5, 30, 70, 120.}\label{fig:Occlusion_AlexNet_Supplement_Individual}
\end{figure*}

\newpage
\clearpage

\begin{figure*}[!t]
\centering
\includegraphics[width=1.0\columnwidth,clip=false,draft=false,]{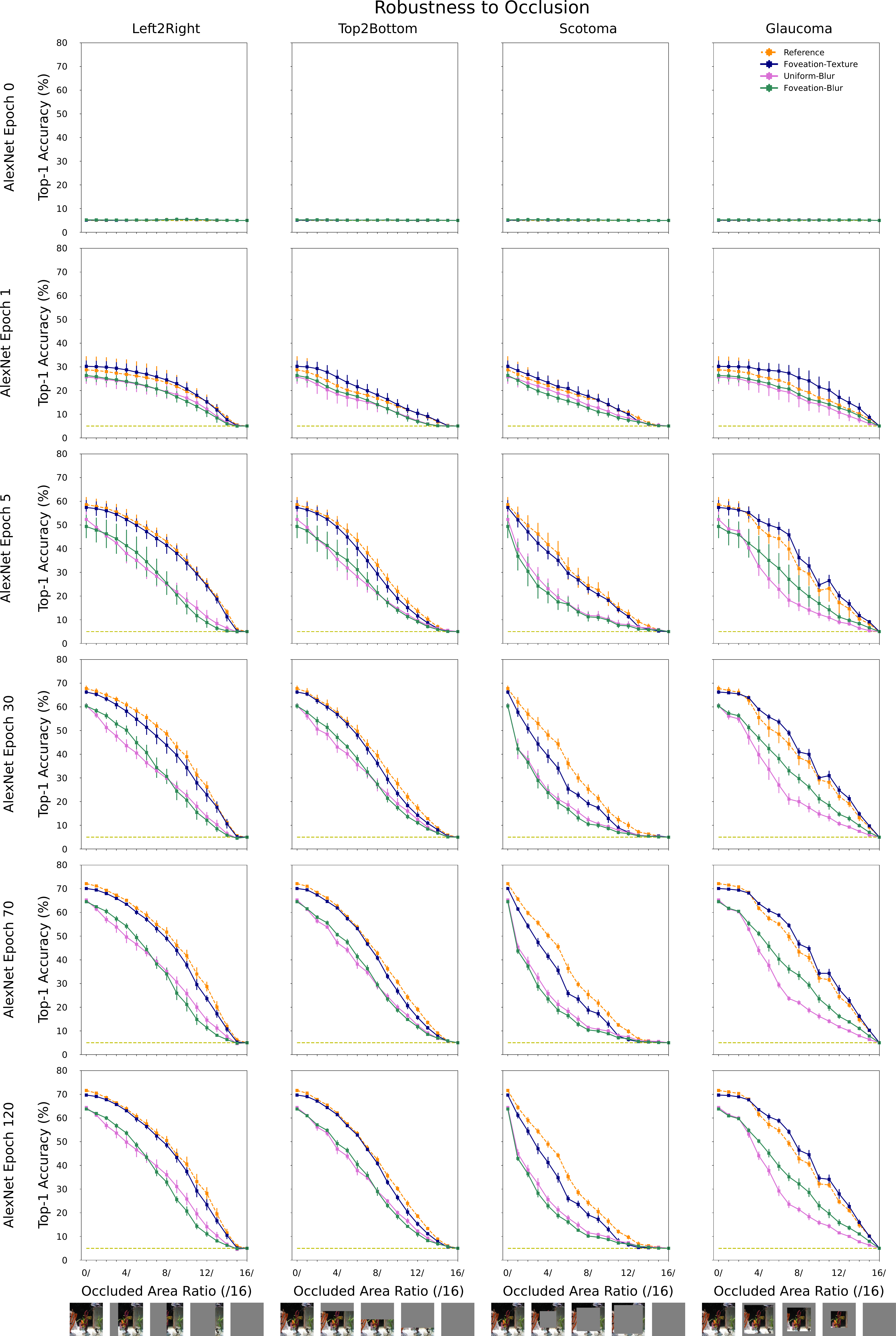}\caption{Aggregate Robustness to Occlusion plot for AlexNet as $g(\circ)$ after epochs 0, 1, 5, 30, 70, 120.}\label{fig:Occlusion_AlexNet_Supplement_Average}
\end{figure*}

\newpage
\clearpage

\begin{figure*}[!t]
\centering
\includegraphics[width=1.0\columnwidth,clip=false,draft=false,]{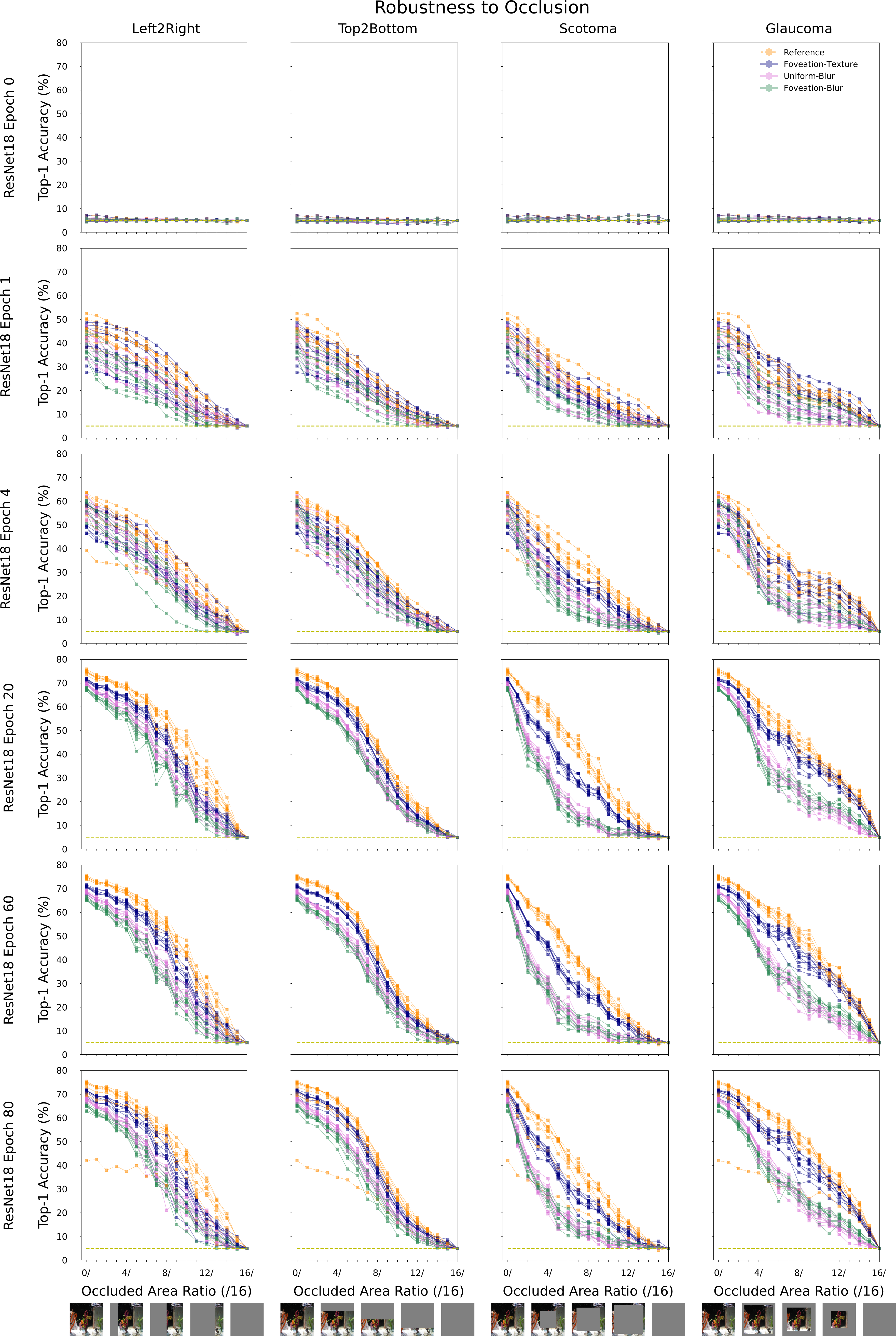}\caption{Individual Robustness to Occlusion plots for ResNet18 as $g(\circ)$ after epochs 0, 1, 4, 20, 60, 80.}\label{fig:Occlusion_ResNet18_Supplement_Individual}
\end{figure*}

\newpage
\clearpage

\begin{figure*}[!t]
\centering
\includegraphics[width=1.0\columnwidth,clip=false,draft=false,]{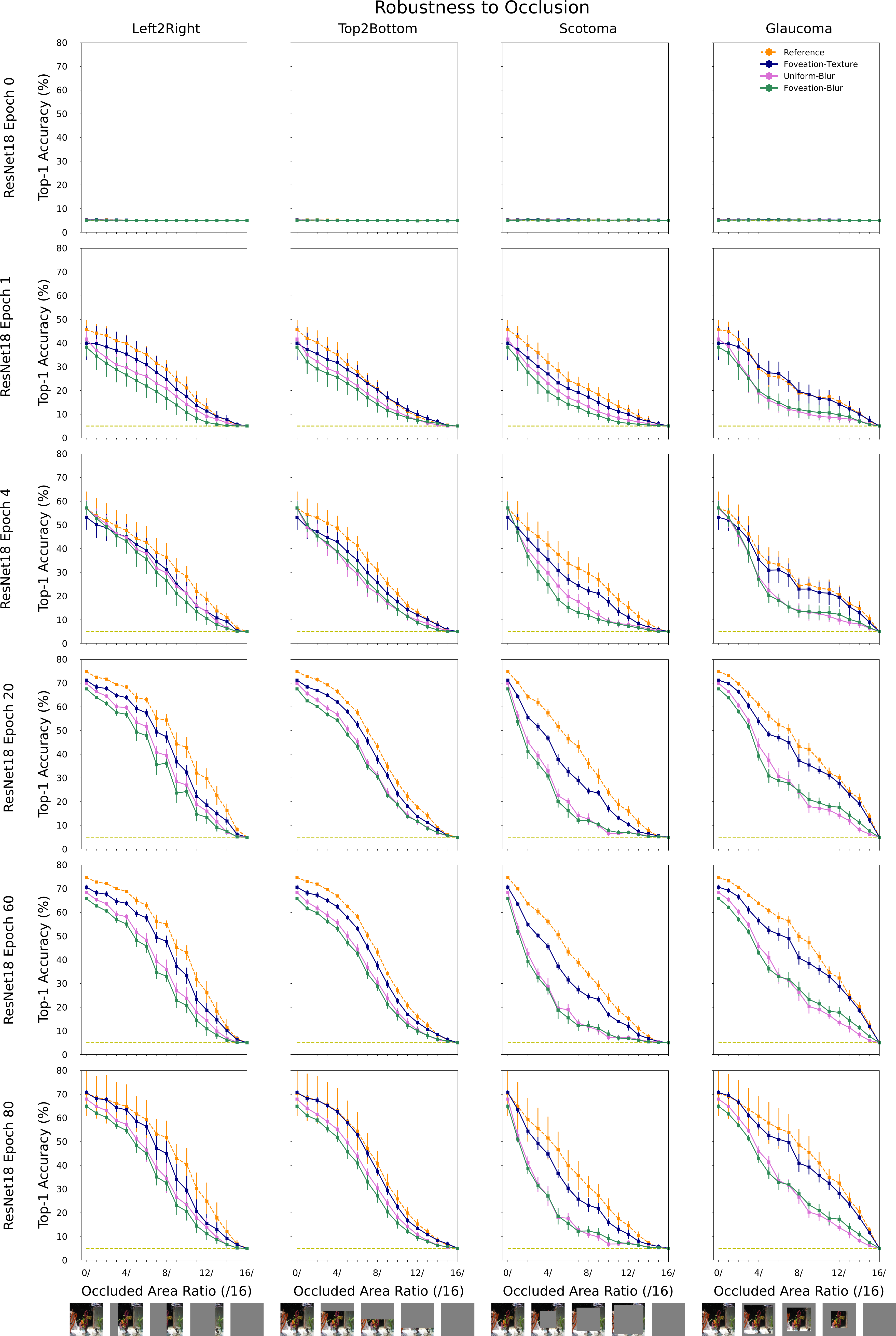}\caption{Aggregate Robustness to Occlusion plot for ResNet18 as $g(\circ)$ after epochs 0, 1, 4, 20, 60, 80.}\label{fig:Occlusion_ResNet18_Supplement_Average}
\end{figure*}

\newpage
\clearpage

\section{Window Cue-Conflict}

\begin{figure*}[!h]
\centering
\includegraphics[width=1.0\columnwidth,clip=false,draft=false,]{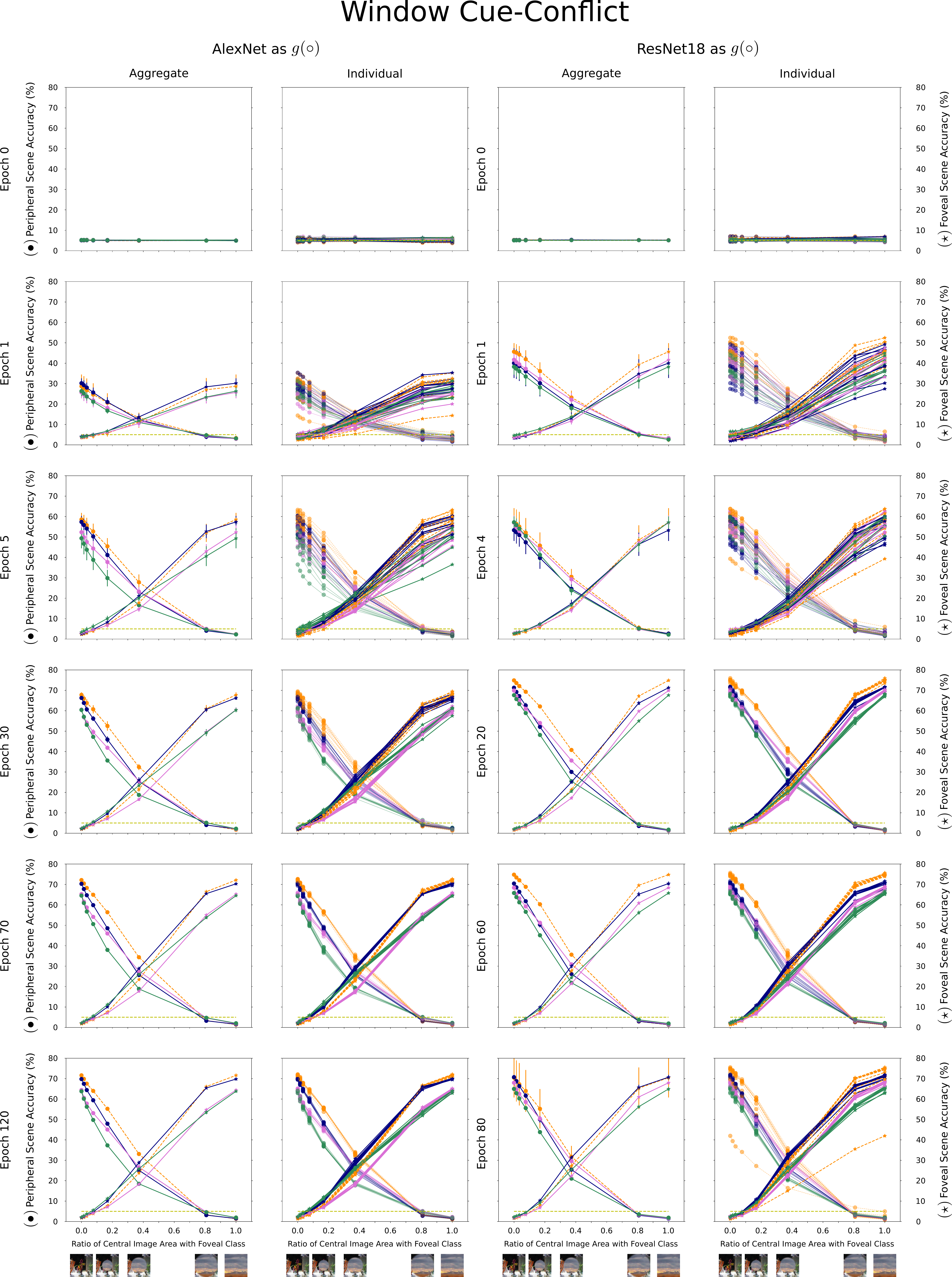}\caption{Aggregate and Individual Window Cue-Conflict plots for AlexNet and ResNet18 as $g(\circ)$ after epochs 0, 1, 5, 30, 70, 120 and 0, 1, 4, 20, 60, 80. respectively}\label{fig:Supplement_Window_Cue_Conflict}
\end{figure*}

\newpage
\clearpage

\begin{figure}[!t]
\centering\includegraphics[width=0.75\columnwidth,clip=true,draft=false,]{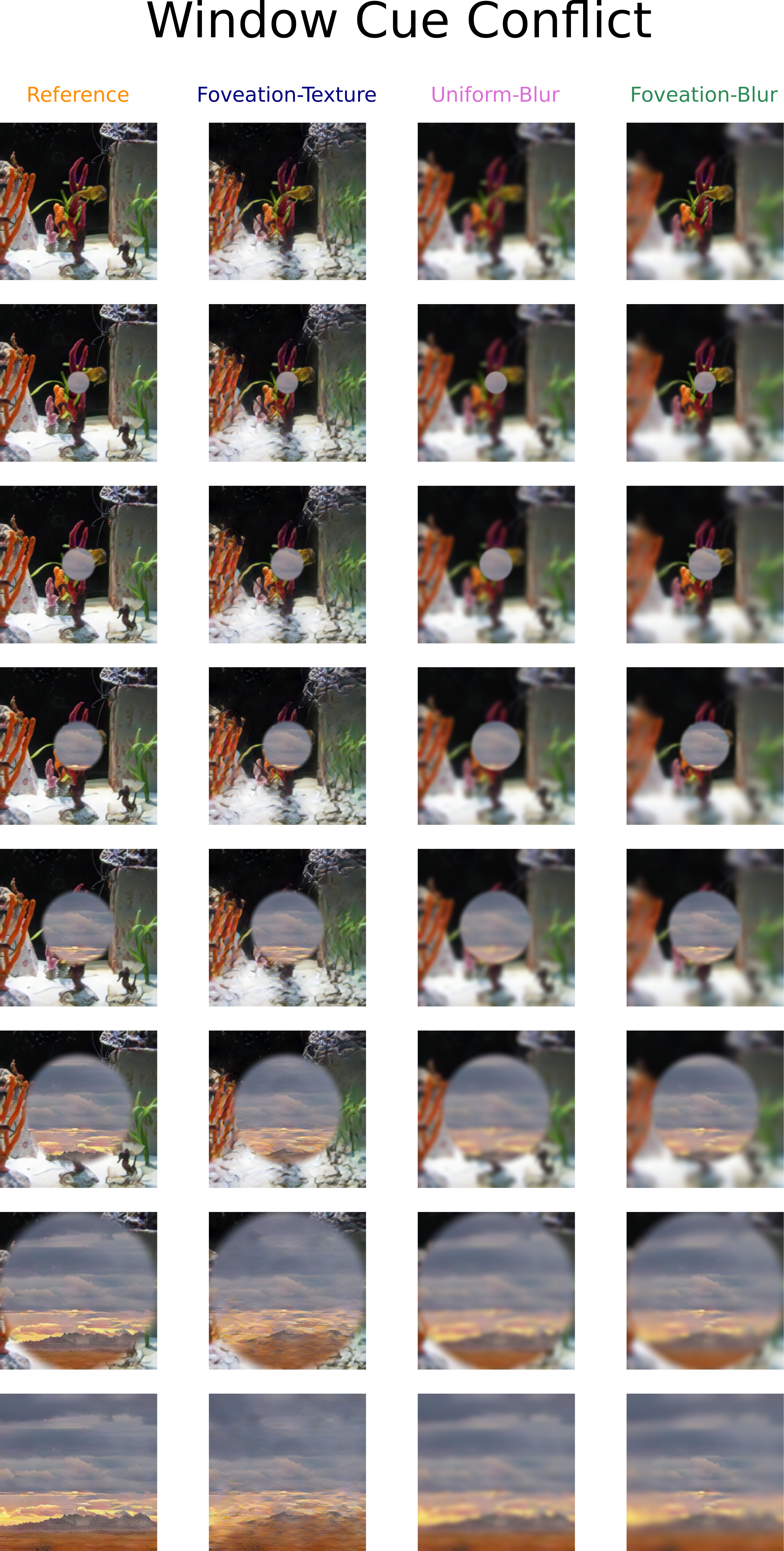}\caption{Sample Window Cue Conflict Stimuli.
}\label{fig:Window_Cue_Conflict_Stimuli}
\end{figure}

\newpage
\clearpage
\section{Square Cue-Conflict}

\begin{figure*}[!h]
\centering
\includegraphics[width=1.0\columnwidth,clip=false,draft=false,]{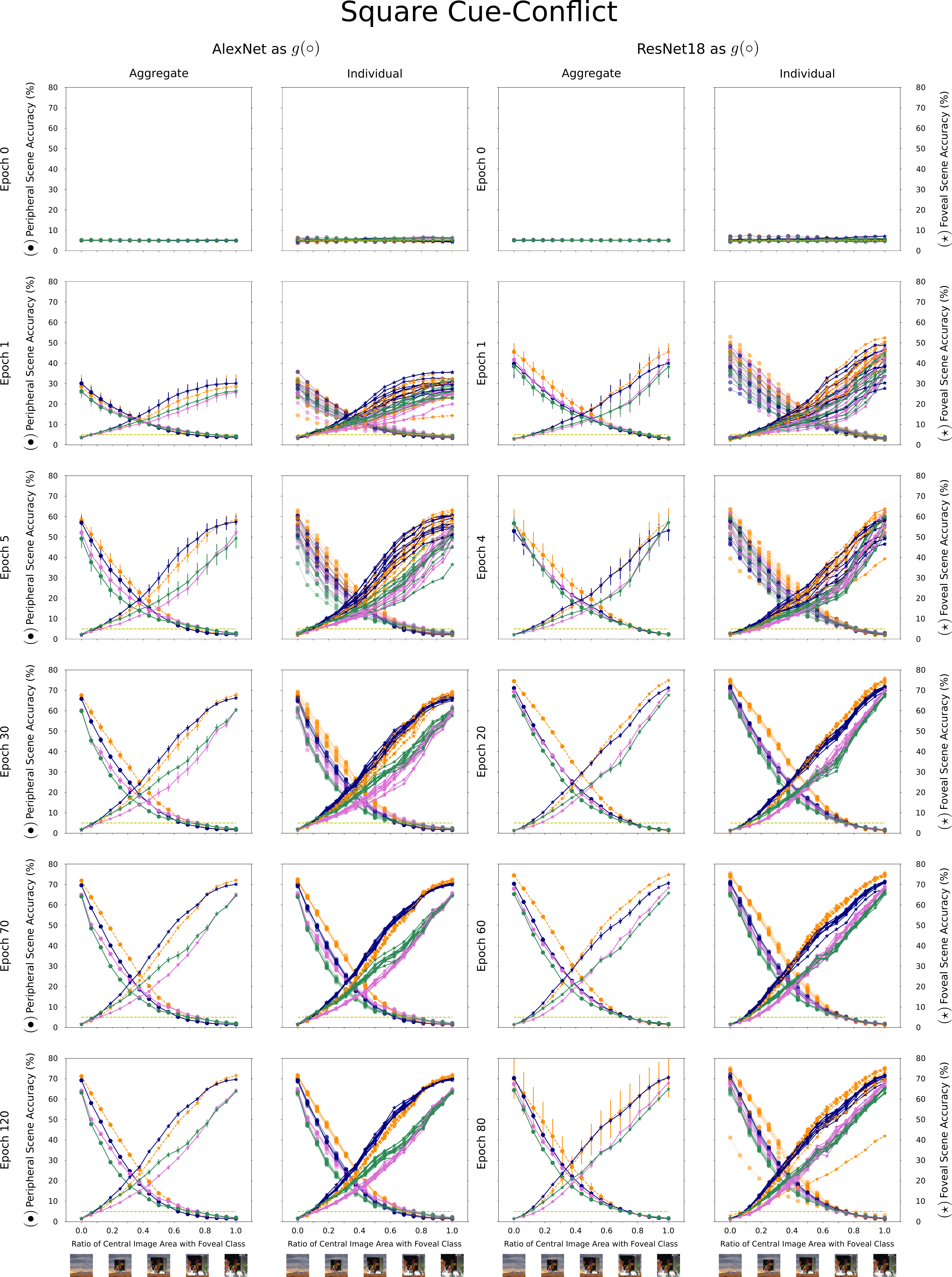}\caption{Aggregate and Individual Square Cue-Conflict plots for AlexNet and ResNet18 as $g(\circ)$ after epochs 0, 1, 5, 30, 70, 120 and 0, 1, 4, 20, 60, 80 respectively}\label{fig:Supplement_Square_Cue_Conflict}
\end{figure*}

\newpage
\clearpage

\begin{figure}[!t]
\centering\includegraphics[width=0.75\columnwidth,clip=true,draft=false,]{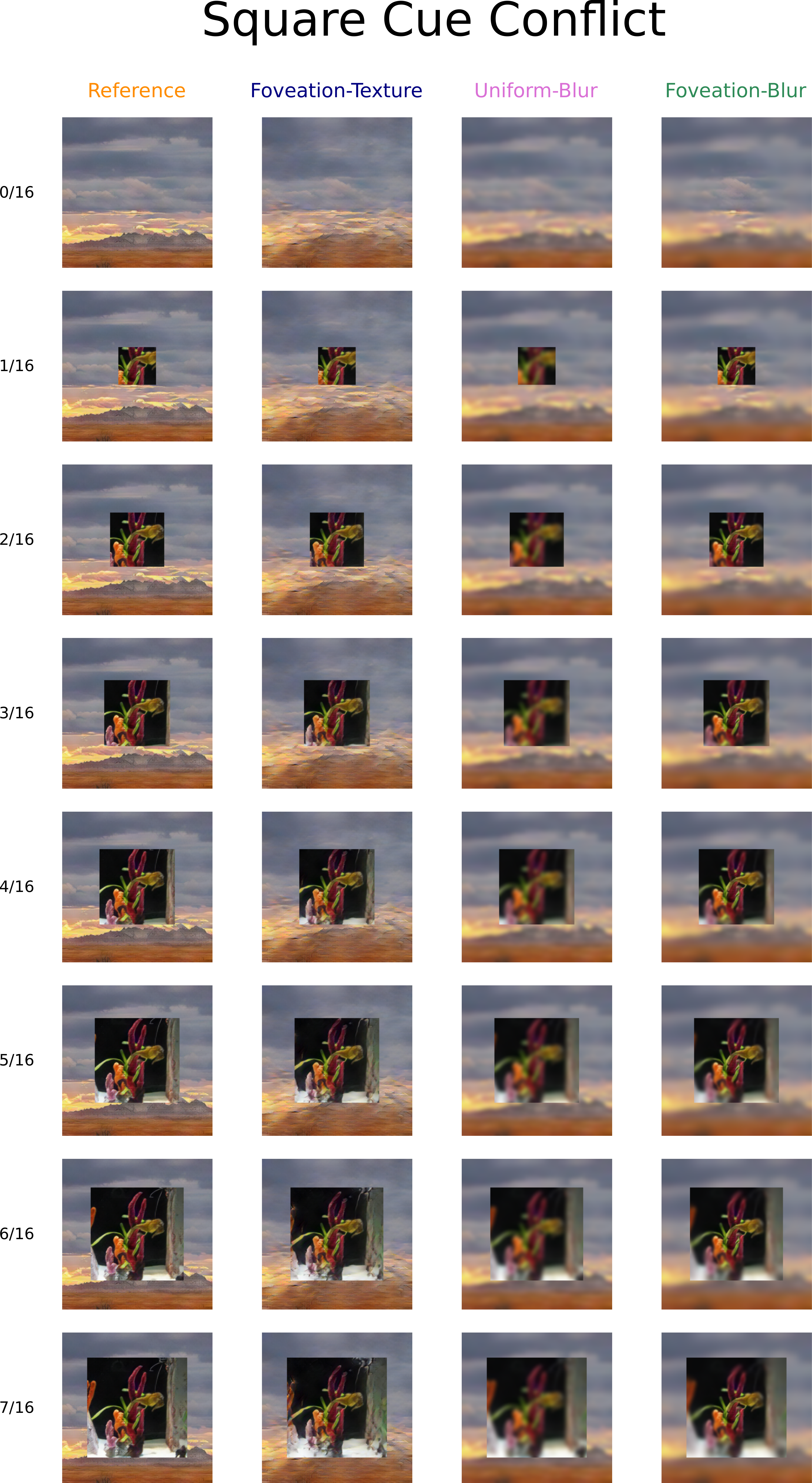}\caption{Sample Square Cue Conflict Stimuli.
}\label{fig:Square_Cue_Conflict_Stimuli_1}
\end{figure}

\newpage
\clearpage

\begin{figure}[!t]
\centering\includegraphics[width=0.75\columnwidth,clip=true,draft=false,]{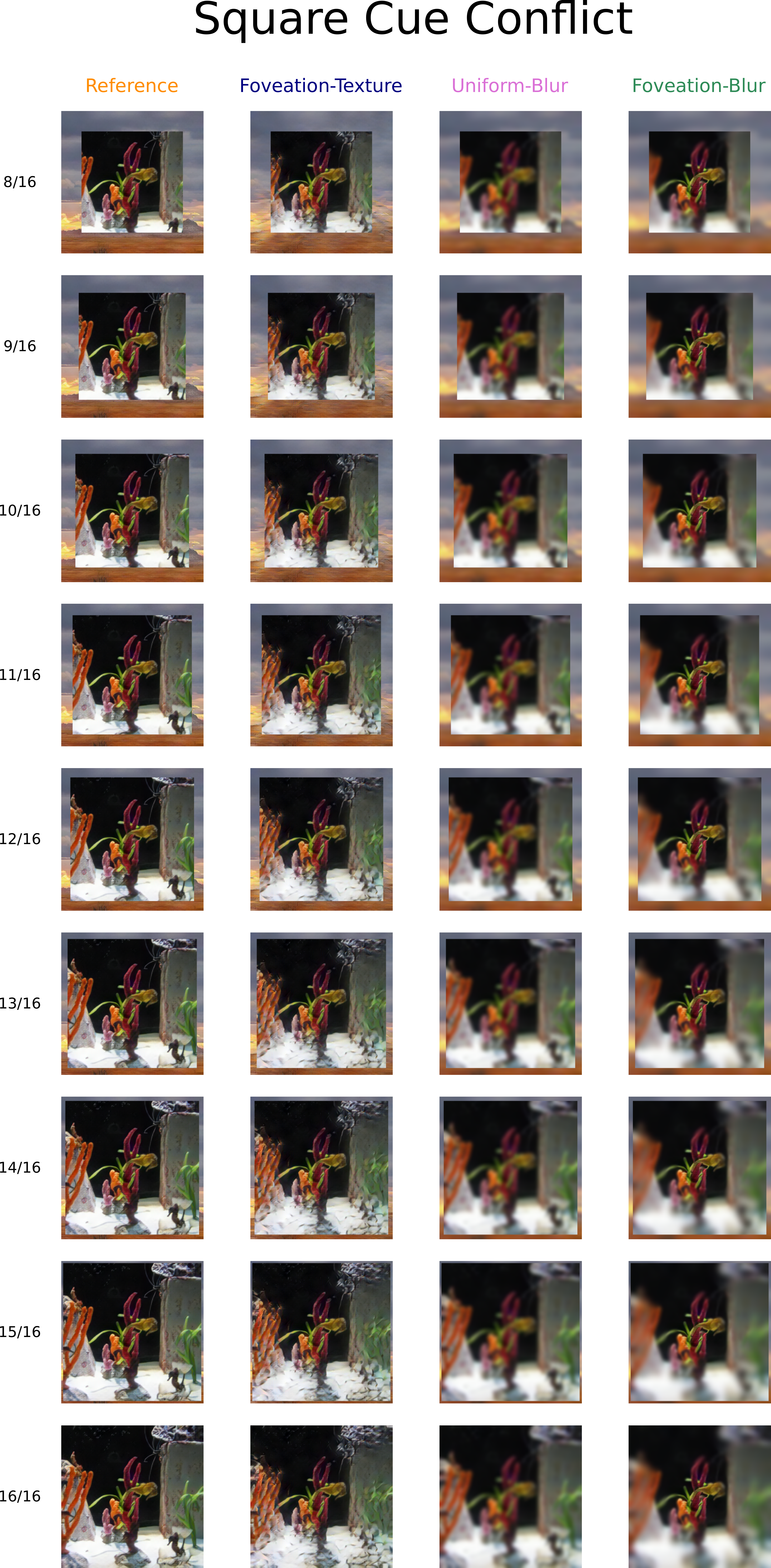}\caption{Sample Square Cue Conflict Stimuli.
}\label{fig:Square_Cue_Conflict_Stimuli_2}
\end{figure}

\newpage
\clearpage

\section{Differences from Previous Manuscript Versions}

[Added; this submission] Improved training and convergence of stage 2 neural networks. AlexNet + ResNet18 now have scheduled learning rates, weight decay and Nesterov momentum when trained with SGD for each image distribution.

[Added; this submission] High Pass and Low Pass Spatial Frequency experiments for grayscale stimuli as suggested in round of review from ICML 2021.

[Added; this submission] Square Uniform cue-conflict experiment to re-verify center image bias as suggested in round of review from ICML 2021.

[Added; this submission] Left2Right \& Top2Bottom experiments moved to main body.

[Added; this submission] both Aggregate and Individual plots for each system to qualitatively check for variance in individual network differences.

[Added; this submission] Visualization of filters from the first convolutional layer for each system.

[Added; this submission] Additional use of Mean Square Error, Mutual Information and 10 more IQA metrics from~\cite{2020arXiv200501338D} as supporting Image Quality Assessment metrics to compare to SSIM for Rate-Distortion Optimization as suggested through reviews in ICML 2021.

[Added; for ICML 2021] Sketched proof of Reference being a Perceptual Upper Bound.

[Added; for ICLR 2021] Rate-Distortion Optimization procedure to compute Uniform-Blur and Foveation-Blur.

[Added; for ICLR 2021] Improved written clarity, and re-emphasized focus of paper on Foveation w.r.t Machines (not humans -- which caused misinterpretation and rejection from NeurIPS 2020).

[Removed; for ICML 2021] Claims about Foveation-Texture inducing a shape bias (currently parallel work) from Submission to ICLR 2021. 

[Removed; for ICLR 2021] Experiments about data-augmentation via eye-movements + classical augmentation schemes such as random cropping + rescaling (parallel work) from Submission to NeurIPS 2020. 

[Bug fix; this submission] Even runs were continuations of odd runs in 10 run randomization across networks due to bug w.r.t distributed parallelization, from submission to ICML 2021. Note: General pattern of results did not change, and all curves have been re-plotted.

[Previous paper scores, decisions, meta-reviews and author opinions:] 
\begin{enumerate}
\item NeurIPS 2020: 5,4,3,4 (reject: Unanimous bad reviews, focus of all reviewers was a need for human psychophysical studies even though the paper was not about human vision -- which prompted us to re-write the paper to make our goals more clear: ``What is the impact of texture-based foveation on machines?; and what can these results tell us about the human visual system -- mainly the visual periphery that has texture-like computation -- from a computational perspective?''. [fixed])
\item ICLR 2021: 7,7,7,3,5 (reject: Mixed reviews \& needed to tone down claims and re-emphasize why texture was used in the periphery [fixed])
\item ICML 2021: 3 Weak Rejects (1 Accept + 1 Weak Accept downgraded their scores post-rebuttal suggesting the work was not a good fit for ICML), 1 Strong Reject (withdrawn: we caught a bug post-rebuttal phase in the process of code/data release that did not affect the main pattern or results, but required re-running all the experiments and overall improved the current version of the paper. Reviewers suggested different IQA metrics beyond SSIM to make comparisons for matched perceptual compression (we added MSE, Mutual Information, and 10 more IQA metrics). This has been added and addressed in our current version.).
\end{enumerate}

A recurrent theme in negative reviews has been that the model does not (in its current state) advance the state of the art by beating a baseline. While these hallmarks are pivotal for computer vision, our goal is complimentary, as we would like to model, and understand the representational consequences -- beyond accuracy -- of spatially-adaptive computation in machines inspired by the foveated visual system of humans.

\end{document}